\documentclass[10pt,twocolumn,letterpaper]{article}

\usepackage[pagenumbers]{cvpr} 

\usepackage[dvipsnames]{xcolor}
\usepackage{algorithm}
\usepackage{algpseudocode}

\usepackage{listings}
\usepackage{xcolor}

\lstdefinelanguage{JSON}{
    string=[s]{"}{"},
    comment=[l]{:},
    morecomment=[l]{:},
    commentstyle=\color{jsonkey},
    stringstyle=\color{jsonstring},
    identifierstyle=\color{jsonvalue},
}

\definecolor{jsonstring}{RGB}{206,145,120}
\definecolor{jsonkey}{RGB}{156,220,254}
\definecolor{jsonvalue}{RGB}{181,206,168}
\definecolor{jsonbracket}{RGB}{212,212,212}

\lstdefinestyle{json}{
    language=JSON,
    basicstyle=\ttfamily\scriptsize,
    numbers=none,
    showstringspaces=false,
    breaklines=true,
    frame=none,
    stringstyle=\color{jsonstring},
    keywordstyle=\color{jsonkey},
    commentstyle=\color{jsonvalue},
    morestring=[b]",
    literate=
     *{0}{{{\color{jsonvalue}0}}}{1}
      {1}{{{\color{jsonvalue}1}}}{1}
      {2}{{{\color{jsonvalue}2}}}{1}
      {3}{{{\color{jsonvalue}3}}}{1}
      {4}{{{\color{jsonvalue}4}}}{1}
      {5}{{{\color{jsonvalue}5}}}{1}
      {6}{{{\color{jsonvalue}6}}}{1}
      {7}{{{\color{jsonvalue}7}}}{1}
      {8}{{{\color{jsonvalue}8}}}{1}
      {9}{{{\color{jsonvalue}9}}}{1}
      {:}{{{\color{jsonbracket}{:}}}}{1}
      {,}{{{\color{jsonbracket}{,}}}}{1}
      {\{}{{{\color{jsonbracket}{\{}}}}{1}
      {\}}{{{\color{jsonbracket}{\}}}}}{1}
      {[}{{{\color{jsonbracket}{[}}}}{1}
      {]}{{{\color{jsonbracket}{]}}}}{1},
}

\definecolor{cvprblue}{rgb}{0.21,0.49,0.74}
\usepackage[pagebackref,breaklinks,colorlinks,allcolors=cvprblue]{hyperref}

\def\paperID{16} 
\def\confName{AI4CC}
\def\confYear{2025}

\title{Generating Animated Layouts as Structured Text Representations}

\author{
    Yeonsang Shin$^{1*}$ \qquad
    Jihwan Kim$^{1*}$ \qquad
    Yumin Song$^{1}$
    \\
    Kyungseung Lee$^{2}$ \qquad 
    Hyunhee Chung$^{2}$ \qquad 
    Taeyoung Na$^{2}$ 
    \\[5pt]
    $^1$Seoul National University \qquad 
    $^2$SK telecom
}

\begin{document}

\twocolumn[{%
    \maketitle
    \begin{center}
        \vspace{-3em}
        \includegraphics[width=0.95\textwidth]{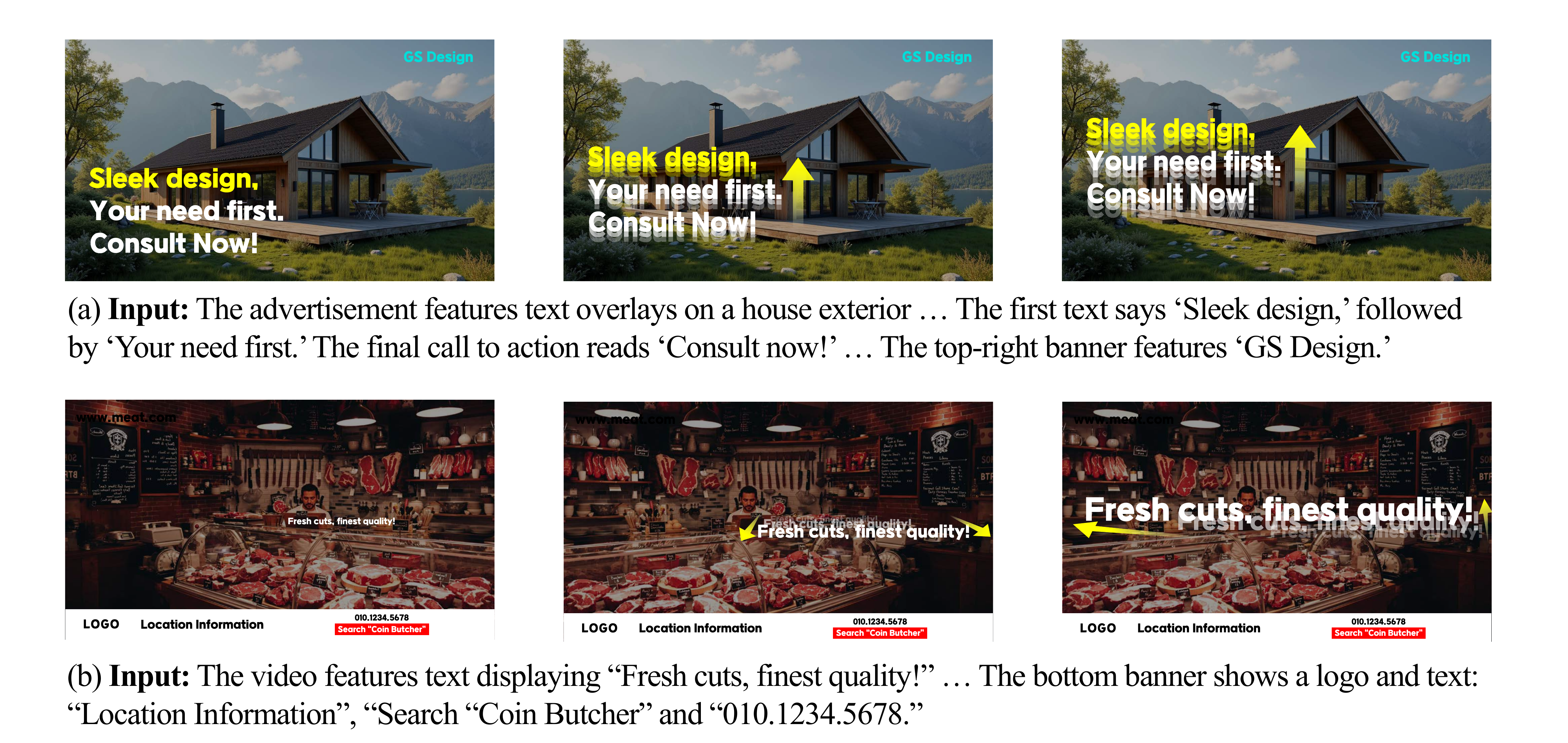}
    \end{center}
    \vspace{-1.5em}
    \captionof{figure}{\textbf{Results generated by VAKER.} Given text prompts that specify the content and style, VAKER creates animated layouts for video advertisements. Each row illustrates the sequential animation of elements layered over the base video content.}
    \vspace{1em}
    \label{fig:teaser}
}]

\newcommand\blfootnote[1]{%
  \begingroup
  \renewcommand\thefootnote{}\footnote{#1}%
  \addtocounter{footnote}{-1}%
  \endgroup
}


\begin{abstract}
\renewcommand{\thefootnote}{ }
\footnotetext{$^*$Equal contribution.}
\footnotetext[1]{Project page: \href{https://yeonsangshin.github.io/projects/Vaker}{https://yeonsangshin.github.io/projects/Vaker}}
\vspace{-2.3em}

Despite the remarkable progress in text-to-video models, achieving precise control over text elements and animated graphics remains a significant challenge, especially in applications such as video advertisements. To address this limitation, we introduce Animated Layout Generation, a novel approach to extend static graphic layouts with temporal dynamics. We propose a Structured Text Representation for fine-grained video control through hierarchical visual elements. To demonstrate the effectiveness of our approach, we present VAKER~(Video Ad maKER), a text-to-video advertisement generation pipeline that combines a three-stage generation process with Unstructured Text Reasoning for seamless integration with LLMs. VAKER fully automates video advertisement generation by incorporating dynamic layout trajectories for objects and graphics across specific video frames. Through extensive evaluations, we demonstrate that VAKER significantly outperforms existing methods in generating video advertisements.
\end{abstract}    

\vspace{-2mm}
\section{Introduction}
\label{sec:intro}
\vspace{-2.5mm}
Recent advances in text-to-video generation have demonstrated impressive capabilities in creating high-quality videos from text descriptions.
While current diffusion-based approaches~\cite{wang2024emu3,blattmann2023stable,yang2024cogvideox} excel at generating diverse videos with realistic scenes and movements, they struggle with readable text and precise control over animated elements.
This limitation is particularly evident in videos that combine dynamic graphical elements with base video content, such as social media videos with animated captions, promotional content with moving visual elements, or informational videos with synchronized text overlays.
The core challenge stems from two fundamental constraints: the difficulty in generating high-quality readable text within videos, and the lack of precise control over animated graphics that must be precisely positioned and timed on video content.

To address these challenges, we present \textit{``Animated Layout Generation"}, a generation task that extends traditional static graphic layouts~\cite{inoue2023layoutdmdiscretediffusionmodel, chai2023layoutdmtransformerbaseddiffusionmodel} to incorporate temporal dynamics.
Whereas conventional layouts organize visual objects through static bounding boxes, animated layouts add temporal control to these objects---managing how they appear, move, and transform over time.
This approach is particularly effective for creating videos with graphical elements that demand precise control over both spatial composition and temporal evolution of multiple visual elements.

The implementation focuses on \textit{Structured Text~(ST) Representation}, a novel representation that enables fine-grained control over video generation by organizing visual elements in a hierarchical structure.
By transforming videos into text sequences, this representation format preserves essential spatial and temporal relationships while enabling consistent mapping from text to visual output.
A key advantage of this text-based representation is its natural compatibility with Large Language Models~(LLMs), which have demonstrated remarkable capabilities across various domains.
Furthermore, the format excels at capturing on-screen text and spatial positioning, making it particularly effective for applications requiring detailed control over animated elements throughout video sequences.

In this paper, we present our approach through VAKER, a text-to-video advertisement generation pipeline that realizes Animated Layout Generation.
VAKER decomposes the generation process into three stages~(Banner, Mainground, and Animation), each leveraging ST-Representation to control spatial layout, visual attributes, and temporal dynamics.
Additionally, we incorporate Unstructured Text~(UT) Reasoning to effectively leverage LLM capabilities in translating user prompts into a structured format.
Extensive evaluations through qualitative assessments, quantitative metrics, and user studies demonstrate that VAKER significantly outperforms existing approaches in generating video advertisements.
Our work shows that the proposed approach successfully addresses the core challenges of precise text rendering and graphical control in video generation, contributing a robust framework for creating videos with dynamic visual elements.

\vspace{0.2em}
\noindent The main contributions of this paper are summarized below:

\begin{itemize}
    \item We introduce \textit{Animated Layout Generation}, a novel approach to extend traditional layout generation to the temporal domain. By incorporating time-based controls into layout composition, our approach enables precise manipulation of dynamic graphics and ensures text readability in videos.

    \item We develop \textit{ST-Representation}, a novel structured video representation that effectively captures temporal and spatial information in controllable text sequences. This design enables natural integration with LLMs while preserving detailed control over essential visual components.

    \item We present VAKER, a comprehensive text-to-video advertisement generation pipeline that demonstrates our approach in practice. Through rigorous evaluations, we show that VAKER generates superior results compared to existing approaches. 

\end{itemize}


\begin{figure*}[!t]
    \centering
    \includegraphics[width=0.8\linewidth]{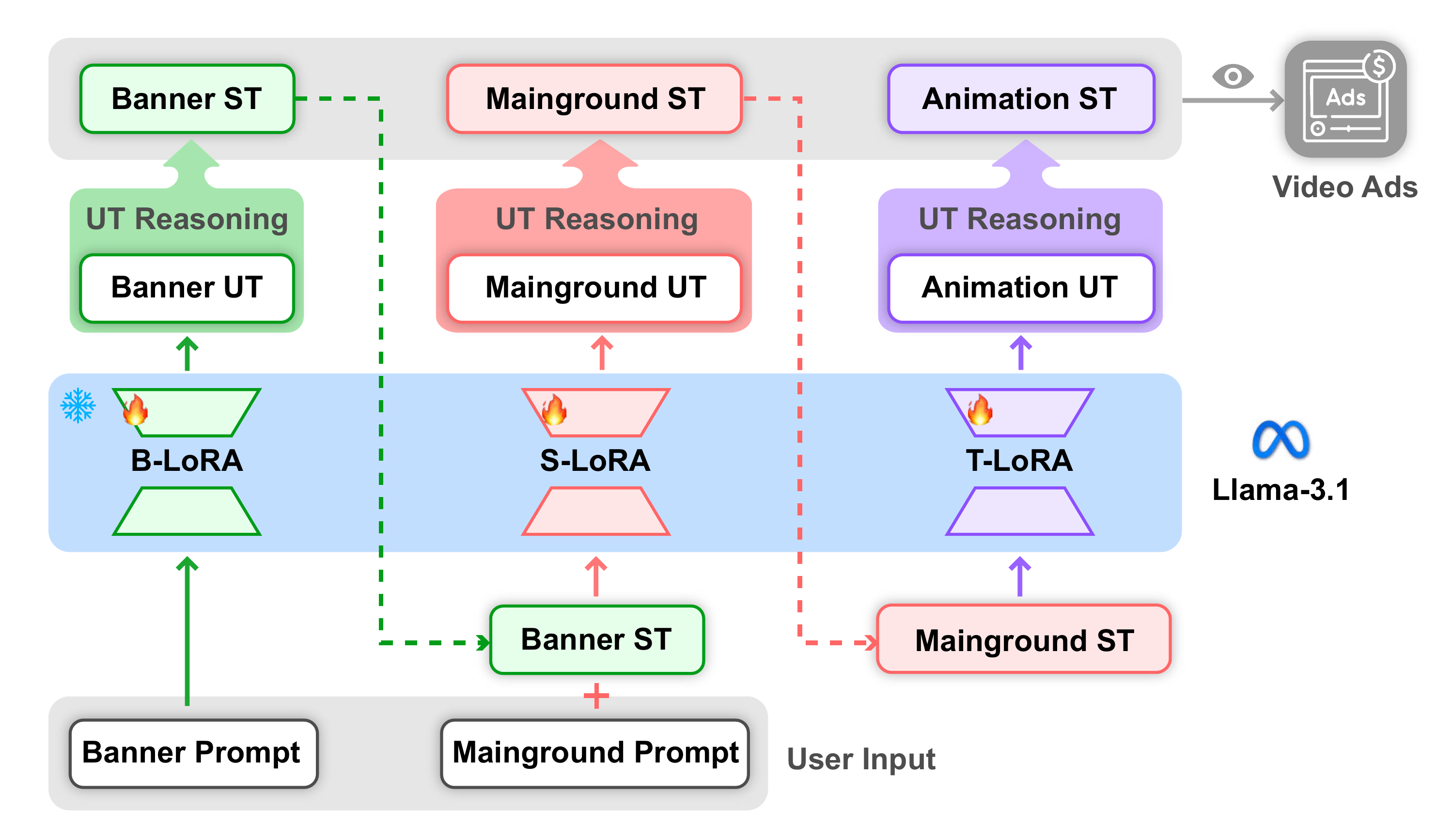}
    \caption{\textbf{Overview of VAKER}. VAKER takes text input from users to generate video advertisements. It does so by producing Structured Text~(ST) Representations, a format we propose for encoding video layouts as structured text. The pipeline breaks down video advertisement creation into three components: Banner, Mainground, and Animation. Each component uses a LoRA-adapted language model that first interprets the design in natural language before generating structured specifications for the final video advertisement. }
    \label{fig:model}
\end{figure*}

\section{Related work}
\label{sec:relwork}

\subsection{Text-to-video diffusion models}
\label{subsec:t2v}

Recent advancements in text-to-video diffusion models have demonstrated substantial capabilities in generating visually diverse and realistic videos from textual prompts, underscoring their potential in general video synthesis tasks.
Despite these strengths, current diffusion models face considerable challenges in handling specialized video content that requires embedded textual elements or fine-grained control of animated overlays.
This limitation stems partly from design choices in state-of-the-art video generation models~\cite{wang2024emu3,blattmann2023stable,yang2024cogvideox} that often exclude videos containing substantial text to mitigate visual artifacts in the dataset, consequently restricting the models' ability to generate integrated and readable text within videos.
Furthermore, most approaches rely on text encoders that transform text into embeddings used to condition the diffusion process, an architecture inherently unsuitable for precise text rendering and spatial control.
In response to these challenges, we introduce a Structured Text~(ST) Representation for animated layouts to enable the generation of video content with embedded text, ensuring clarity, spatial precision, and temporal consistency.

\subsection{Graphic layout generation}

Graphic layout generation is central to automated design, supporting efficient, visually compelling content across various media types.
Existing approaches in layout generation can be divided into two main categories based on their input format: constraint-based methods~\cite{chai2023layoutdmtransformerbaseddiffusionmodel, chen2024alignedlayoutgenerationdiffusion} and Text-to-layout methods~\cite{lin2023layoutprompterawakendesignability, lin2023parsethenplaceapproachgeneratinggraphic}.

\vspace{-1.5em}
\paragraph{Constraint-based.} Constraint-based layout generation relies on predefined rules including element categories, sizes, relationships, as well as refinement and completion requirements. 
The field has evolved from early approaches using GANs~\cite{Zheng19, zhou2022composition} and VAEs~\cite{yamaguchi2021canvasvaelearninggeneratevector, Arroyo_2021_CVPR} to recent advances in Transformer-based models~\cite{Gupta_2021_ICCV, inoue2023flexiblemultimodaldocumentmodels} and diffusion approaches \cite{chen2024alignedlayoutgenerationdiffusion, chai2023layoutdmtransformerbaseddiffusionmodel, inoue2023layoutdmdiscretediffusionmodel, hui2023unifyinglayoutgenerationdecoupled}.
Recent works have also explored LLM applications~\cite{tang2023layoutnuwarevealinghiddenlayout, lin2023layoutprompterawakendesignability, lin2023parsethenplaceapproachgeneratinggraphic}.
While these methods excel at creating structured layouts, their rigid constraint-based input format struggles to capture complex requirements for specialized content such as readable text overlays or synchronized visual elements, motivating the need for more flexible natural language specifications.

\vspace{-1.5em}
\paragraph{Text-to-layout.} Text-to-layout models~\cite{lin2023layoutprompterawakendesignability, lin2023parsethenplaceapproachgeneratinggraphic} generate layouts directly from natural language descriptions.
Recent approaches leverage LLMs: LayoutPrompter~\cite{lin2023layoutprompterawakendesignability} employs in-context learning without fine-tuning, while Parse-then-Place~\cite{lin2023parsethenplaceapproachgeneratinggraphic} utilizes a fine-tuned model to convert text into intermediate representations.
These methods enhance accessibility by enabling natural language specifications rather than technical constraints.
While these approaches handle static layouts effectively, they lack the capability to manage dynamic layouts where elements change over time---a limitation that our Animated Layout Generation specifically addresses.

\subsection{Large Language Models}

Large Language Models~(LLMs)~\cite{openai2024gpt4technicalreport, touvron2023llama2openfoundation} have transformed natural language processing by understanding and generating human-like text.
Beyond basic text generation, LLMs demonstrate sophisticated reasoning capabilities through various prompting techniques, including Chain-of-Thought prompting~\cite{wei2023chainofthoughtpromptingelicitsreasoning} for step-by-step problem solving.
Recent studies have revealed that LLMs possess inherent visual understanding capabilities. Despite being trained solely on text data, they can effectively reason about spatial relationships and visual concepts~\cite{sharma2024visioncheckuplanguagemodels}. 
These capabilities, combined with their proven success across diverse applications, have motivated our exploration of LLMs' potential for Animated Layout Generation.

\subsection{Video advertisement analysis}

Video advertisement analysis has primarily focused on understanding viewer behavior and commercial effectiveness.
Prior works have examined advertising rhetoric~\cite{hussain2017automaticunderstandingimagevideo}, narrative structures~\cite{Ye2018StoryUI}, and multimodal elements~\cite{bose2023mmautowardsmultimodalunderstandingadvertisement}.
However, the automated generation of video advertisements, particularly considering temporal dynamics and element positioning, remains largely unexplored.


\section{ST-Representation for videos}
\label{sec:st_rep}

The Structured Text~(ST) Representation forms the foundation of our approach to video generation, transforming animated layouts into a structured textual format with multiple advantages.
First, encoding video layouts as text sequences enables leveraging Large Language Models (LLMs), which excel at processing and generating text-based data.
Second, this text-based representation directly preserves on-screen text, making it particularly effective for applications requiring precise text rendering.
Third, the hierarchical key-value structure of ST-Representation effectively organizes complex visual information, maintaining essential spatial and temporal relationships within the content.
Finally, this structured format facilitates a reliable mapping from text to visual output, ensuring consistent visualization of text-based representations in the generated videos.

\vspace{-1em}
\paragraph{Structured video representation.} To efficiently manage diverse video components, we define a video clip $\mathbf V$ as a 4-tuple 
\vspace{-2mm}
\begin{equation}
\mathbf V=(\mathbf{B}, \mathbf{FG}, \mathbf{BG}, \mathbf{A}),
\label{eq:video_tuple}
\end{equation}
where $\mathbf{B}$, $\mathbf{FG}$, $\mathbf{BG}$, and $\mathbf{A}$ represent the Banner, Foreground, Background, and Animation information, respectively. 
We define each component as a structured form, which can be represented as structured text, such as JSON, XML, or YAML object.

The Banner
\vspace{-2mm}
\begin{equation}
\mathbf{B} = \{ B_j \}_{j\in \mathcal{J}}
\label{eq:B}
\end{equation}
denotes specific fixed regions and the objects inside them, with each $B_j$ defined as 
\vspace{-2mm}
\begin{equation}
B_j = \left( b_j^{\text{ban}}, y_j^{\text{ban}}, \{ o_m^{\text{ban}} \}_{m=1}^M \right).
\label{eq:banner}
\end{equation}
Here, $b_j^{\text{ban}}$ specifies the bounding box coordinates, $y_j^{\text{ban}}$ encodes banner attributes, and $\{ o_m^{\text{ban}} \}_{m=1}^M$ represents the set of $M$ objects within the banner.
We define attributes $y$ as supplementary properties, such as content and color, which are critical for accurate rendering within the layout.

The Foreground
\vspace{-2mm}
\begin{equation}
\mathbf{FG} = \{ o_n^{\text{fg}} \}_{n=1}^N
\label{eq:FG}
\end{equation}
comprises movable objects, such as images and text, that appear prominently within the scene and interact dynamically with other elements.

The Background
\vspace{-2mm}
\begin{equation}
\mathbf{BG} = (c^{\text{bg}}, y^{\text{bg}})
\label{eq:BG}
\end{equation}
represents type of the background $c^{\text{bg}}$ and its associated attributes $y^{\text{bg}}$.

The Animation
\vspace{-2mm}
\begin{equation}
\mathbf{A} = \left( l, \{ a_n^{\text{fg}} \}_{n=1}^N \right)
\label{eq:A}
\end{equation}
describes scene dynamics, where $l$ represents total scene duration in frame count, and each animation $a_n$ consists of a sequence of $K$ keyframes for the $n^\text{th}$ object. Each animation $a_n$ is defined as 
\vspace{-2mm}
\begin{equation}
a_n = \{ (f_k, b_k, t_k) \}_{k=1}^K,
\label{eq:animation}
\end{equation}
where $f_k$ is the keyframe index within the sequence, $b_k$ specifies the bounding box coordinates of the object at that frame, and $t_k$ represents additional transformation parameters~(\textit{e.g.}, rotation, transparency, scaling) that can be straightforwardly integrated to enable richer motion effects beyond positional changes.

\vspace{-1em}
\paragraph{Object information.} Each object $o_i$ within a scene is represented as a 3-tuple 
\vspace{-2mm}
\begin{equation}
o_i = (c_i, b_i, y_i),
\label{eq:object}
\end{equation}
where $c_i$ denotes the object class~(e.g., text, logo, image, etc.), $b_i$ specifies the bounding box coordinates as $b_i = [x_1, y_1, w, h]$, with $(x_1, y_1)$ representing the top-left corner coordinates and $(w, h)$ denoting the object's width and height, and $y_i$ encapsulates the object's specific attributes.

\begin{algorithm}[!t]
    \caption{Three-stage generation procedure of VAKER}
    \label{alg:vaker}
    \begin{algorithmic}
        \Require~$B_\theta$, $S_\phi$, $T_\psi$
        \State \textbf{Input:} $p^\text{b}, p^\text{m}$
        \State \textbf{Output:} $(\mathbf{B}, \mathbf{FG}, \mathbf{BG}, \mathbf{A})$
        \vspace{2mm}
        \State $\mathbf{B} \leftarrow B_\theta(p^\text{b})$ \Comment{B-LoRA}
        \State $\mathbf{FG}, \mathbf{BG} \leftarrow S_\phi(p^\text{m}, \mathbf{B})$ \Comment{S-LoRA}
        \State $\mathbf{A} \leftarrow T_\psi(\textbf{FG})$ \Comment{T-LoRA}
        \vspace{2mm}
        \State \textbf{return} $(\mathbf{B}, \mathbf{FG}, \mathbf{BG}, \mathbf{A})$
    \end{algorithmic}
\end{algorithm}

\begin{figure*}[!t]
    \centering
    \includegraphics[width=0.9\linewidth]{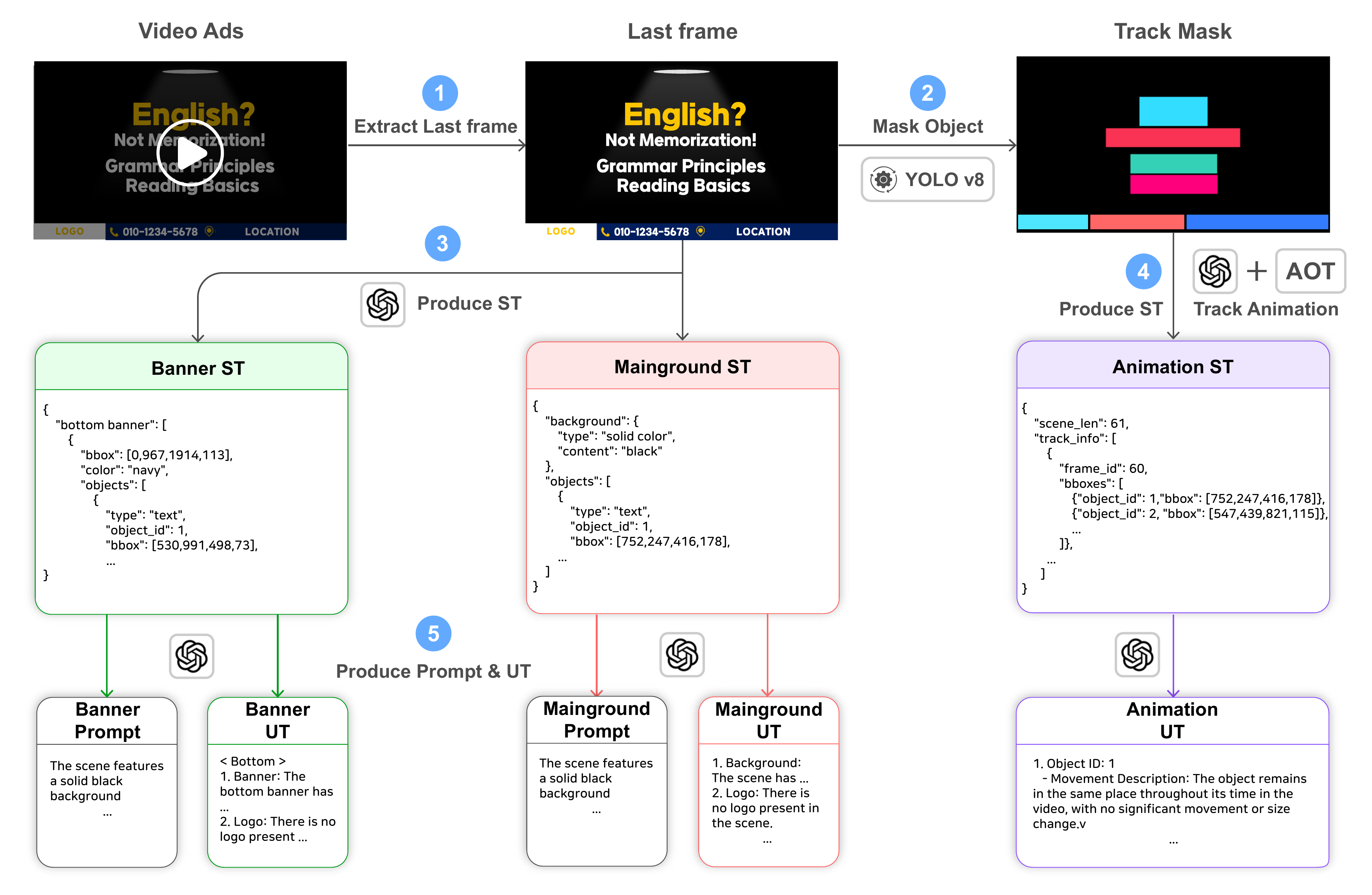}
    \caption{\textbf{Video Advertisement Dataset Construction Pipeline.} Our automated pipeline converts video advertisements into three types of data for training VAKER: Structured Text~(ST), Unstructured Text~(UT) Reasoning, and natural language prompts. Given a video advertisement, the pipeline performs the following steps: (1) extracts the last frame, (2) detects and classifies objects using fine-tuned detection models~\cite{vargheseyolov8}, and (3) generates Banner ST and Mainground ST from the spatial layout, (4) tracks object movements using tracking models~\cite{yang2021associatingobjectstransformersvideo} to generate Animation ST, and finally, (5) uses template-based prompting with LLMs to convert these ST-Representations into UT Reasonings and natural language prompts.}
    \label{fig:dataset}
\end{figure*}

\section{VAKER: Video Ad maKER}
\label{sec:vaker}
This section describes how we leverage ST-Representation and LLMs to generate realistic video advertisements.

\subsection{ST-Representation for video advertisements}
\label{subsec:st_rep_ads}

We adapt the Structured Text~(ST) Representation introduced in \cref{sec:st_rep} to address the specific requirements of video advertisements.

In video advertisements, banners are typically positioned at the bottom, top-left, or top-right of the frame, indicating:
\vspace{-2mm}
\begin{equation}
\mathcal{J} = \{\text{bottom}, \text{top-left}, \text{top-right}\},
\label{eq:J}
\end{equation}
for $\mathcal{J}$ in \cref{eq:B}. The attribute $y_j^\text{ban}$ in \cref{eq:banner} represents the banner color in this context.

The Background can be either a solid color or an image with varying attributes such that
\vspace{-2mm}
\begin{equation}
(c^{\text{bg}}, y^{\text{bg}})=\text{(solid color, color)}
\end{equation}
or
\vspace{-2mm}
\begin{equation}
(c^{\text{bg}}, y^{\text{bg}})=\text{(image, text caption)}.
\end{equation}

We focus on two primary object classes: text and logo. Each object $o_i$ in \cref{eq:object} is defined as either:
\vspace{-2mm}
\begin{equation}
c_i=\text{text}, y_i=\{\text{raw text, text color, textbox color}\}
\end{equation}
or
\vspace{-2mm}
\begin{equation}
c_i=\text{logo}, y_i=\text{None}
\end{equation}

For animation, we simplify to $a_n = \{(f_k, b_k)\}_{k=1}^K$, omitting transformation parameters $t_k$ to focus on positional changes in this first exploration of animated layouts.

Importantly, our ST-Representation incorporates color schemas, enabling VAKER to generate both layouts and visual attributes cohesively.
Detailed information about each component of the ST-Representation can be found in the supplementary materials.

\vspace{1em}
\subsection{Three-stage generation}
\label{subsec:MoE}

Training an LLM to generate dynamic content with precise spatial positioning and temporal movement would require extensive training data and computational resources.
To enable effective layout generation even with limited data, we decompose the process into three distinct stages, each handled by a specialized expert model fine-tuned with LoRA~\cite{hu2021loralowrankadaptationlarge}.

Our pipeline sequentially generates three Structured Text (ST) objects---Banner ST, Mainground ST, and Animation ST---based on user input prompts for the banner ($p^{\text{b}}$) and mainground ($p^{\text{m}}$).
Each ST object is encoded in JSON format.
We integrate both Foreground and Background into a single Mainground ST to effectively contextualize background information within the overall layout.

The generation process follows a sequential architecture.
Initially, the Banner $\mathbf{B}$ is generated by the banner expert $B_\theta$~(B-LoRA), using the banner prompt $p^{\text{b}}$.
Subsequently, the spatial expert $S_\phi$~(S-LoRA) generates Mainground ST based on the mainground prompt $p^{\text{m}}$ and the previously generated Banner $\mathbf{B}$.
Lastly, the temporal expert $T_\psi$~(T-LoRA) produces the animation $\mathbf{A}$, defining the movements of the foreground objects.
The complete generation process is presented in \cref{alg:vaker}.

\subsection{UT Reasoning}
\label{subsec:UT_Reasoning}

Generating structured text containing numerous bounding box coordinates directly from natural language prompts places a substantial computational burden on the model.
To address this challenge, we propose an Unstructured Text~(UT) Reasoning approach, inspired by the Chain-of-Thought methodology~\cite{wei2023chainofthoughtpromptingelicitsreasoning}, which first generates detailed natural language descriptions before converting them to structured text.

As illustrated in \cref{fig:model}, VAKER incorporates an explicit reasoning phase that produces comprehensive specifications prior to generating the Banner, Mainground, and Animation ST.
This intermediate reasoning step facilitates more precise and contextually appropriate content generation, effectively bridging the gap between natural language prompts and structured layout specifications.

\subsection{Dataset construction}

Manually converting video advertisements into ST-Representations would require a time-consuming and labor-intensive process, necessitating frame-by-frame annotation of object coordinates, text content, and color attributes.
To build a comprehensive dataset efficiently, we develop an automated pipeline~\cref{fig:dataset} that converts video advertisements into three types of data: Structured Text~(ST), Unstructured Text~(UT) Reasoning, and natural language prompts.
This pipeline extracts key visual information from videos using computer vision models and converts it into structured representations, which are then transformed into natural language descriptions using language models.
The complete pipeline processes each video in under a minute, with technical details provided in the supplementary materials.
Our dataset comprises 2,224 real video advertisements, provided specifically for research purposes, originally aired on commercial television broadcasts.
Each sample represents a single scene at 1920$\times$1080 resolution and 30 fps, with an average length of 50 frames.
The dataset encompasses diverse commercial categories, including restaurants, services, and retail products, capturing a wide spectrum of advertising layouts and visual design patterns.

\begin{figure*}[t]
    \includegraphics[width=\linewidth]{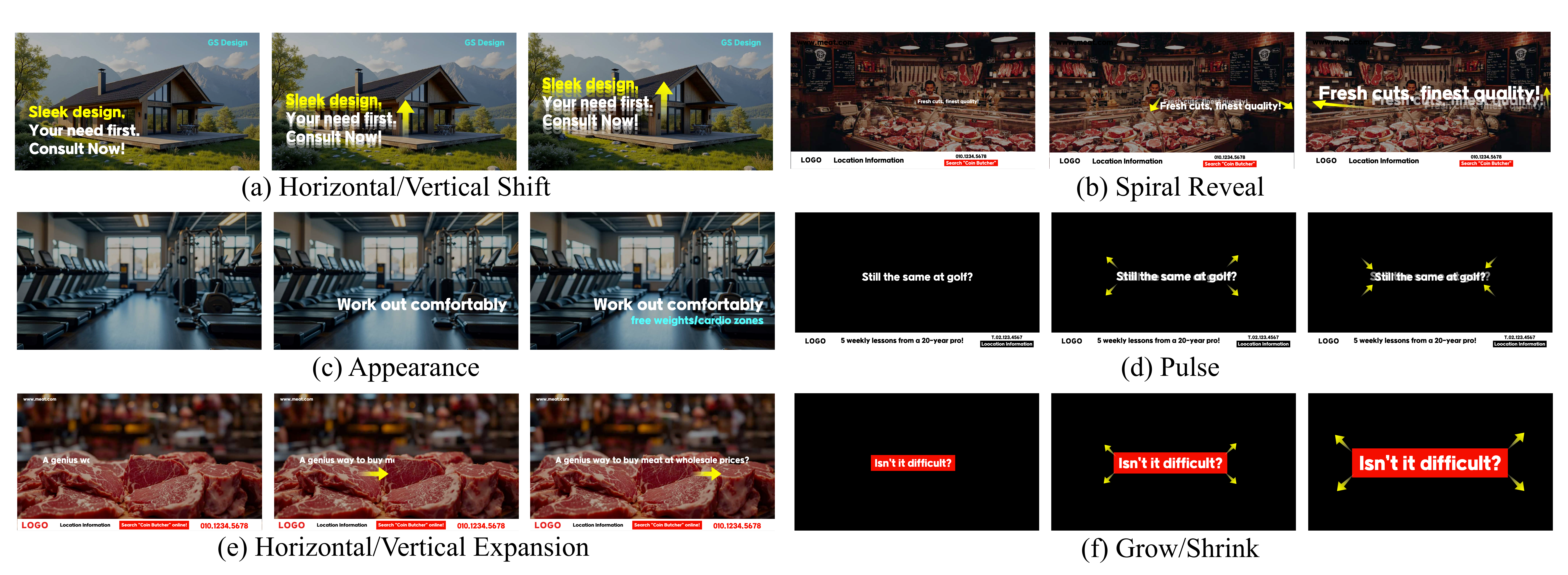}\vspace{-3mm}
    \caption{\textbf{Qualitative examples of animations generated by VAKER.} VAKER demonstrates diverse animation capabilities including~(a) horizontal/vertical shift,~(b) spiral reveal,~(c) appearance,~(d) pulse,~(e) horizontal/vertical expansion, and~(f) grow/shrink.}
    \label{fig:qual_results}
\end{figure*}

\begin{figure*}[t]
    \centering
    \includegraphics[width=\linewidth]{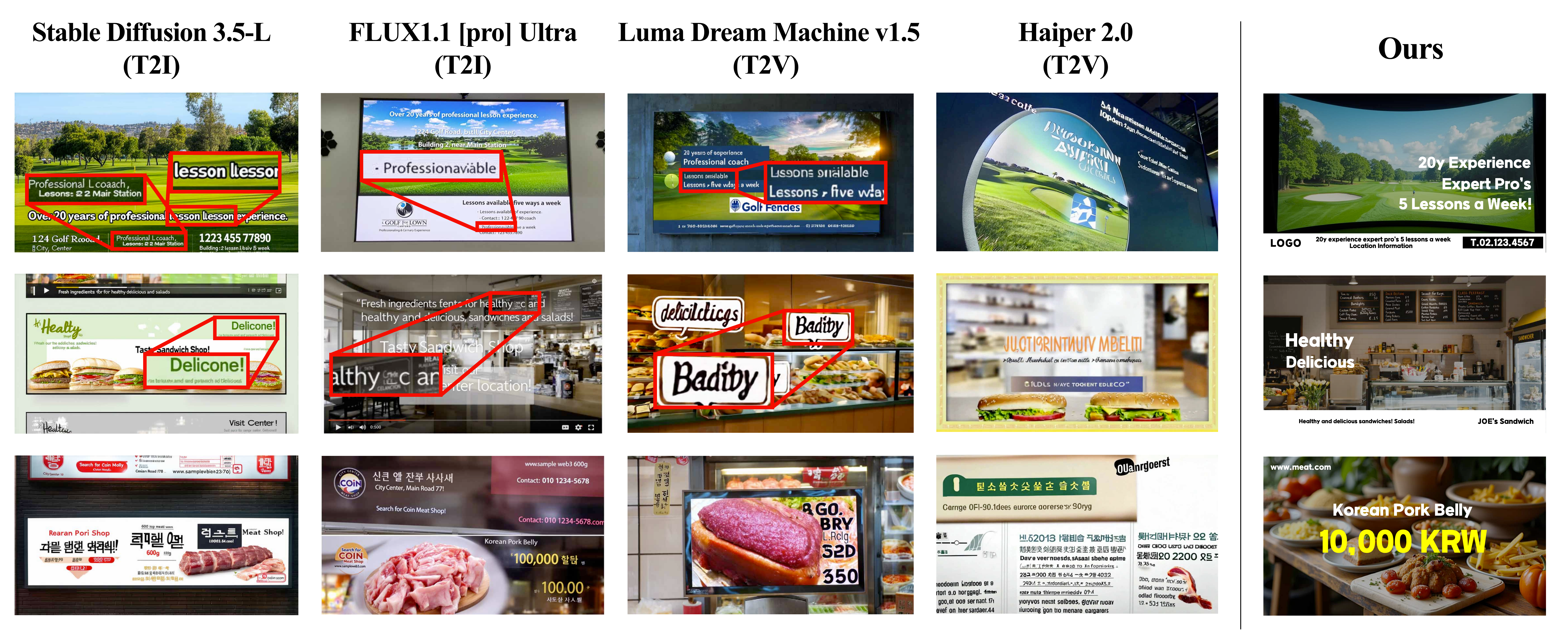}\vspace{-3mm}
    \caption{\textbf{Qualitative comparisons with diffusion models.} We compare VAKER with text-driven diffusion models~(T2I and T2V) for generating videos with the same prompts.\vspace{-3mm}}
    \label{fig:qual_DM}
\end{figure*}

\section{Experiment}
\label{sec:experiment}

\subsection{Experiment setting}

\paragraph{Implementation details.}
To train each expert, we fine-tune the Llama-3.1 70B-Instruct model~\cite{dubey2024llama3herdmodels} using LoRA.
Our dataset, consisting of 2,224 samples, is split into training~(90\%) and validation~(10\%) sets, with each expert trained until validation loss is minimized.
Each expert is guided by a task-specific system prompt that includes persona, task description, context, and format specifications, following the approach outlined in~\cite{google_gemini_prompting_guide}.
The additional details and complete prompts used for training are provided in the supplementary materials.

For qualitative visualization, we use OpenCV to render the generated ST-Representation, which is formatted as a JSON object.
Note that the background images shown in our results are generated from captions in $\mathbf{BG}$ using off-the-shelf image models: FLUX 1.1 dev~\cite{blackforestlabs_flux1_dev} and FLUX 1.1 Pro Ultra~\cite{blackforestlabs_flux}.

\vspace{-0.5em}
\paragraph{Baselines.}
As VAKER represents the first attempt at text-to-video advertisement generation, no directly comparable baselines exist for the complete framework.
To evaluate the two integrated stages of the framework (Banner ST and Mainground ST), we compare against established graphic layout generation models: LayoutPrompter~\cite{lin2023layoutprompterawakendesignability} for text-to-layout and constraint-explicit tasks, Desigen~\cite{weng2024desigen}, and Parse-Then-Place~\cite{lin2023parsethenplaceapproachgeneratinggraphic}.
Additionally, we compare VAKER against state-of-the-art diffusion models, including two text-to-video models~(Luma Dream Machine v1.5~\cite{lumalabs_dream_machine} and Haiper 2.0~\cite{haiper_haiper_2_0}) and two text-to-image models~(FLUX 1.1 Pro Ultra~\cite{blackforestlabs_flux} and Stable Diffusion 3.5 Large~\cite{huggingface_stable_diffusion_3_5_large}).
For the text-to-image models, we evaluate a two-stage approach (T2I + I2V) for video advertisement generation.

\vspace{-0.5em}
\paragraph{Metric for motion.}
Since our work is the first to generate animated layouts, we propose the Fréchet Motion Distance~(FMD) metric. FMD is based on the assumption that high-quality motion should resemble the distribution of motions in the training data. It quantifies this resemblance by comparing relative motion vectors $\Delta b(t)=b(t)-b(0)$ normalized by frame dimensions. Lower FMD values indicate more realistic motion. The formal equation and implementation details are provided in the supplementary material.

\vspace{-0.5em}
\paragraph{Layout evaluation metrics.} 
To quantitatively assess layout quality and compare performance with baseline models, we evaluate our framework using three additional metrics:
(1) Overlap~\cite{li2020attributeconditionedlayoutganautomatic} measures the average Intersection over Union~(IoU) between element pairs in a generated layout.
Lower values indicate better spacing between elements, ensuring text and visual components remain distinct and readable throughout the video.
(2) Maximum IoU~(mIoU)~\cite{Kikuchi_2021} calculates the highest IoU between elements in generated and real layouts derived from the same prompt. Higher values indicate better alignment with real-world design practices.
(3) Failure rate represents the percentage of generated layouts with invalid structures.
This includes JSON parsing failures~(malformed JSON syntax), missing required keys~(e.g., omitting attributes), incorrect structural hierarchies, missing requested elements, etc.

\begin{table}[!t]
    \centering
    \begin{small}
    \begin{tabular}{lcccc}
    \toprule
    Model & FMD $\downarrow$ & Overlap $\downarrow$ & mIoU $\uparrow$ & Failure $\downarrow$ \\
    \midrule
    LP~(T2L)~\cite{lin2023layoutprompterawakendesignability} & 0.0273 & 0.5990 & 0.3172 & 19.28\% \\
    PTP~\cite{lin2023parsethenplaceapproachgeneratinggraphic} & 0.0273 & 0.4343 & 0.2600 & 35.24\% \\
    \midrule
    Ours & \textbf{0.0103} & \textbf{0.4221} & \textbf{0.3376} & \textbf{5.41}\%\\
    \bottomrule
    \end{tabular}
    \end{small}
    \caption{\textbf{Comparison with text-to-layout baseline models.} Bold indicates best performance.\vspace{-2mm}}
    \label{tab:model_comparison}
\end{table}

\vspace{1em}
\subsection{Qualitative results}

\paragraph{Qualitative analysis.}

\cref{fig:teaser,fig:qual_results} demonstrate that VAKER generates high-quality layouts that accurately align with natural language input prompts.
As shown in \cref{fig:teaser}, VAKER successfully converts text descriptions into video advertisements with appropriate element placement and aesthetically pleasing color schemes.
\cref{fig:qual_results} further showcases VAKER's diverse animation capabilities.
Interestingly, several of these animation patterns emerged without explicit training examples, demonstrating the model's exceptional generative capabilities.

\vspace{-0.5em}
\paragraph{Comparison with diffusion models.}
\cref{fig:qual_DM} presents a qualitative comparison of our proposed method against existing diffusion-based models renowned for their typography generation, evaluating their capabilities in generating videos with embedded text and graphical overlays.
As shown in the third and fourth columns, T2V models~\cite{lumalabs_dream_machine,haiper_haiper_2_0} consistently fail to generate clear, legible text, often producing distorted or illegible text within the scene.
Similarly, T2I models~\cite{huggingface_stable_diffusion_3_5_large,blackforestlabs_flux}, despite claiming enhanced typography generation capabilities, struggle to produce clear text elements when extensive textual content is required, as demonstrated in the first two rows.
In contrast, VAKER excels in producing clear, precise typography by directly encoding text elements within the ST-Representation.

\begin{figure*}[t]
    \includegraphics[width=\linewidth]{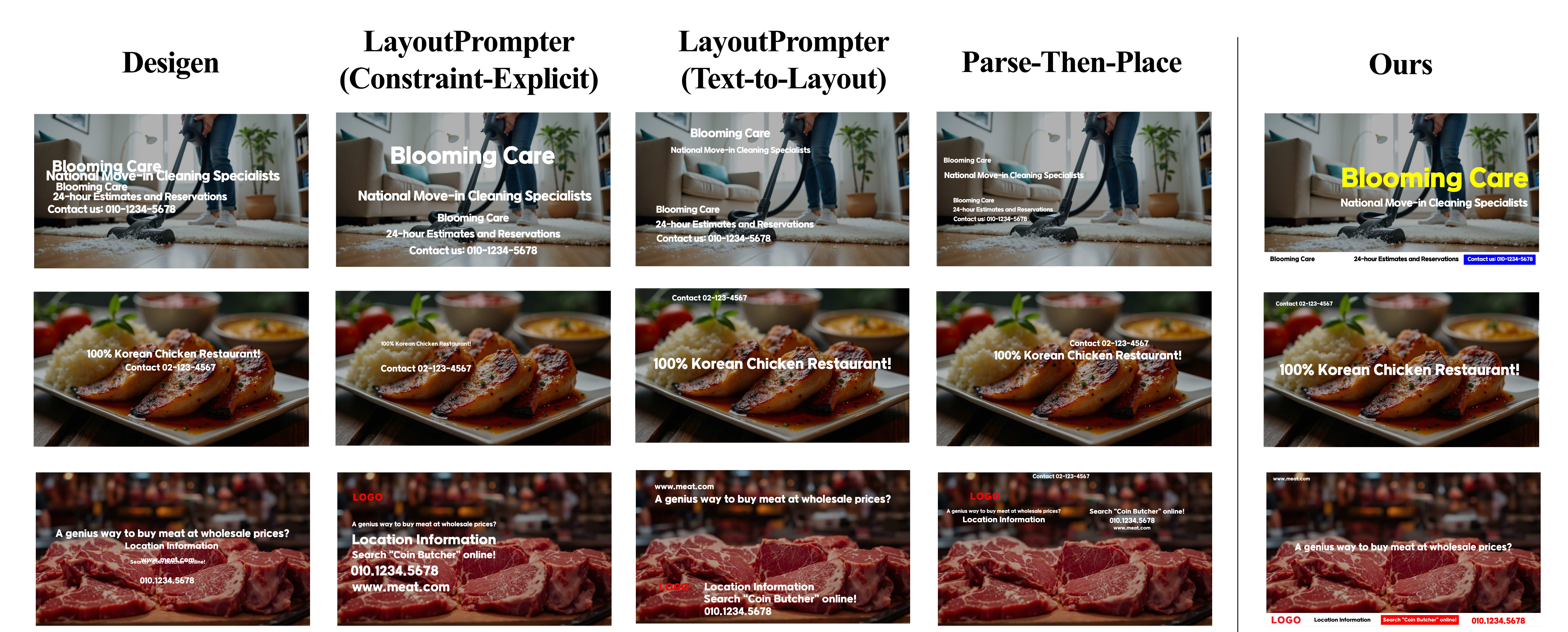}
    \caption{\textbf{Qualitative comparison with GLGs.} We compare VAKER with four baseline graphic layout generation models.}
    \label{fig:qual_GLG}
\end{figure*}

\vspace{-0.5em}
\paragraph{Comparison with layout generation models.}
Since our work is the first to introduce \textit{dynamic} layout generation, we compare our model against \textit{static} layout generation approaches.
\cref{fig:qual_GLG} presents a qualitative comparison with state-of-the-art graphic layout generation model baselines.
For text-to-layout models, LayoutPrompter~(Text-to-Layout)~\cite{lin2023layoutprompterawakendesignability} and Parse-Then-Place~\cite{lin2023parsethenplaceapproachgeneratinggraphic}, we provide input prompts identical to those used for our proposed model.
For Desigen~\cite{weng2024desigen} and LayoutPrompter~(Constraint-Explicit), which accept only object class information as input, we provide inputs consisting solely of object classes that align with our prompts.
To facilitate direct comparison in visualization, we uniformly render input text in white to match bounding box information.
As shown in \cref{fig:qual_GLG}, our approach demonstrates significant advantages over competing models by effectively incorporating text information.
Results indicate superior control over text length, refined color use, and precise layout placement for banner-related elements, highlighting our method's unique strengths in generating well-structured and visually cohesive layouts.

\vspace{1em}
\subsection{Quantitative results}

\paragraph{Quantitative analysis. }

\cref{tab:model_comparison} demonstrates that VAKER outperforms all baseline models across all metrics, achieving the lowest FMD and Overlap, highest mIoU, and lowest failure rate.
Notably, the failure rate of VAKER shows a significant reduction compared to baseline models, with approximately 3 to 6 times fewer instances of invalid layout generation.
As expected, LP and PTP are unable to produce motion, resulting in their low~(and identical) FMD scores.
This substantial improvement highlights VAKER's enhanced ability to maintain proper structure, generate dynamic motion, and consistently produce layouts that faithfully adhere to input prompts.

\begin{table}[htbp]
    \centering
    \begin{tabular}{lcccc}
    \toprule
    Setting & FMD $\downarrow$ & Overlap $\downarrow$ & mIoU $\uparrow$ & Failure $\downarrow$ \\
    \midrule
    Ours & \textbf{0.0103} & \textbf{0.4221} & \textbf{0.3376} & \textbf{5.41\%} \\
    \midrule
    w/o CoT & 0.0272 & 0.4391 & 0.2285 & 8.11\% \\
    w/o MoE & 0.0196 & 0.7565 & 0.2368 & 7.61\% \\
    w/o both & 0.0273 & 0.6401 & 0.2330 & 23.33\%  \\
    \bottomrule
    \end{tabular}
    \caption{\textbf{Ablation study for layout generation.} Bold indicates best performance.}
    \label{tab:ablation}
\end{table}

\subsection{Ablation studies}

To evaluate the design choices in the proposed pipeline, we conduct ablation experiments on two key components: Chain-of-Thought~(CoT) implemented via UT Reasoning~(\cref{subsec:UT_Reasoning}), and Mixture-of-Experts~(MoE) implemented through the three-stage generation process~(\cref{subsec:MoE}).
We assess their impact using four metrics: FMD, Overlap, mIoU, and Failure rate.

The results presented in \cref{tab:ablation} highlight the contributions of both components.
The exclusion of CoT results in higher failure rates, reduced layout quality, and degraded motion quality, underscoring its importance in the generation process.
Similarly, the absence of MoE notably affects element spacing, as reflected in the Overlap metric, and diminishes motion quality.
Removing both components leads to a substantial decline in performance across most metrics, with a particularly significant increase in failure rates and motion quality equivalent to static baselines~(\cref{tab:model_comparison}).
These findings indicate the complementary roles of CoT and MoE in ensuring stable and high-quality layout generation with effective motion.


\section{Conclusion}
\label{sec:conclusion}

In this work, we introduced Animated Layout Generation, extending traditional layout generation into the temporal domain through our novel Structured Text~(ST) Representation. 
We demonstrated its effectiveness through VAKER, a comprehensive system that employs a three-stage generation pipeline for creating video advertisements with precise control over text, visual attributes, and animated objects.
Despite current limitations---such as supporting only simple movements and partial information loss during the conversion from videos to STs---our experimental results demonstrate superior performance compared to both Video Diffusion Models and Graphic Layout Generation approaches.
This work establishes a new direction for video generation that enables precise control over both text rendering and animated graphics, opening new possibilities for automated content creation while maintaining control over spatial and temporal elements.
Future work could explore more complex animation patterns, dynamic visual effects, and improved information preservation during conversion, further expanding the capabilities of animated layout generation.

\newpage
{
    \small
    \bibliographystyle{ieeenat_fullname}
    \bibliography{main}
}

\clearpage
\onecolumn
\setcounter{page}{1}
\setcounter{section}{0} 
\setcounter{figure}{0}
\setcounter{algorithm}{0}

\maketitlesupplementary

\section{Acknowledgements}
This research was supported by Culture, Sports, and Tourism R\&D Program through the Korea Creative Content Agency~(KOCCA) grant funded by the Ministry of Culture, Sports, and Tourism~(MCST) in 2024~(Project Name: Development of Technology for Convergence Performance Planning and Production Platform to Revitalize the Production of Convergence Performance by Traditional Artist Dance Music, Project Number: RS-2024-00398536, Contribution Rate: 100\%)

\section{Details of VAKER}
\label{sec:details_vaker}

\subsection{Implementation details}

\paragraph{Training details.}
To train each expert, we fine-tune the Llama-3.1 70B-Instruct model~\cite{dubey2024llama3herdmodels} using LoRA with a rank of 16 and $\alpha = 32$.
The training is performed on 8$\times$ A6000 GPUs,  with a learning rate of 2e-4, a batch size of 1, 5 epochs for S-LoRA and T-LoRA, and 8 epochs for B-LoRA. The total training time ranged from 5 to 10 hours, depending on the context length.

\vspace{0.5em}
\subsection{Dataset construction}

\paragraph{ST-Representation extraction.}
The ST-Representation extraction process consists of two main components: spatial and temporal information extraction. 
Given that object appearances outnumber disappearances in our videos, we adopt reverse-chronological processing starting from the last frame.
For spatial extraction, we first analyze this last frame, where all information is encoded through bounding boxes---$b_j^{\text{ban}}$ for banners and $b_i$ for objects.
To detect and classify visual objects across banner positions $\mathcal{J}$ and object categories $\mathcal{C}$, we implement two fine-tuned YOLOv8~\cite{vargheseyolov8} models.
These models detect and classify objects into banners, texts, and logos, with banners further classified into positions $j \in \mathcal{J}$ based on their spatial coordinates.
We then leverage Vision-Language Models~\cite{openai2024gpt4technicalreport} to extract additional attributes: banner colors~($y_j^{\text{ban}}$), background information~($y^{\text{bg}}$, as semantic captions or colors), and object attributes~($y_i$, including text content and colors).

For temporal information extraction, we employ a pixel-wise tracking model~\cite{yang2021associatingobjectstransformersvideo} to obtain animation trajectories~($a_n^{\text{fg}}$) of the extracted foreground objects~($o_i^{\text{fg}}$).
The pixel-level outputs are transformed into bounding box representations through a post-processing algorithm. 
Following our reverse-chronological strategy, we track objects backward from the analyzed last frame, which ensures robust tracking by starting from the most complete set of objects.

\vspace{-0.75em}
\paragraph{UT Reasoning and prompt extraction.}
Using template prompts with LLMs~\cite{openai2024gpt4technicalreport} through in-context learning, we generate two types of descriptions from ST-Representations. 
For UT Reasonings, we convert ST-Representations into template-formatted descriptions.
For prompts, we synthesize ST-Representations into natural user-like descriptions of 2-3 sentences.
These generated prompts, UT Reasonings, and their corresponding ST-Representations are then used for fine-tuning our model.

\vspace{-0.75em}
\paragraph{YOLO fine-tuning}
To automate the extraction of bounding boxes from videos, we fine-tune two YOLOv8~\cite{vargheseyolov8} detection models---one specialized for detecting banner regions and the other for detecting logos and text.
These models are trained on a dataset of 632 video clips annotated by humans, ensuring high-quality ground truth annotations.
We employ two distinct models for detection to optimize performance and ensure efficient processing.
For the full dataset of 2,224 video clips, the remaining 1,592 clips were annotated using the trained models, leveraging the human-annotated data as a foundation.

\vspace{-0.75em}
\paragraph{Bounding box post-processing}
We introduce a robust bounding box post-processing algorithm~(\cref{alg:bbox}) that converts pixel-wise tracking results into rectangular bounding boxes while handling tracking errors. Since pixel-level tracking can contain noise, our algorithm first searches between minimum and maximum coordinates of the mask, employing bidirectional scanning to detect empty rows (all-zero lines). To handle cases where noise causes premature detection of empty rows, we employ early-stop detection with 30-70\% thresholds: if an empty row is found too early (above 70\% of height in top-down scan or below 30\% in bottom-up scan), we trigger additional middle-region verification. The subsequent boundary refinement uses 50\% active ratio thresholds, and boxes smaller than 20\% of the maximum historical size are excluded for temporal consistency.

\clearpage
\section{User study}
\begin{figure}[!h]
    \centering
    \includegraphics[width=0.5\linewidth]{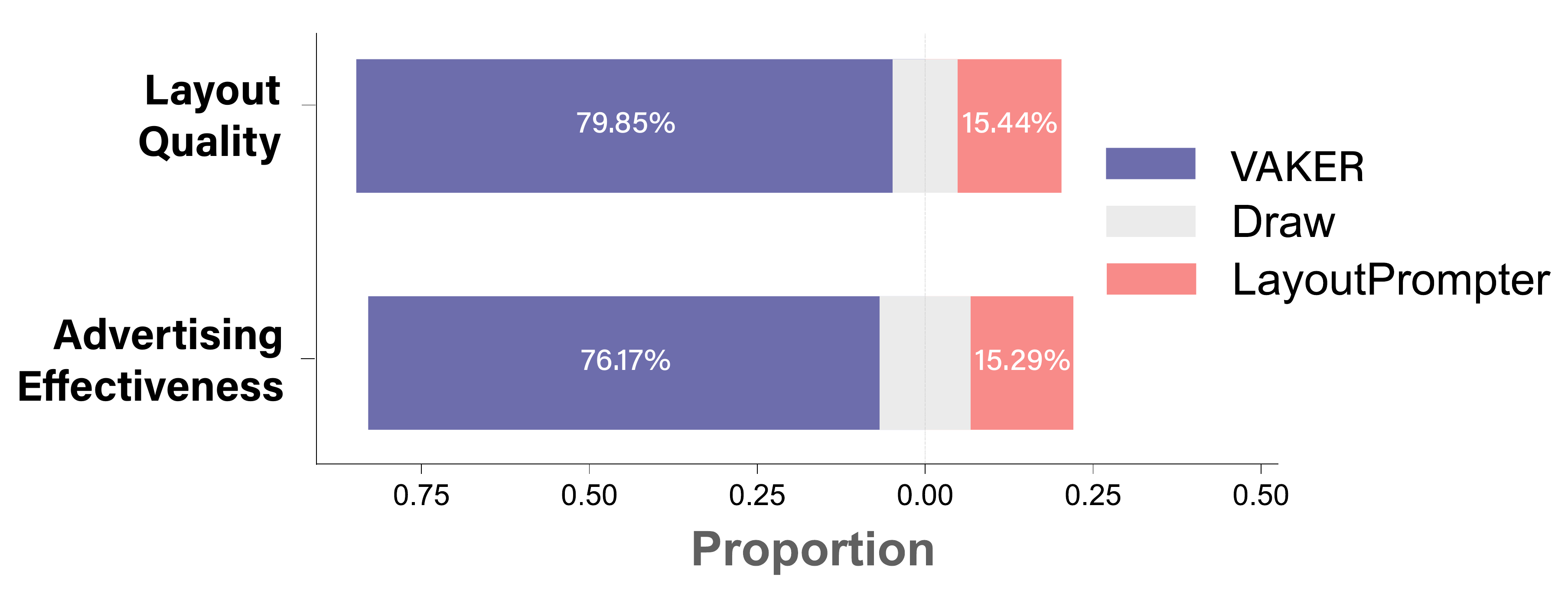}
    \caption{\textbf{User study results.} We compare VAKER with LayoutPrompter for Layout Quality and Advertising Effectiveness. VAKER outperforms the baseline in both criteria.}
    \label{fig:user_study}
\end{figure}

We conduct a user study to evaluate our results against LayoutPrompter~(T2L)~\cite{hsu2023posterlayoutnewbenchmarkapproach}, which exhibits the lowest failure rate among the baselines.
We recruit 30 participants and assign 20 comparison examples, focusing on two criteria: layout quality and advertising effectiveness.
Layout quality assesses the aesthetic arrangement of elements, while advertising effectiveness measures the promotional impact.
Participants are asked to select the superior result or indicate a draw for each criterion.
As shown in \cref{fig:user_study}, our approach significantly outperforms the state-of-the-art text-to-layout model in both criteria.

\section{Fréchet Motion Distance (FMD)}

Fréchet Motion Distance (FMD) quantifies the similarity between distributions of motion trajectories in generated and ground-truth data. This section provides detailed information about our implementation.

\vspace{1em}
\subsection{Motion Vector Extraction}

Given a sequence of bounding boxes for each object, we extract motion vectors as follows:

For each object, we identify its initial bounding box $b(0) = (x_0, y_0, w_0, h_0)$ at frame $t=0$. We normalize all coordinates by the frame dimensions (width $W$ and height $H$):

\begin{equation}
    b_{norm}(t) = \left(\frac{x_t}{W}, \frac{y_t}{H}, \frac{w_t}{W}, \frac{h_t}{H}\right)
\end{equation}

For each subsequent frame $t > 0$, we compute the relative motion vector:

\begin{equation}
    \Delta b(t) = b_{norm}(t) - b_{norm}(0) = \left(\Delta x_t, \Delta y_t, \Delta w_t, \Delta h_t\right)
\end{equation}

where $\Delta x_t, \Delta y_t$ capture position changes and $\Delta w_t, \Delta h_t$ capture size changes.

This results in a set of 4-dimensional vectors representing how each object moves relative to its initial position and size.

\vspace{1em}
\subsection{Fréchet Distance Computation}

Given two sets of motion vectors (from ground truth and generated sequences), we compute the mean vector $\mu \in \mathbb{R}^4$ and covariance matrix $\Sigma \in \mathbb{R}^{4 \times 4}$ for each distribution. The FMD is then calculated using:

\begin{equation}
    \text{FMD} = \|\mu_r - \mu_g\|^2 + \text{Tr}(\Sigma_r + \Sigma_g - 2\sqrt{\Sigma_r\Sigma_g})
\end{equation}

where $\mu_r, \Sigma_r$ are the mean and covariance of the real data distribution, $\mu_g, \Sigma_g$ are the mean and covariance of the generated data distribution, and $\sqrt{\Sigma_r\Sigma_g}$ is the matrix square root of the product $\Sigma_r\Sigma_g$.

\clearpage

\section{Qualitative results}
\label{sec:more_qual}

\begin{figure*}[h]
    \centering
    \renewcommand{\arraystretch}{1.5}
    \scalebox{0.9}{
    \setlength{\tabcolsep}{1pt} \hspace{-0.5mm}
        \hspace{-3mm}
        \begin{tabular}{ccccc}
	        \includegraphics[width=0.2\linewidth]{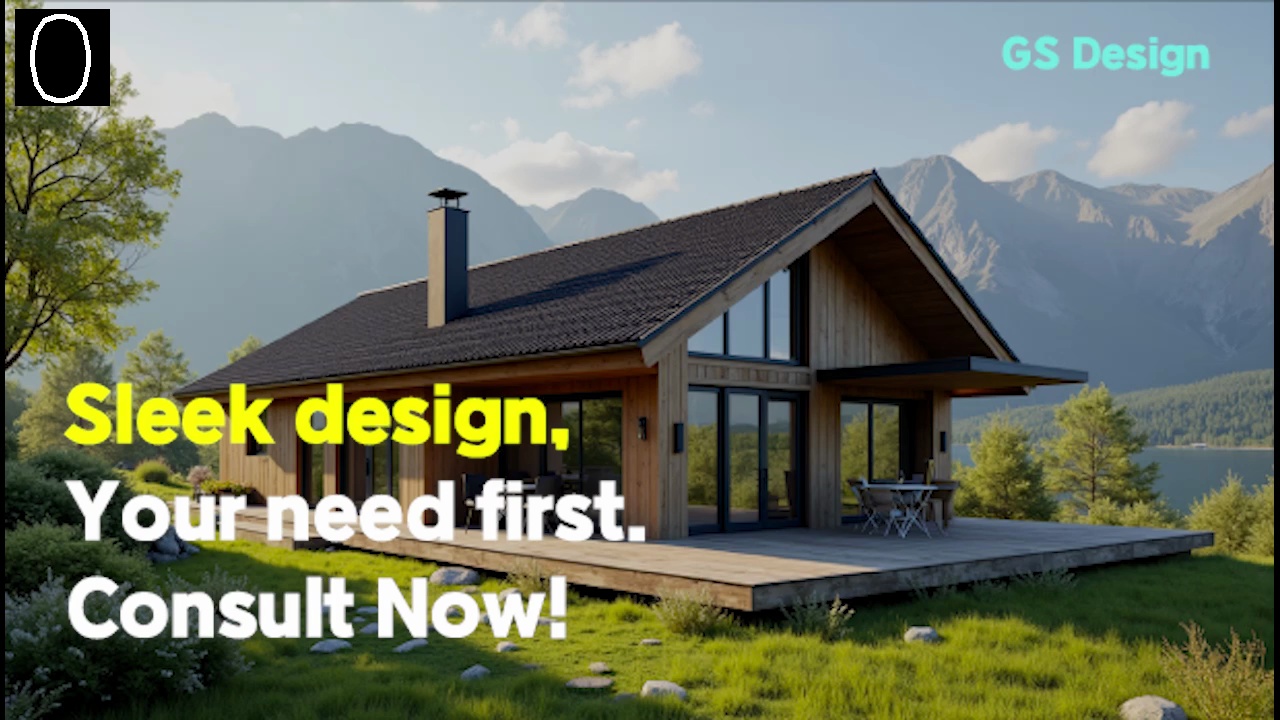} &
            \includegraphics[width=0.2\linewidth]{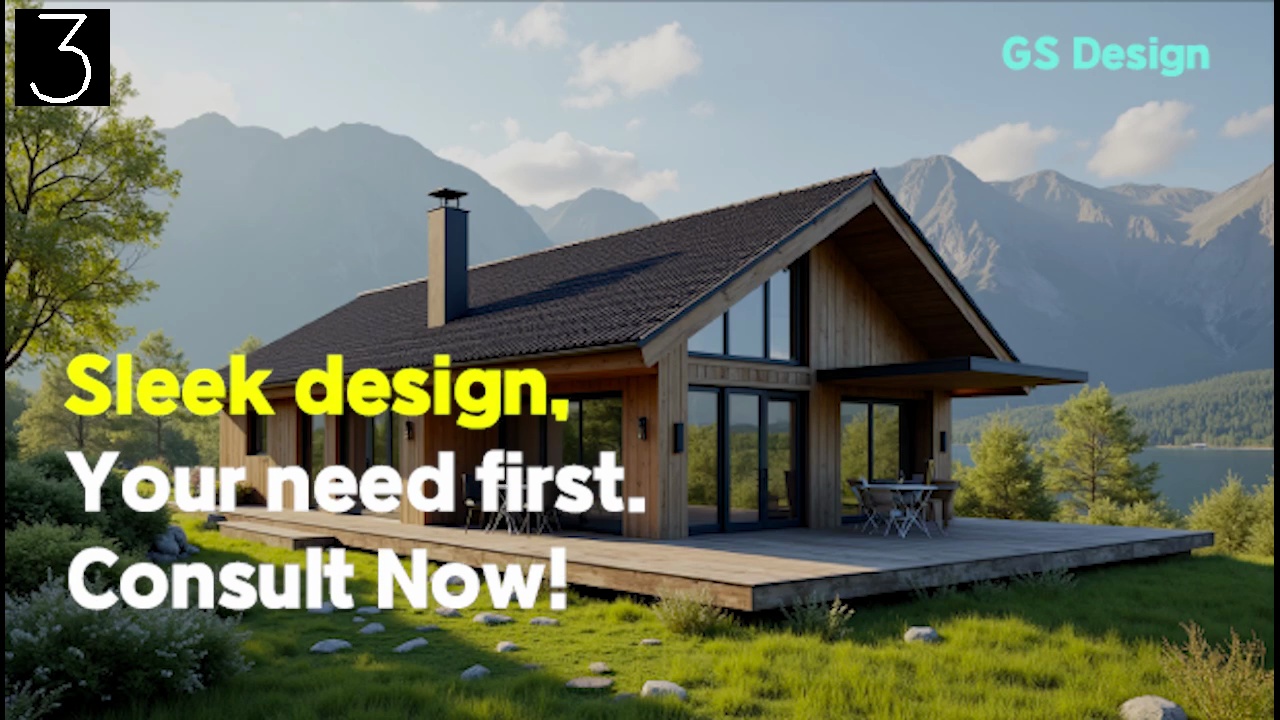} &
            \includegraphics[width=0.2\linewidth]{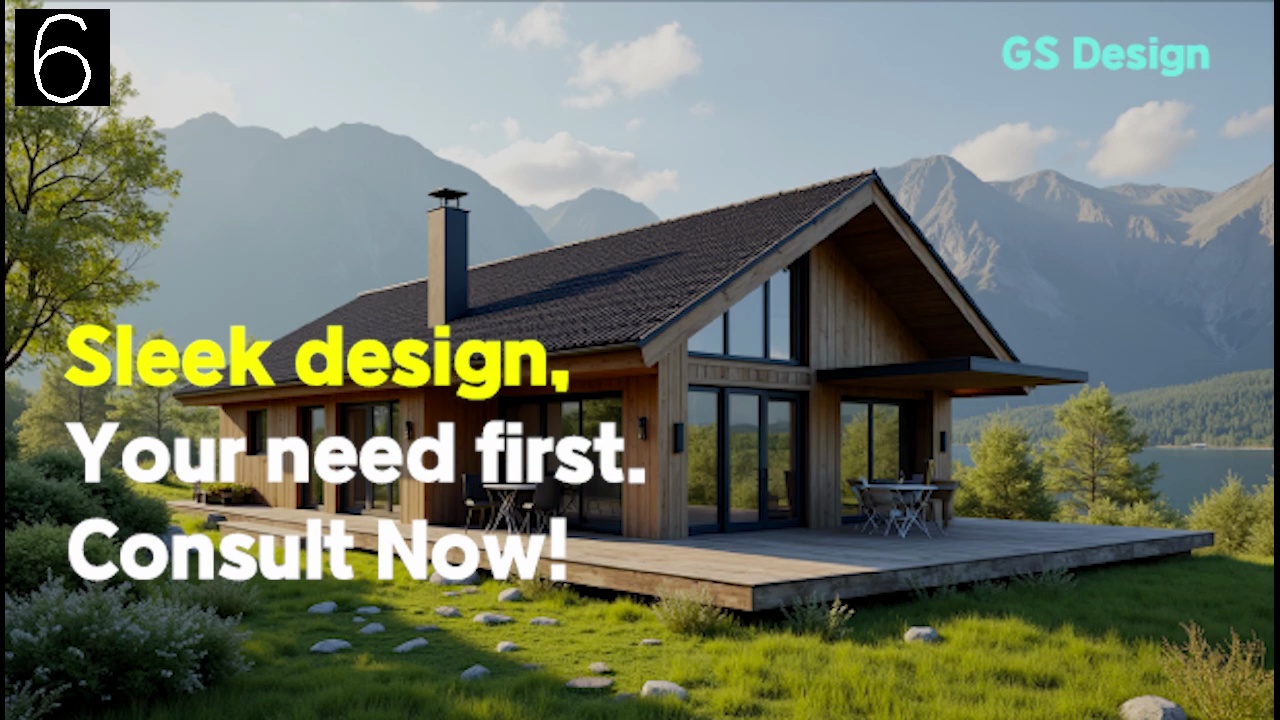} &
            \includegraphics[width=0.2\linewidth]{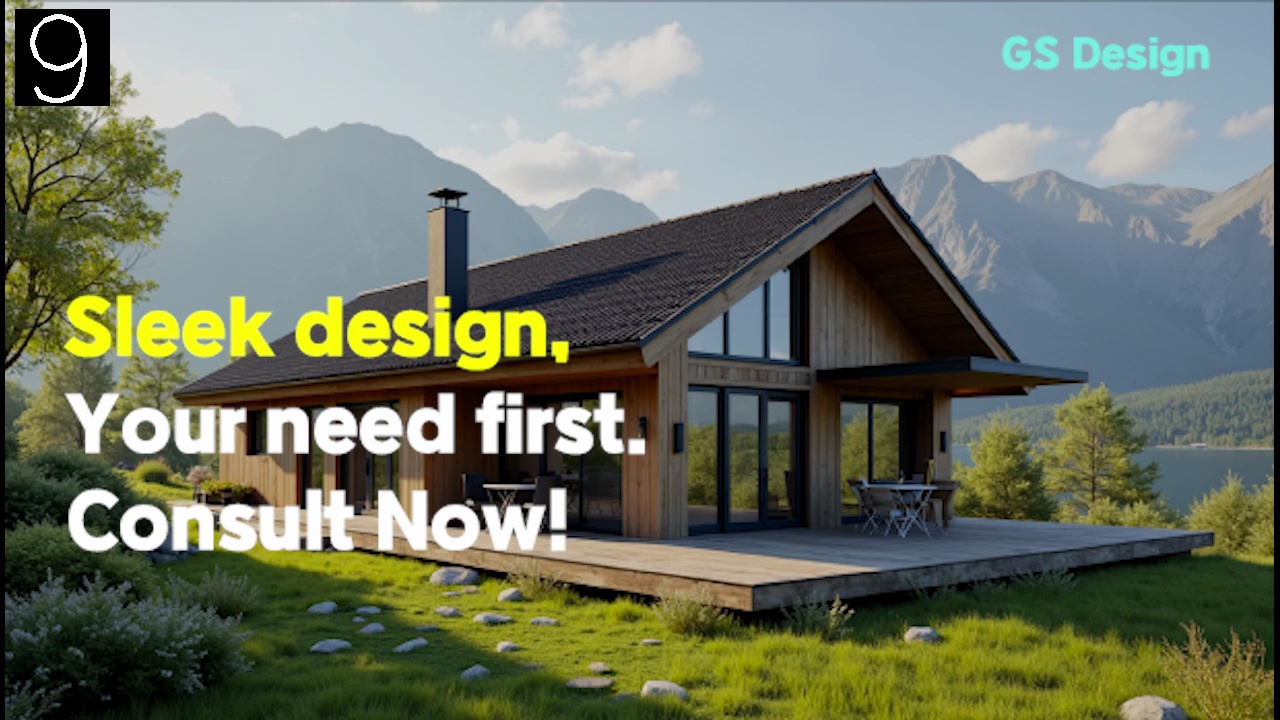} &
            \includegraphics[width=0.2\linewidth]{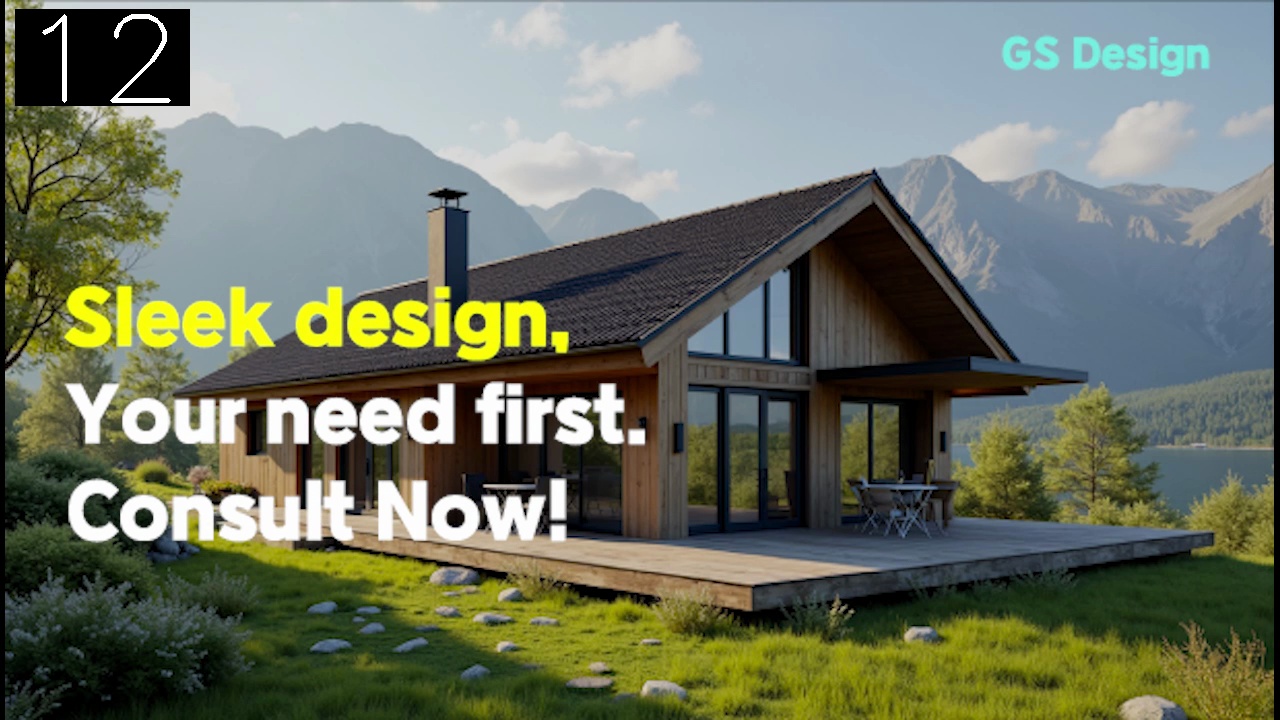} \\
            \multicolumn{5}{c}{\parbox{\linewidth}{\small (a) \textsf{The top-right banner features the text ``GS Design." There are no banners at the bottom.
The advertisement features a series of text elements over an image of a house with trees and mountains in the background. It begins with ``Sleek design," followed by ``Your need first." and ``Consult Now!" promoting expertise in various construction types.}\vspace{3mm}}}\\
            \includegraphics[width=0.2\linewidth]{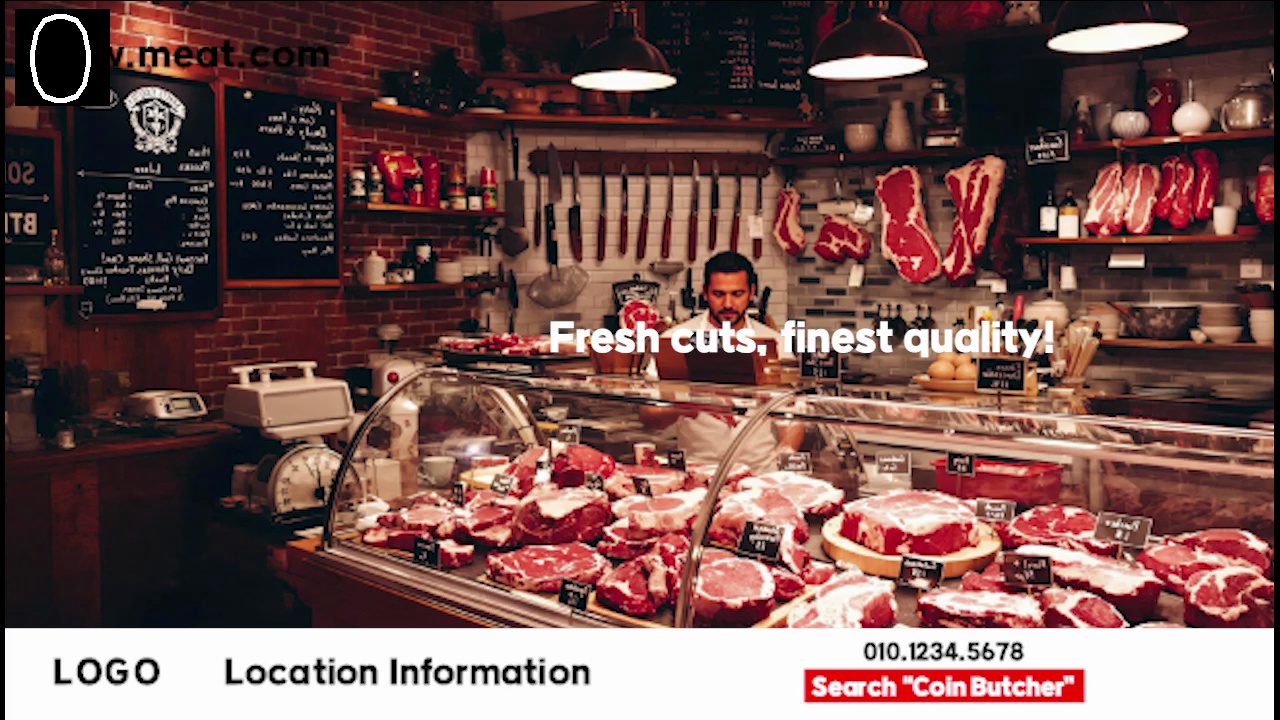} &
            \includegraphics[width=0.2\linewidth]{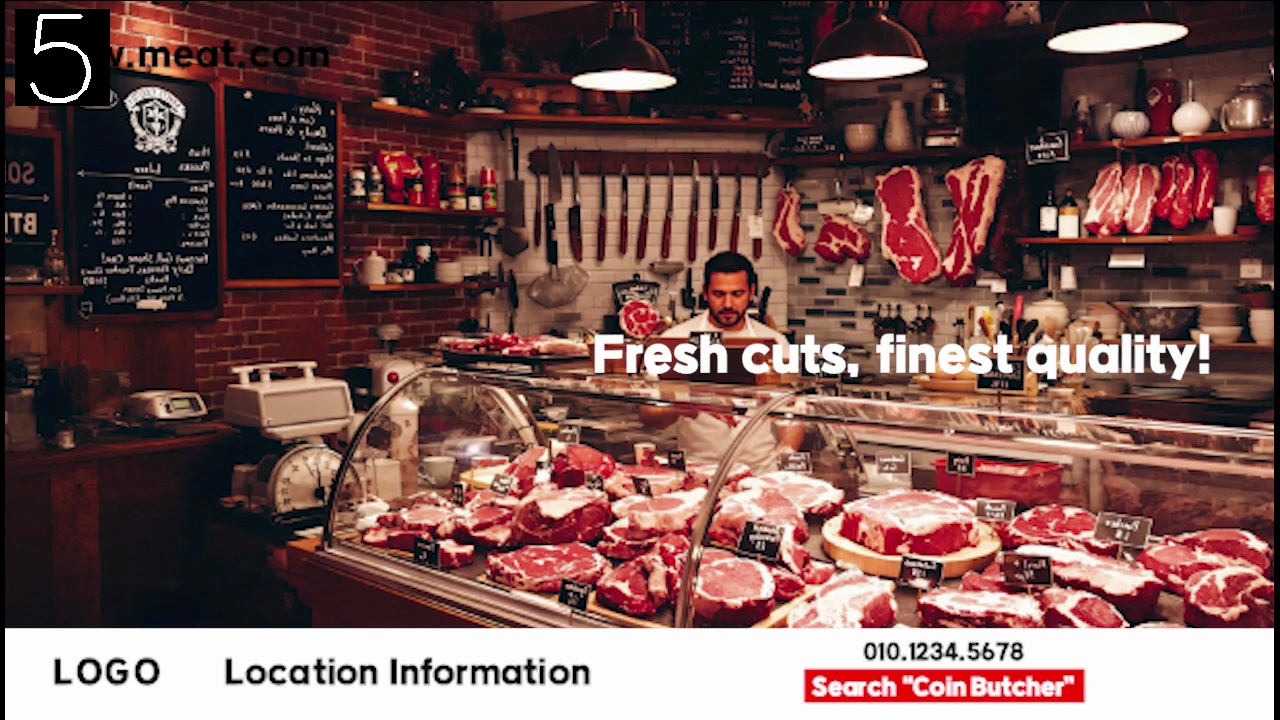} &
            \includegraphics[width=0.2\linewidth]{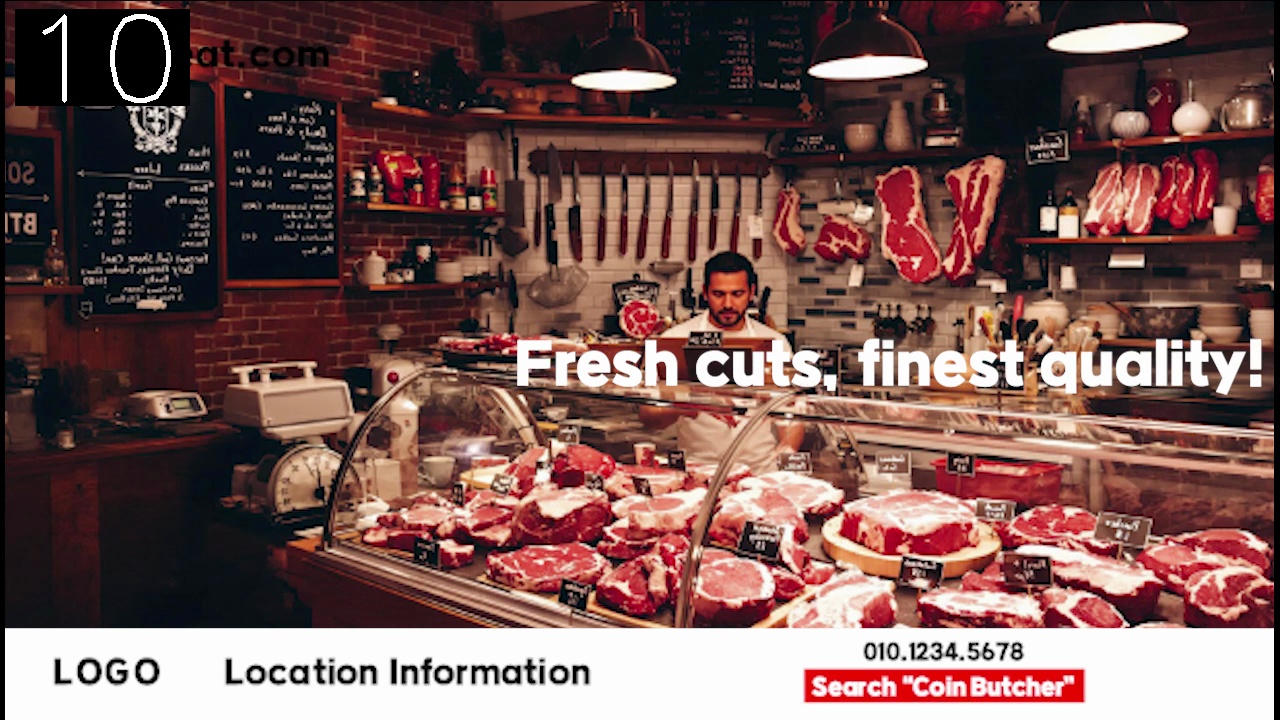} &
            \includegraphics[width=0.2\linewidth]{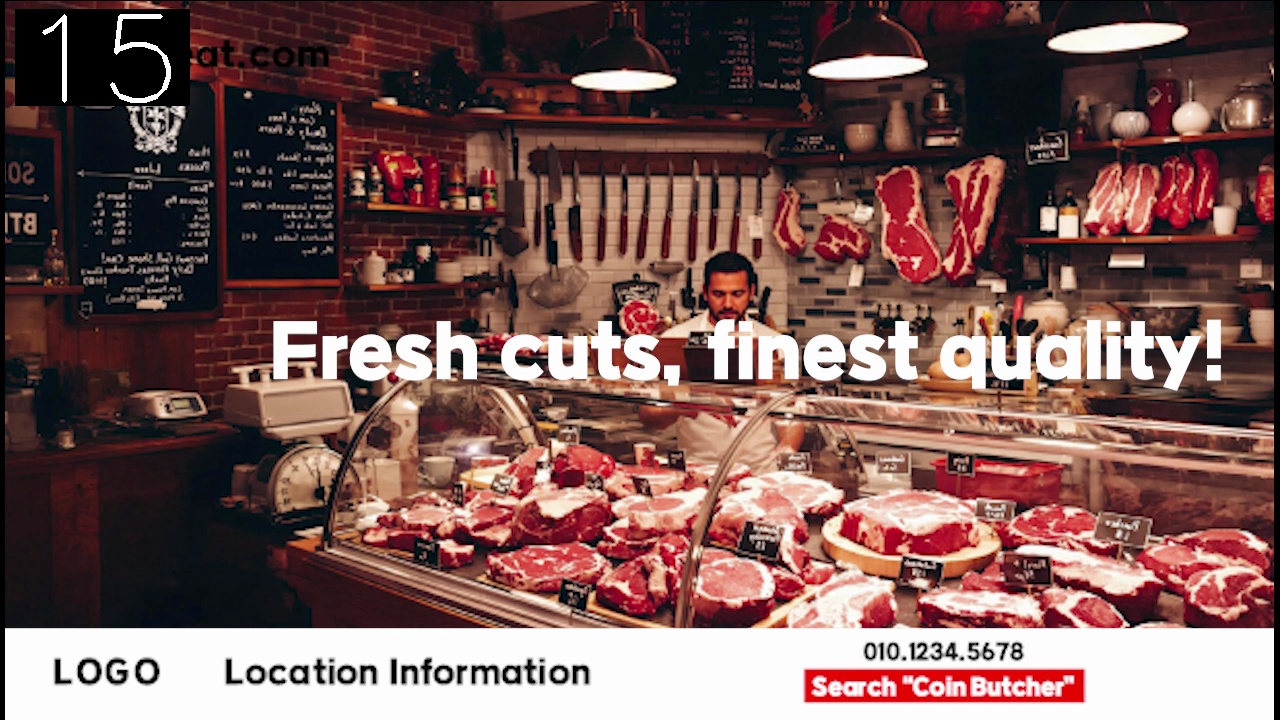} &
            \includegraphics[width=0.2\linewidth]{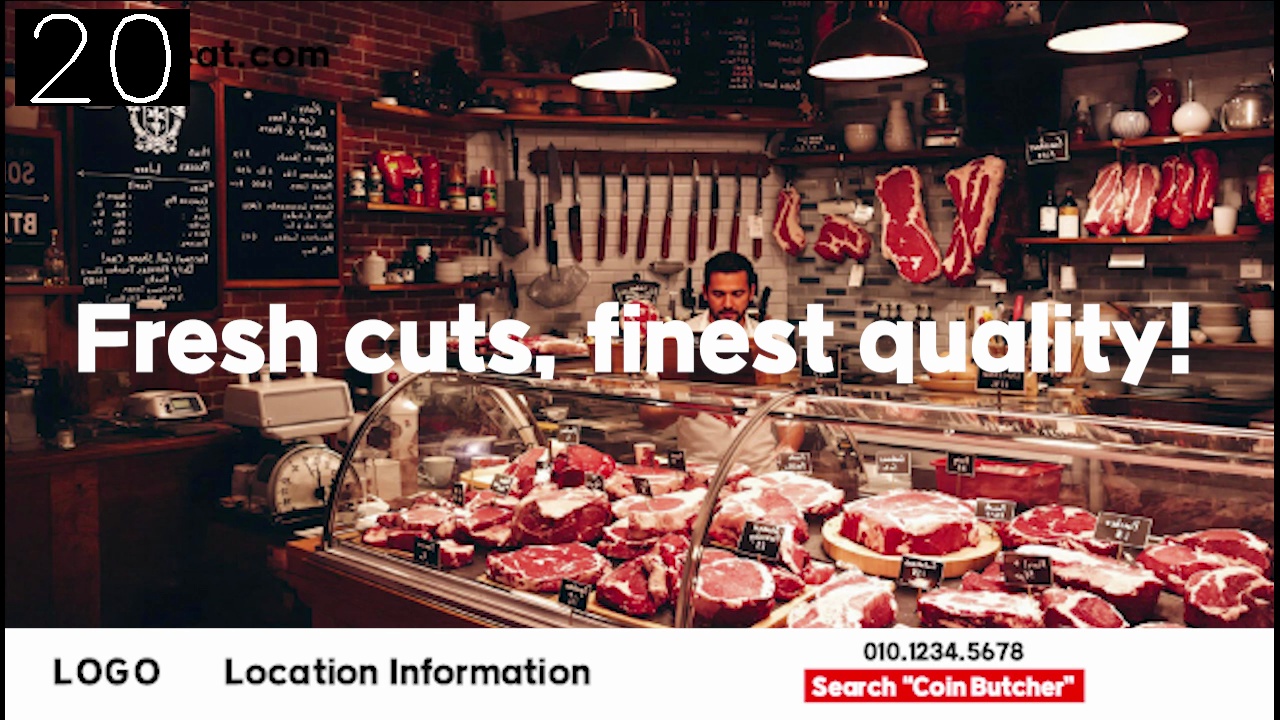} \\
            \multicolumn{5}{c}{\parbox{\linewidth}{\small (b) \textsf{The bottom banner contains a logo and three text elements: ``Location Information," ``Search ``Coin Butcher"" and ``010.1234.5678." The top-left banner includes a single text stating, ``www.meat.com." There is no top-right banner.
The video advertisement presents a scene containing text with the content ``Fresh cuts, finest quality!".}\vspace{3mm}}}\\
            \includegraphics[width=0.2\linewidth]{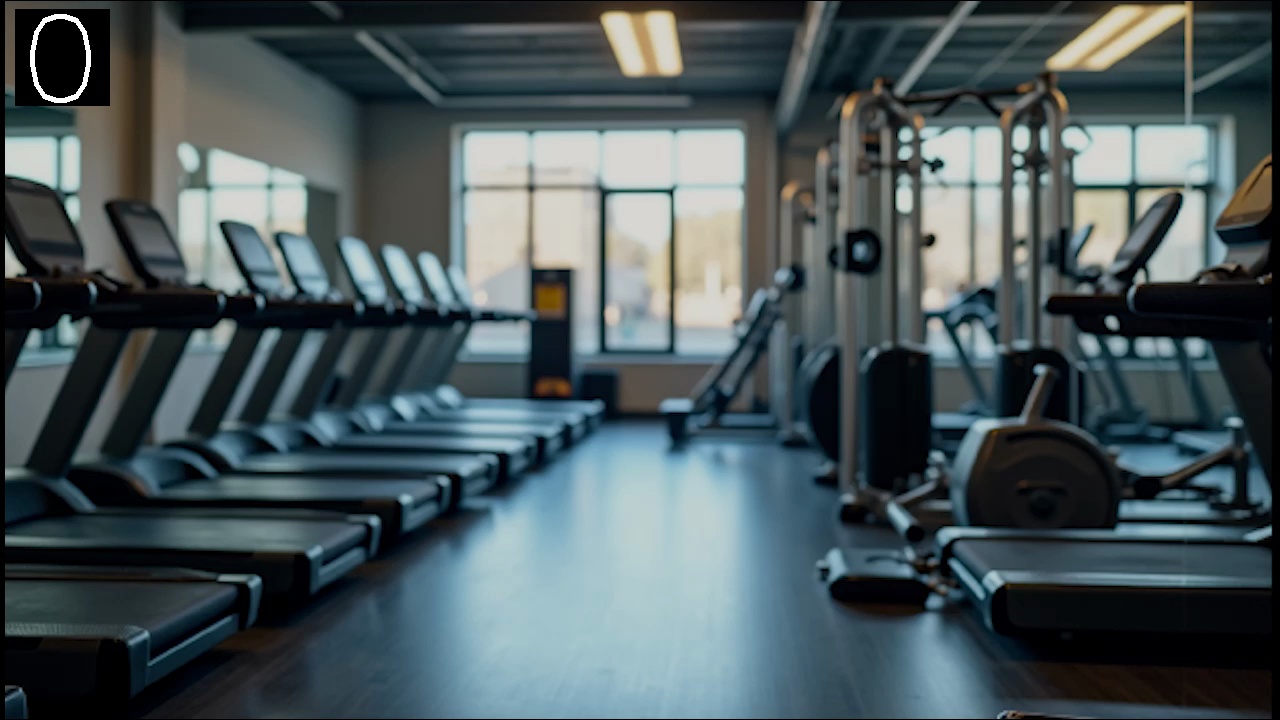} &
            \includegraphics[width=0.2\linewidth]{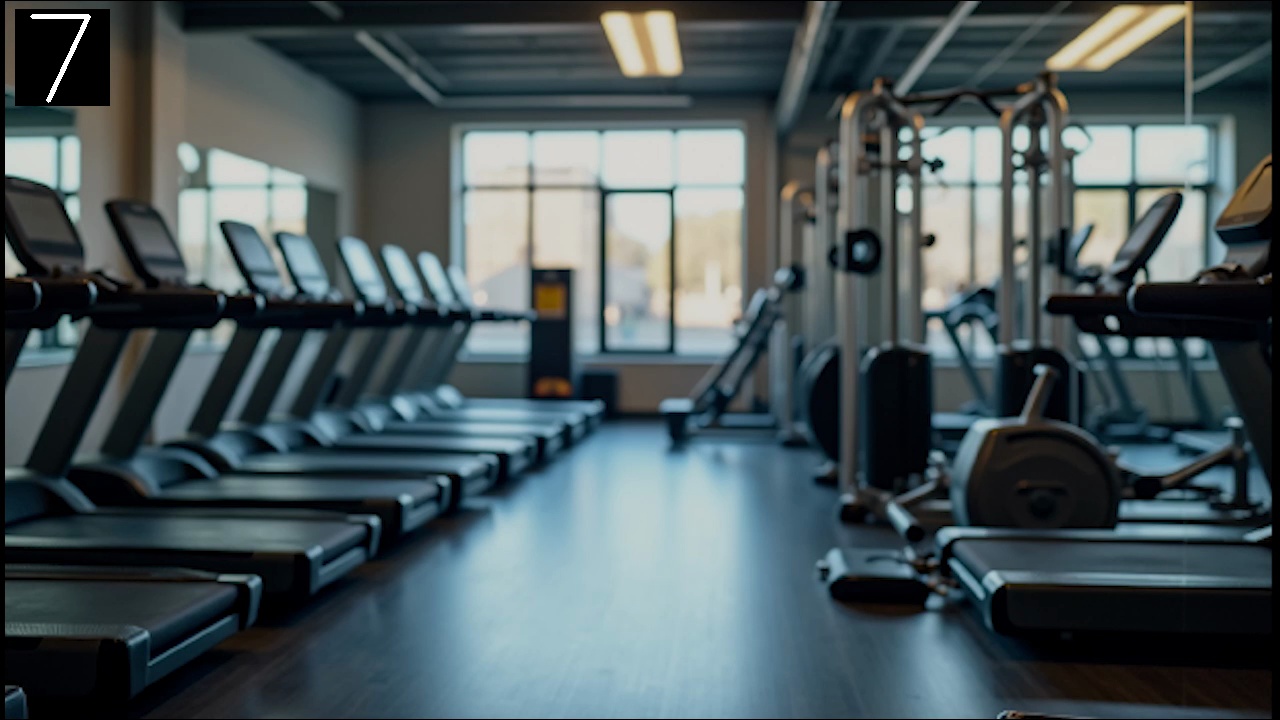} &
            \includegraphics[width=0.2\linewidth]{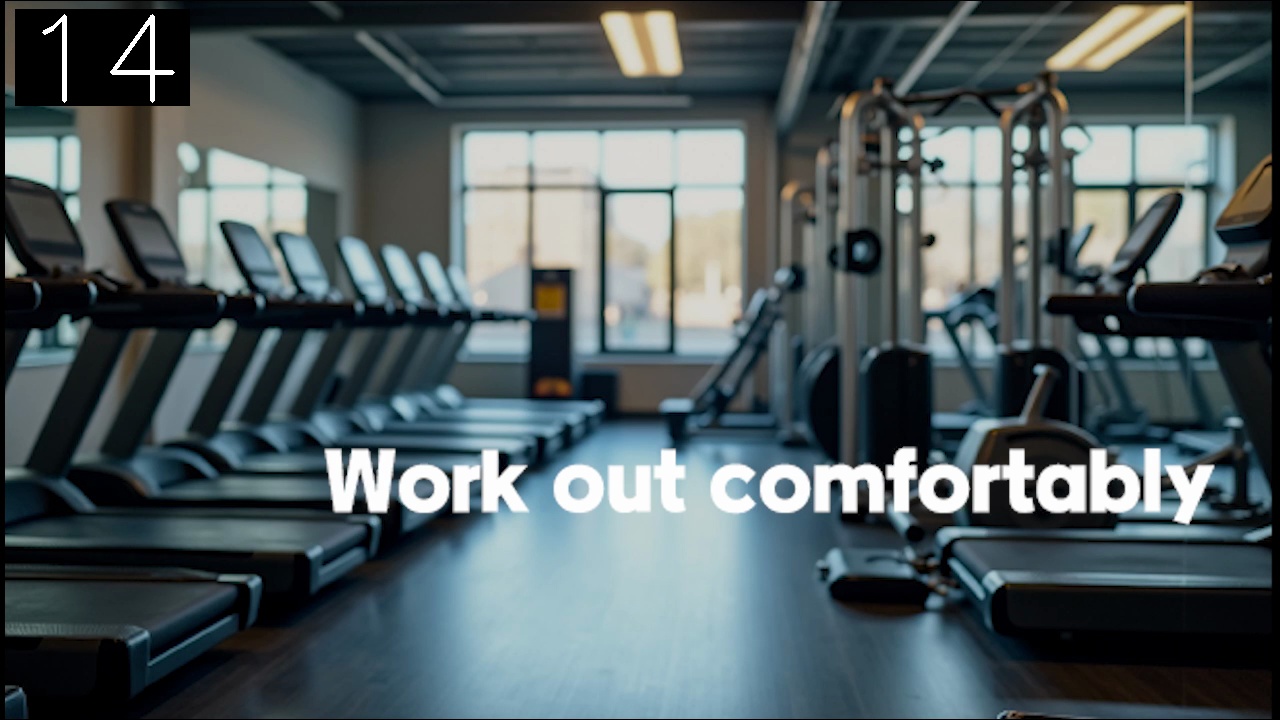} &
            \includegraphics[width=0.2\linewidth]{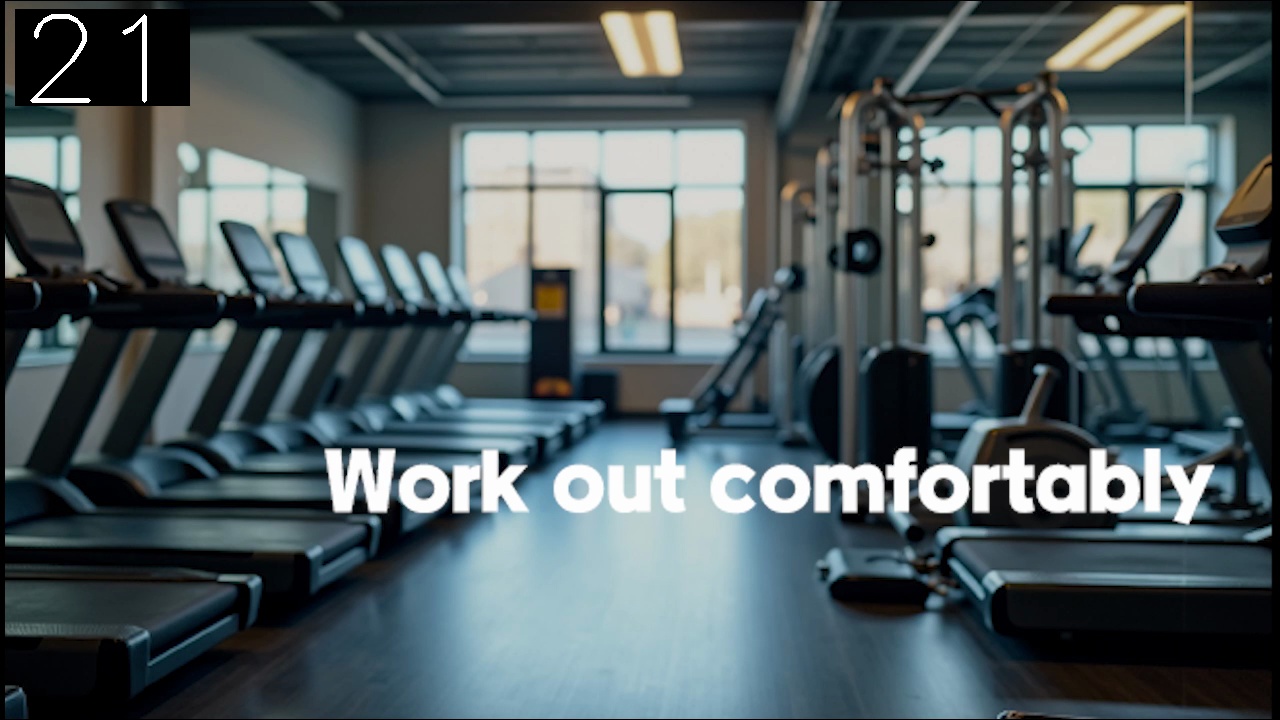} &
            \includegraphics[width=0.2\linewidth]{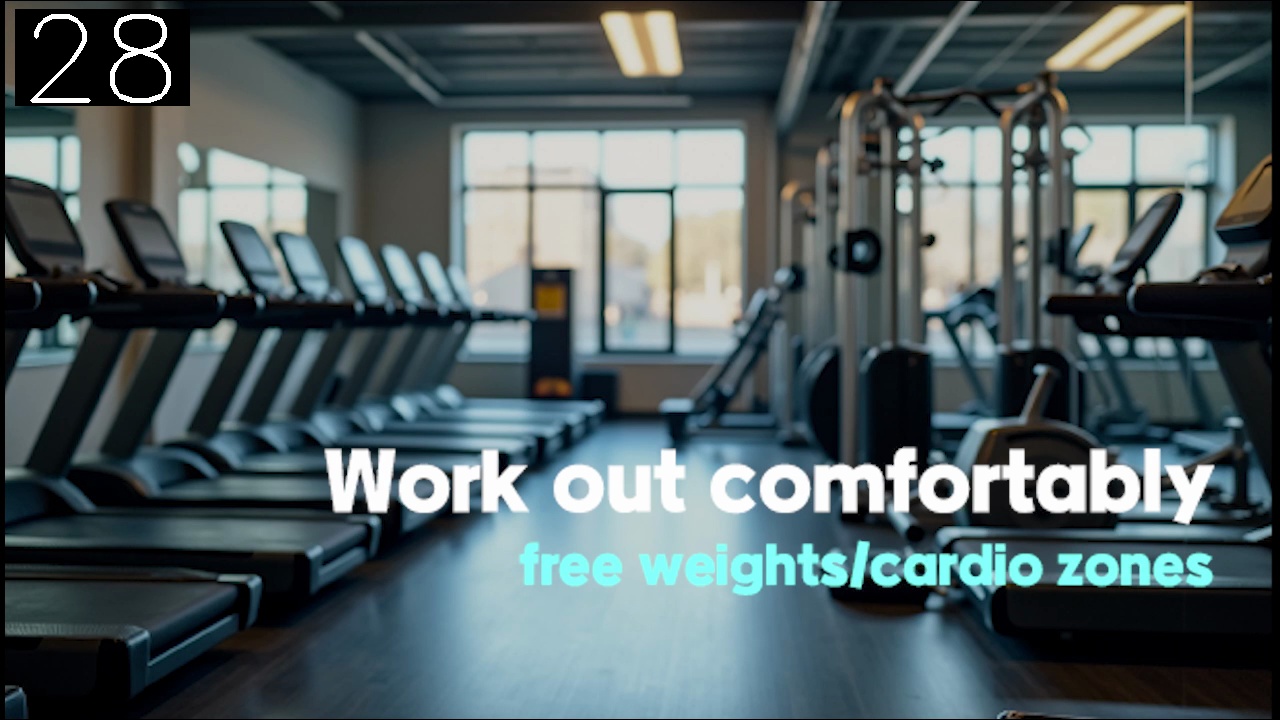} \\
            \multicolumn{5}{c}{\parbox{\linewidth}{\small (c) \textsf{There are no banners.
The scene depicts an image of gym equipment. Accompanying this image are two pieces of text, one stating ``Work out comfortably" and the other mentioning ``free weights/cardio zones."}\vspace{3mm}}}\\
            \includegraphics[width=0.2\linewidth]{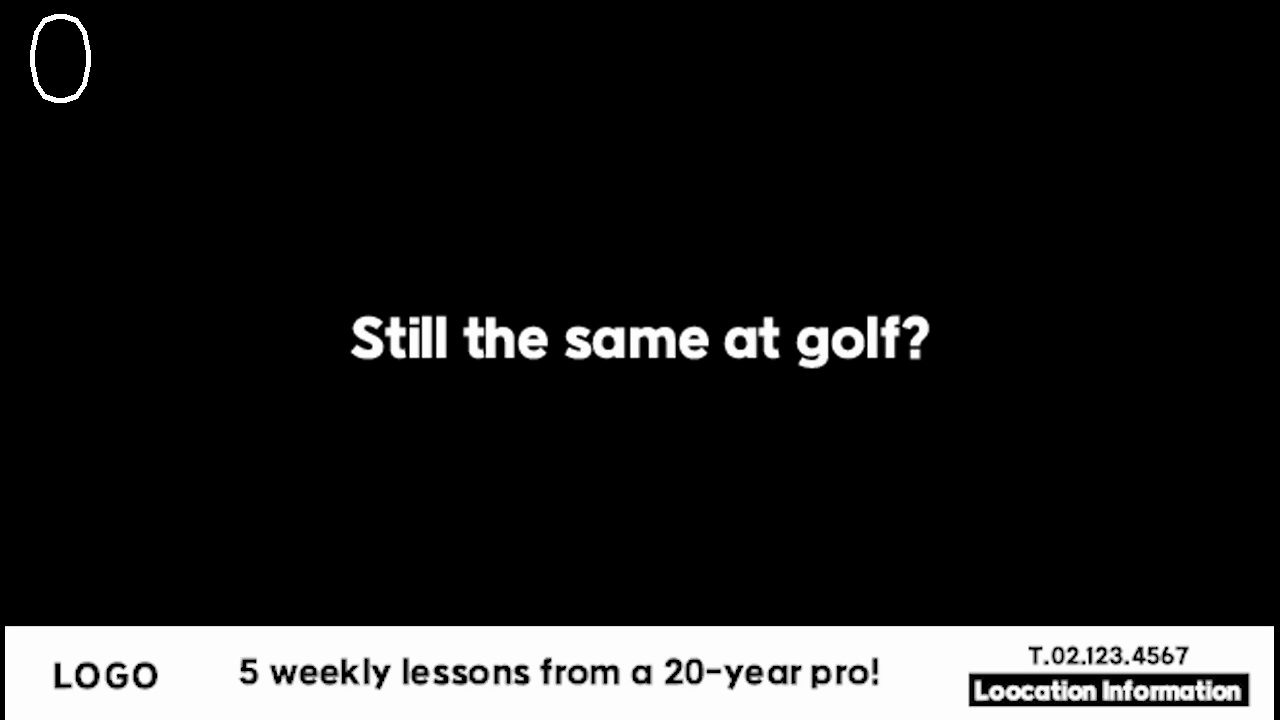} &
            \includegraphics[width=0.2\linewidth]{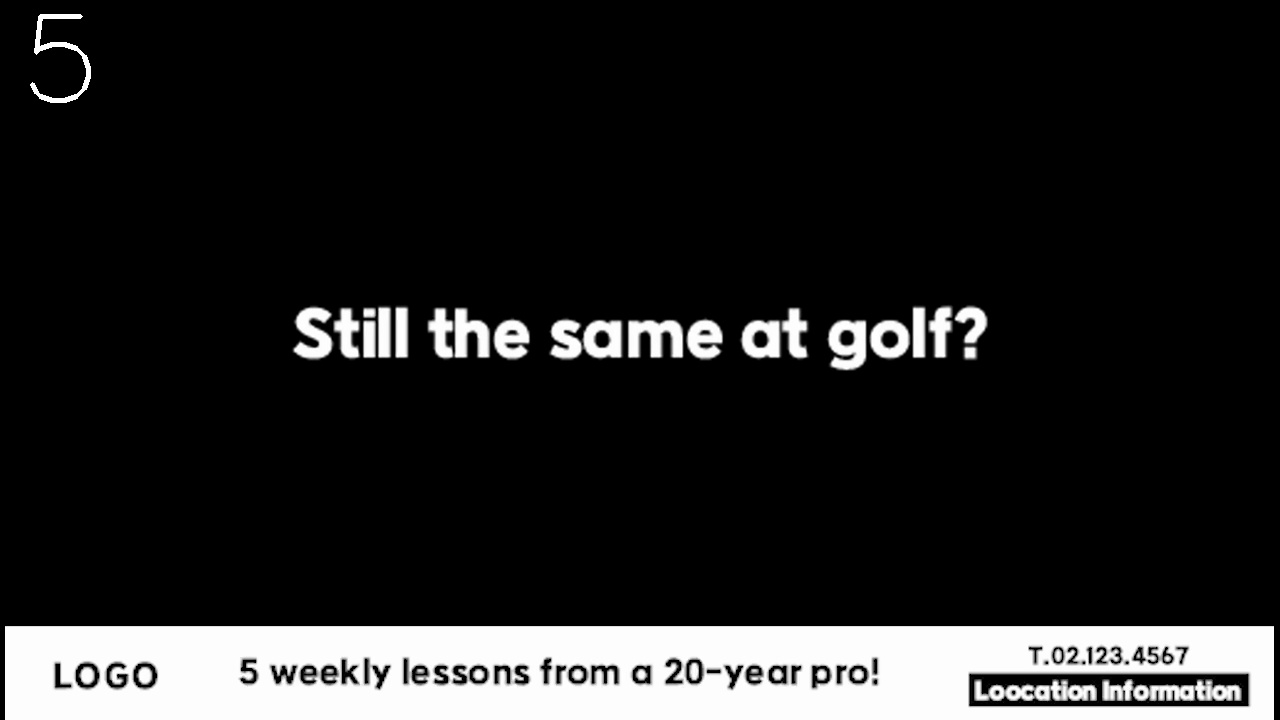} &
            \includegraphics[width=0.2\linewidth]{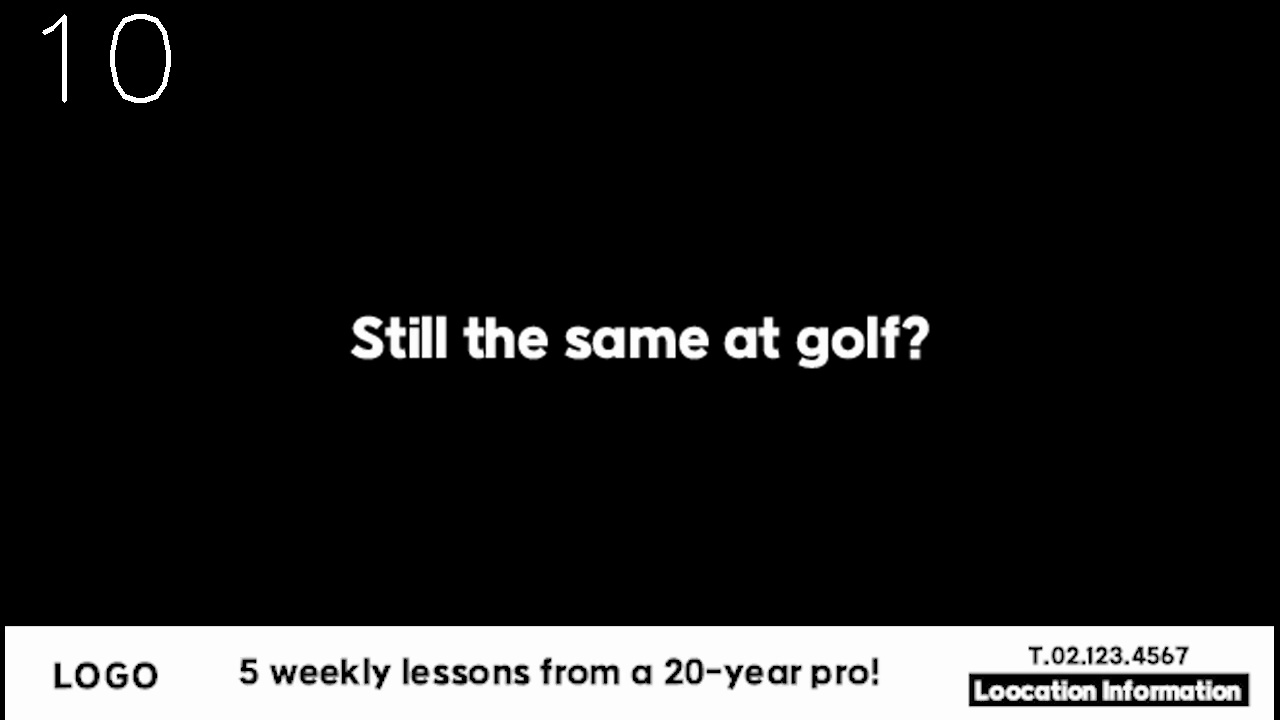} &
            \includegraphics[width=0.2\linewidth]{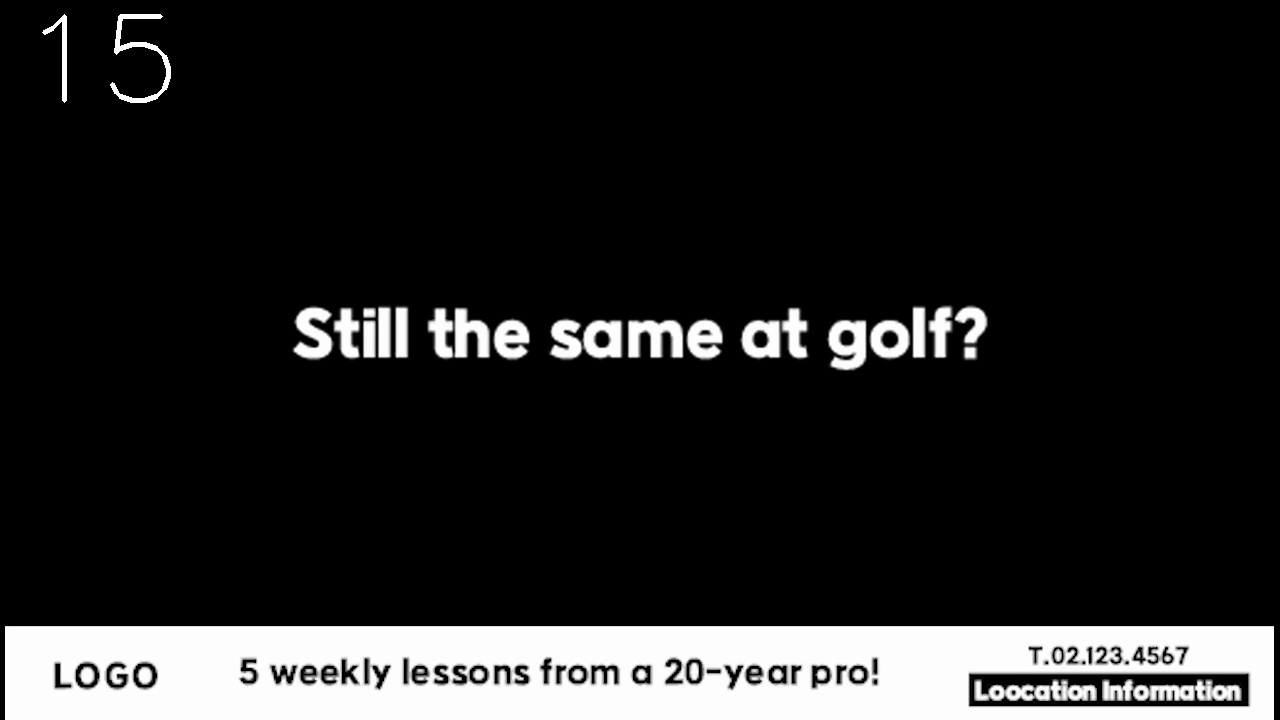} &
            \includegraphics[width=0.2\linewidth]{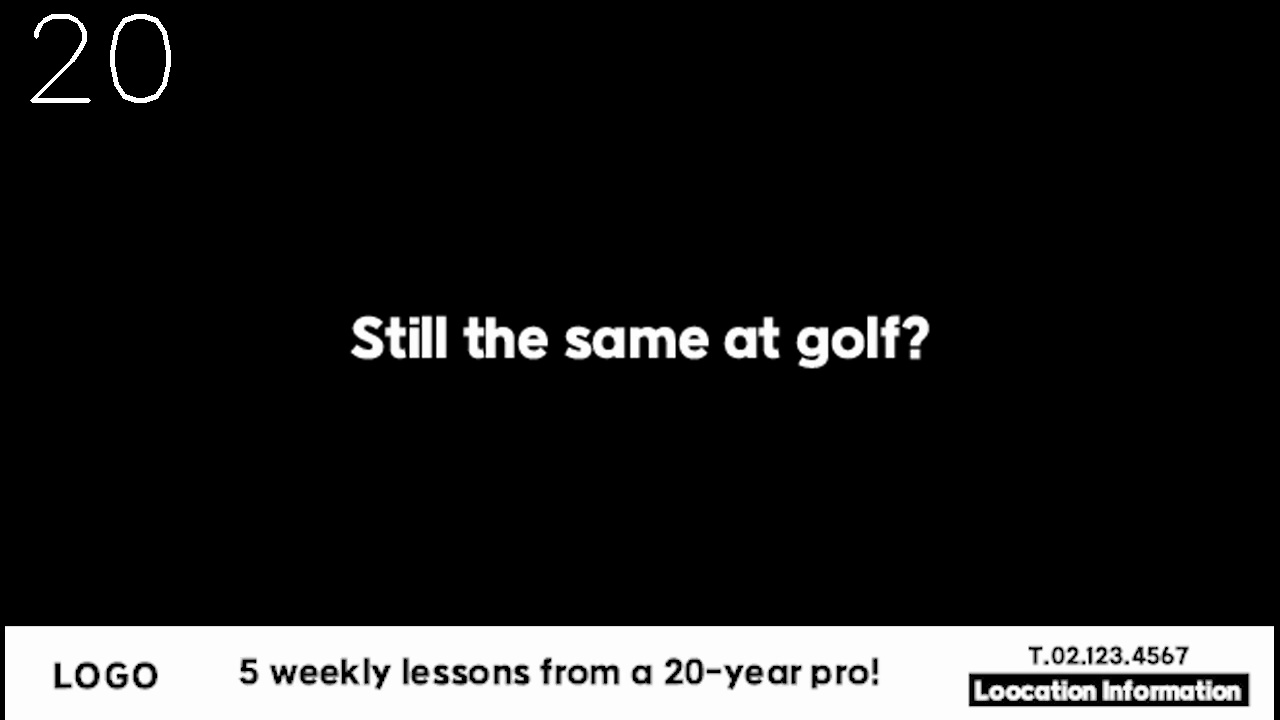} \\
            \multicolumn{5}{c}{\parbox{\linewidth}{\small (d) \textsf{The bottom banner contains a logo and three text elements. The texts include ``5 weekly lessons from a 20-year pro!", ``Location Information" and ``T.02.123.4567." There are no banners on the top-left or top-right of the screen.
The scene features a single piece of text asking, ``Still the same at golf?".}\vspace{3mm}}}\\
            \includegraphics[width=0.2\linewidth]{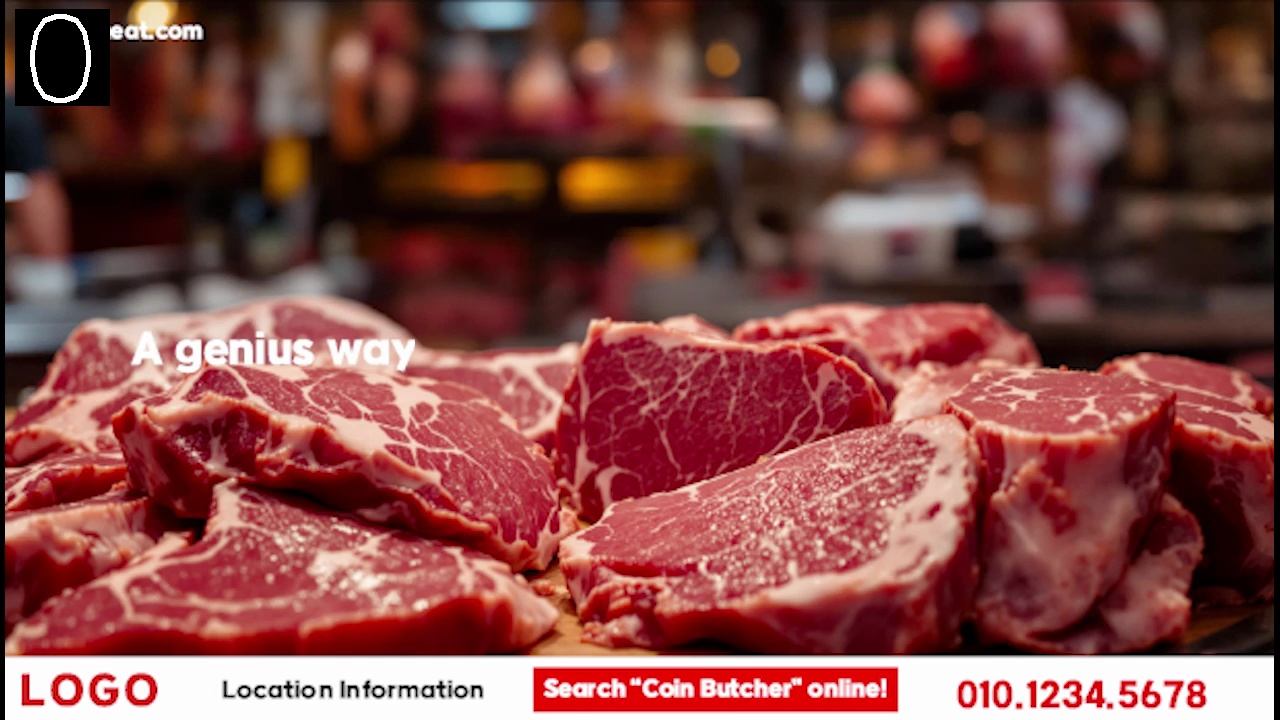} &
            \includegraphics[width=0.2\linewidth]{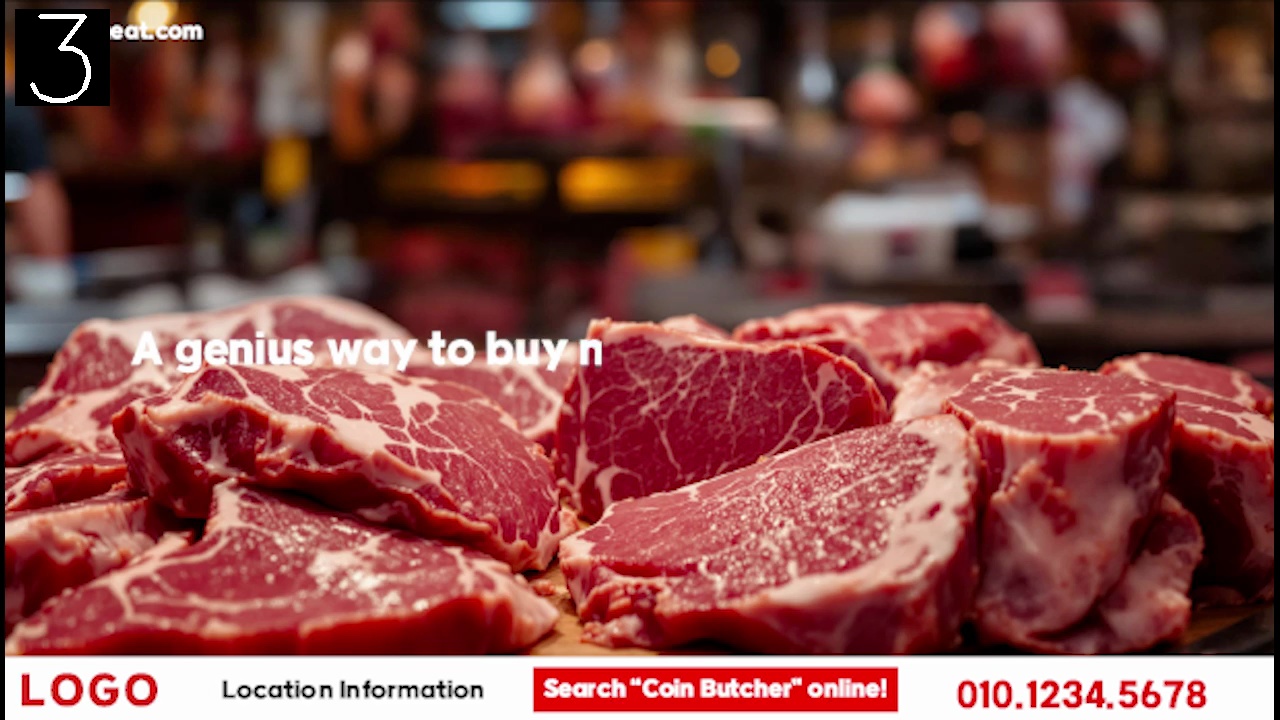} &
            \includegraphics[width=0.2\linewidth]{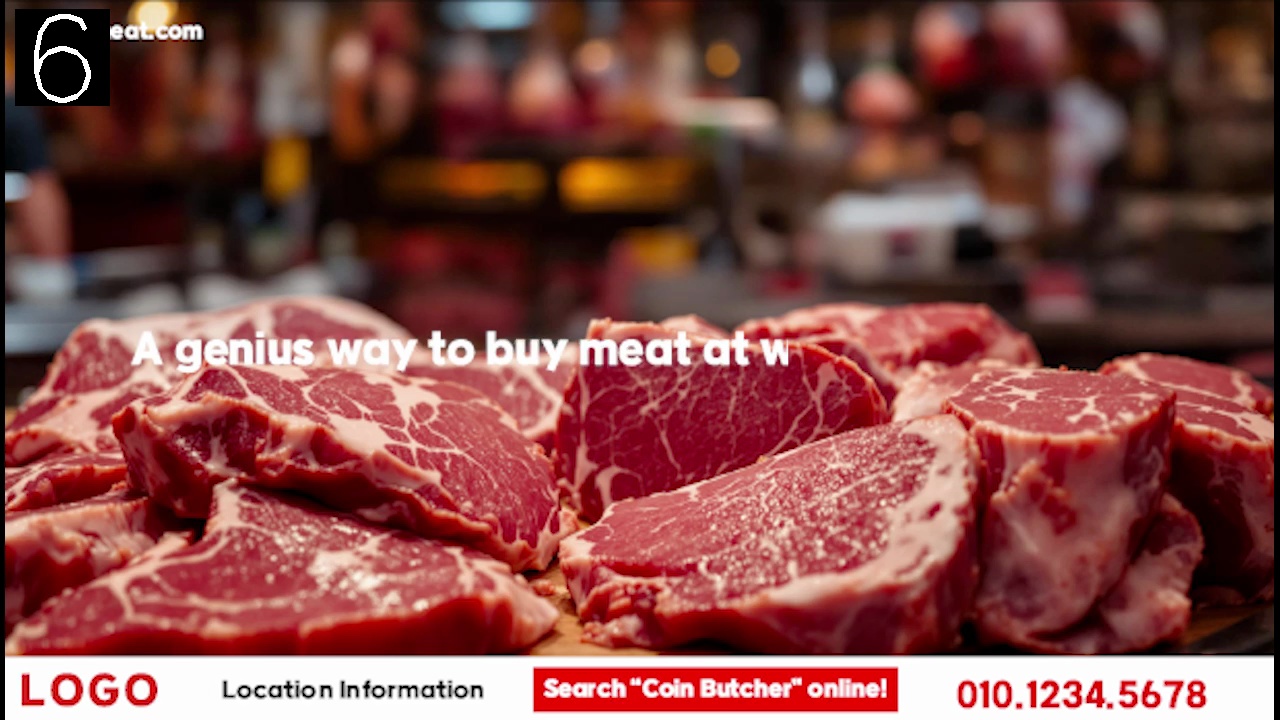} &
            \includegraphics[width=0.2\linewidth]{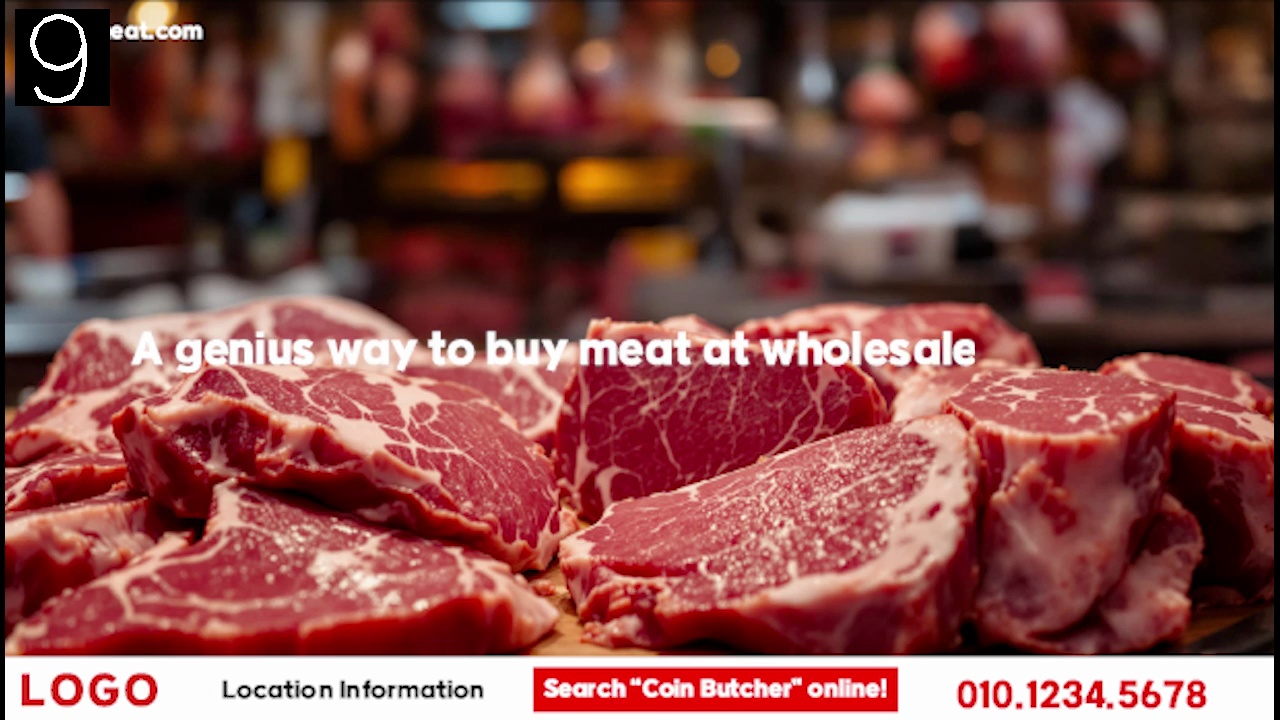} &
            \includegraphics[width=0.2\linewidth]{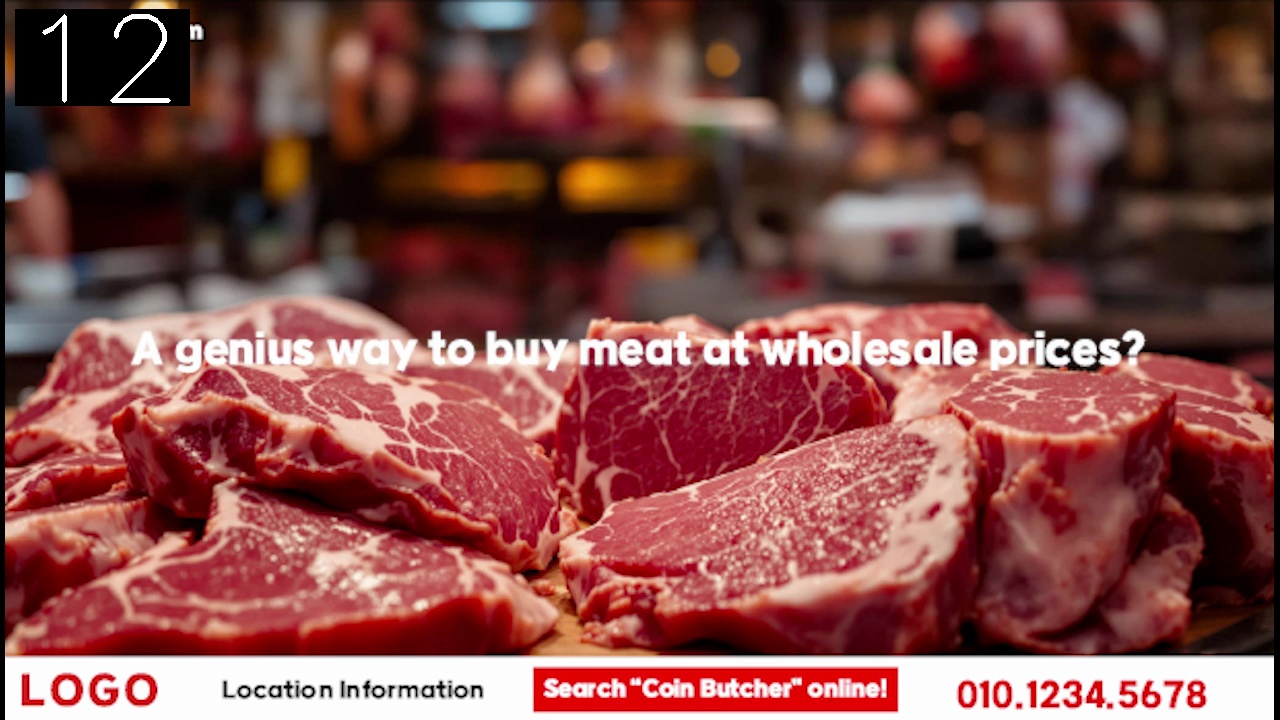} \\
            \multicolumn{5}{c}{\parbox{\linewidth}{\small (e) \textsf{The bottom banner contains a logo and three text elements: ``Location Information" ``Search ``Coin Butcher" online!" and ``010.1234.5678." The top-left banner includes a single text stating, ``www.meat.com" There is no top-right banner.
The scene features a text overlay with the message, ``A genius way to buy meat at wholesale prices?". The text appears over an image of raw meat slices.}\vspace{3mm}}}\\
        \end{tabular}
    }
        \caption{
            Videos generated by VAKER.
            The number on the top-left corner of each frame indicates the frame index.
        } \label{fig:qual_video}
\end{figure*}

\begin{figure*}[h]
    \centering
    \renewcommand{\arraystretch}{1.5}
    \scalebox{0.9}{
    \setlength{\tabcolsep}{1pt} \hspace{-0.5mm}
        \hspace{-3mm}
        \begin{tabular}{ccccc}
            \includegraphics[width=0.2\linewidth]{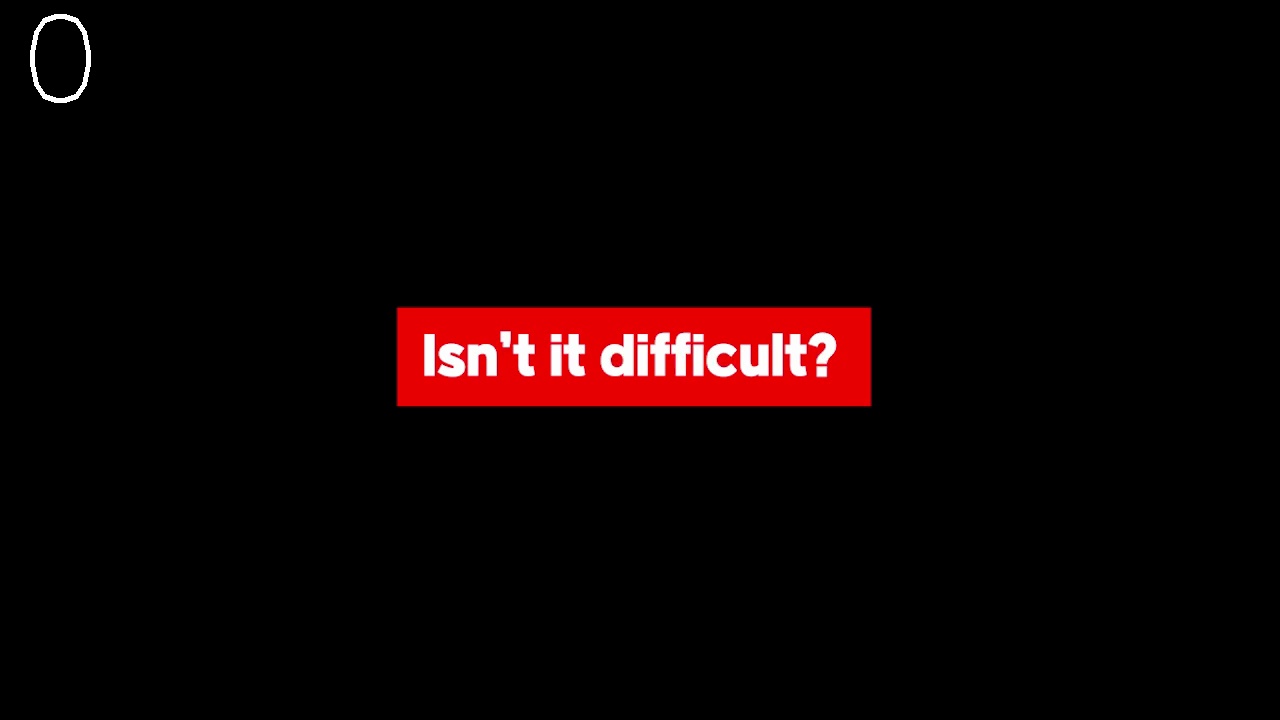} &
            \includegraphics[width=0.2\linewidth]{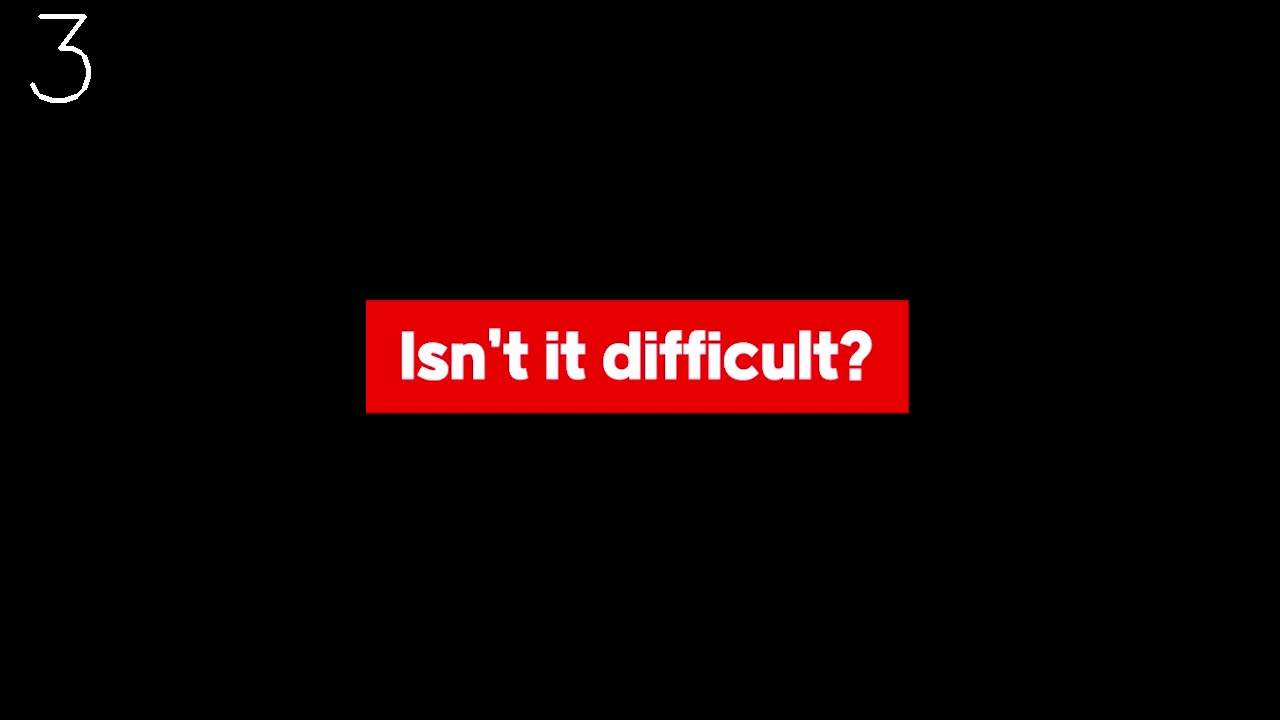} &
            \includegraphics[width=0.2\linewidth]{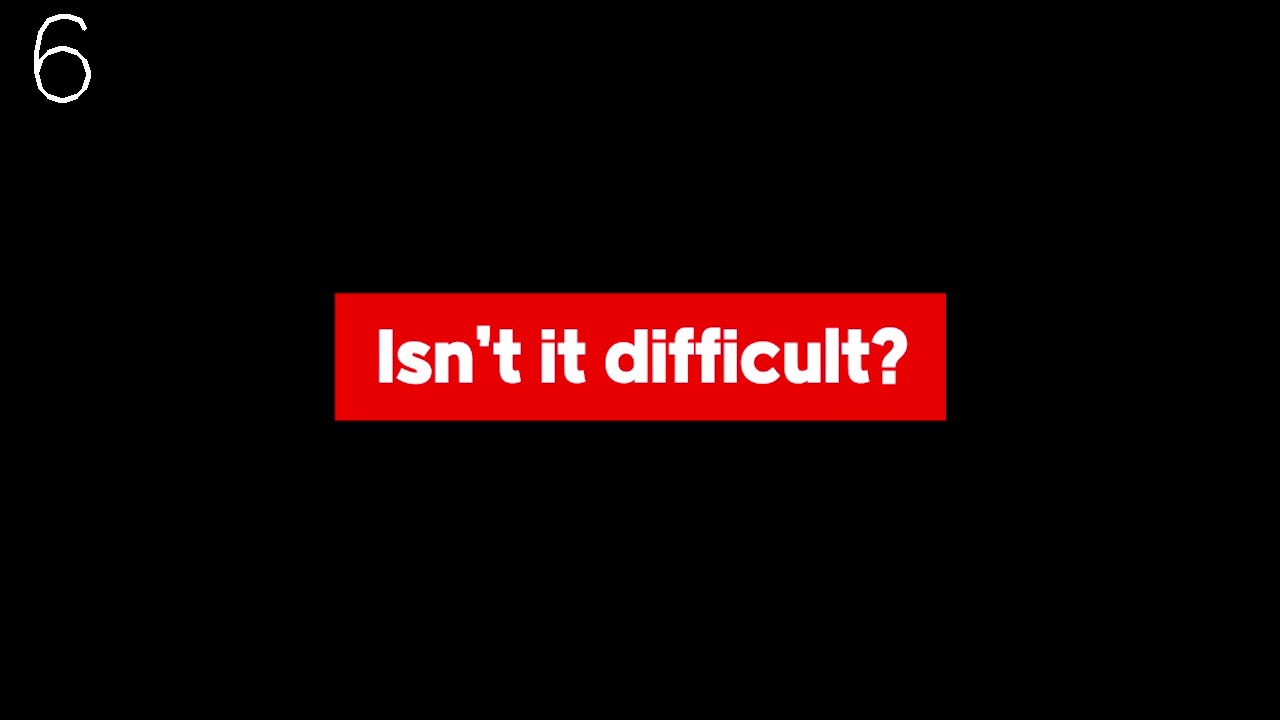} &
            \includegraphics[width=0.2\linewidth]{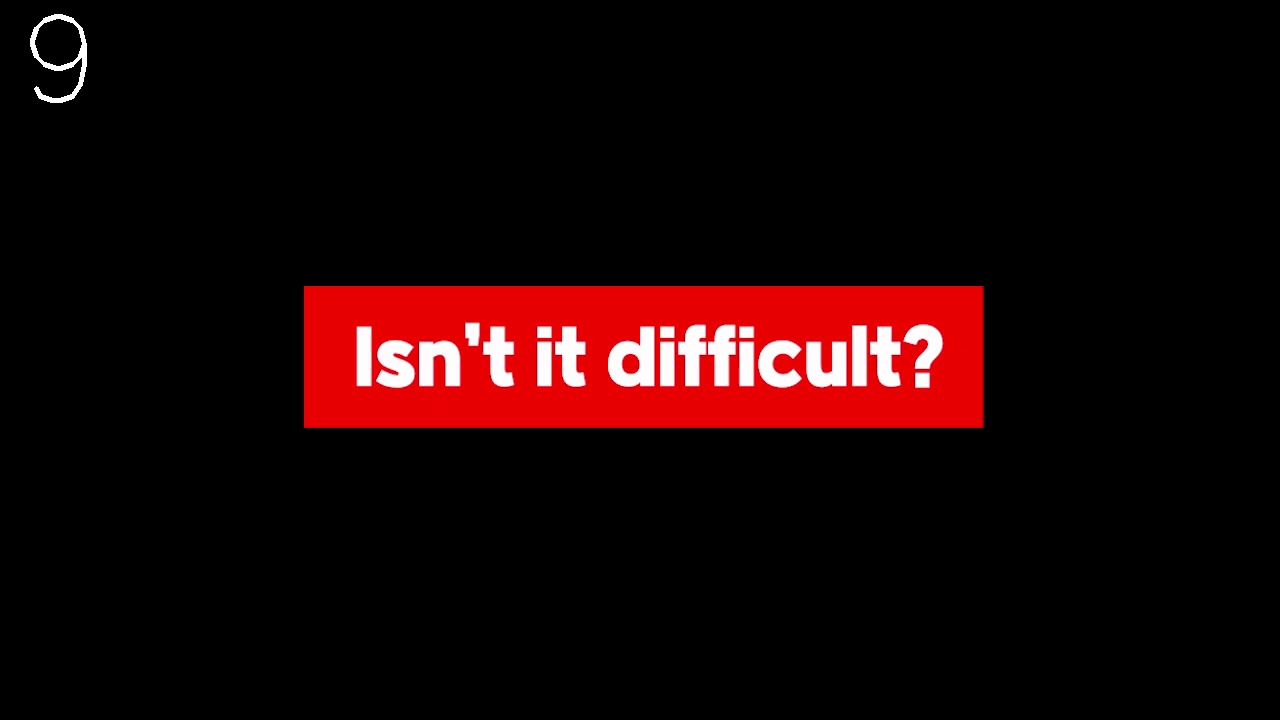} &
            \includegraphics[width=0.2\linewidth]{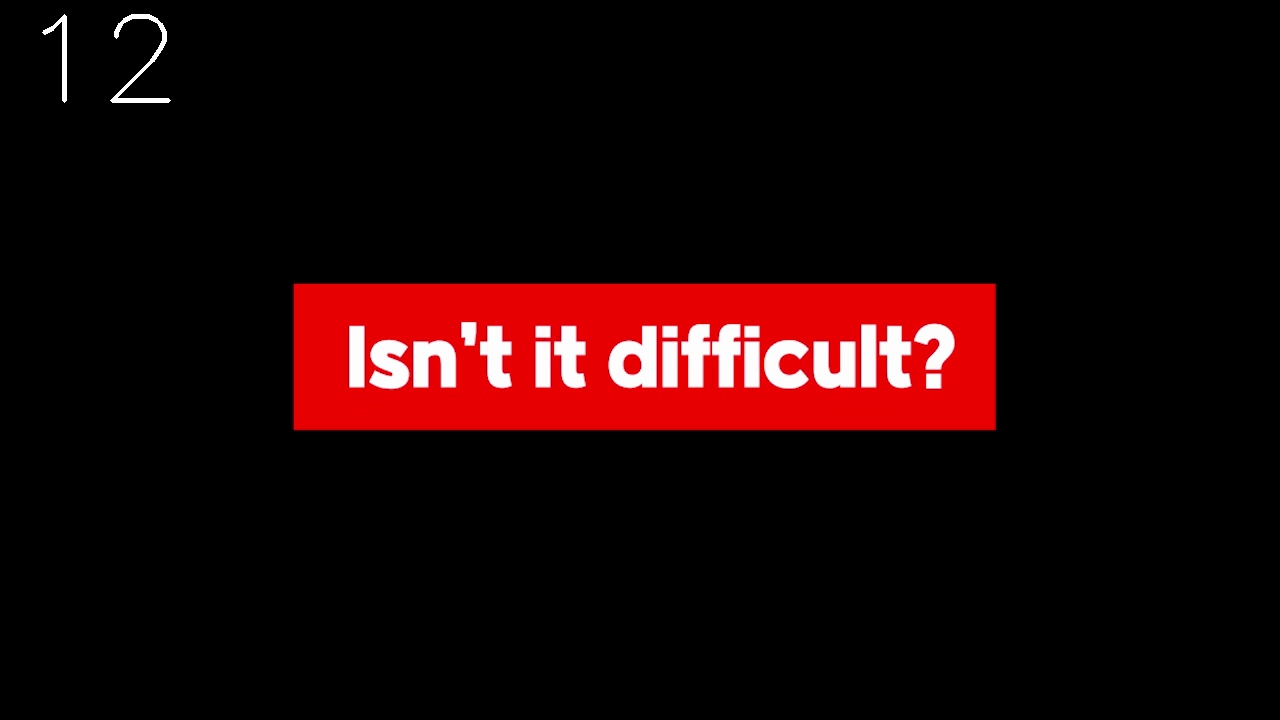} \\
            \multicolumn{5}{c}{\parbox{\linewidth}{\small (a) \textsf{There are no banners.
The scene features a piece of text with the content ``Isn't it difficult?", which is displayed prominently on the screen.}\vspace{3mm}}}\\
            \includegraphics[width=0.2\linewidth]{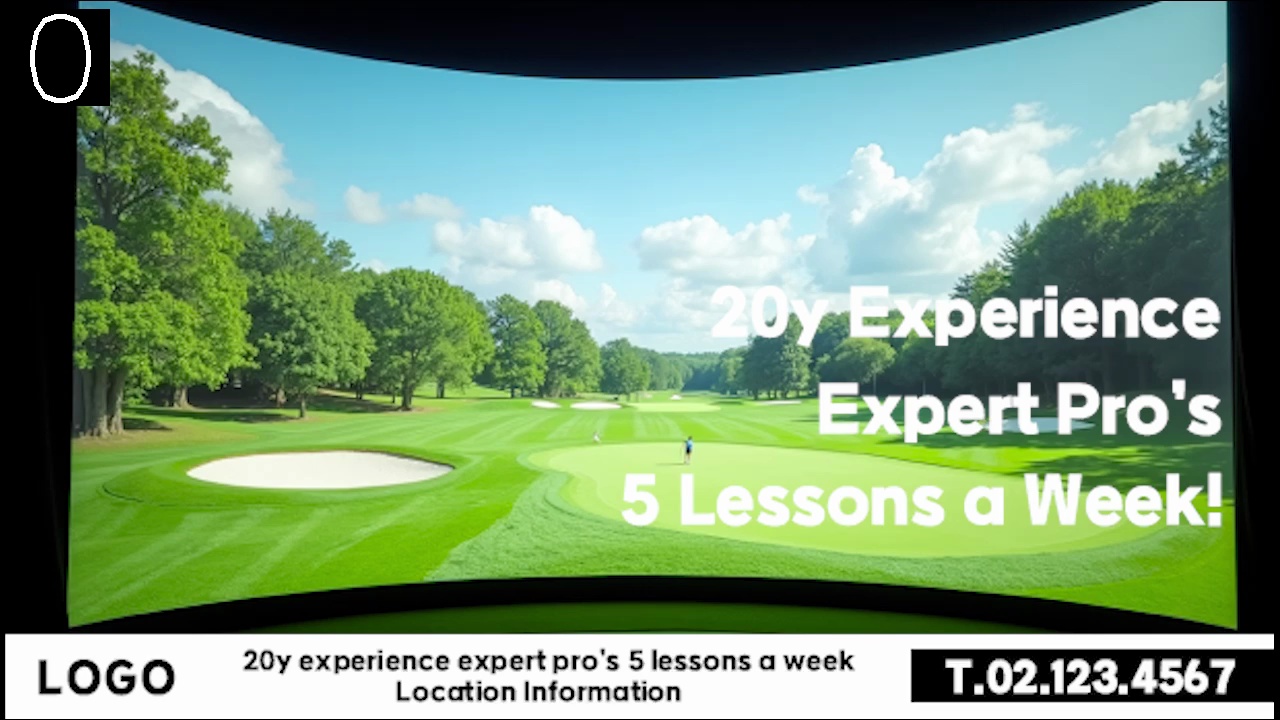} &
            \includegraphics[width=0.2\linewidth]{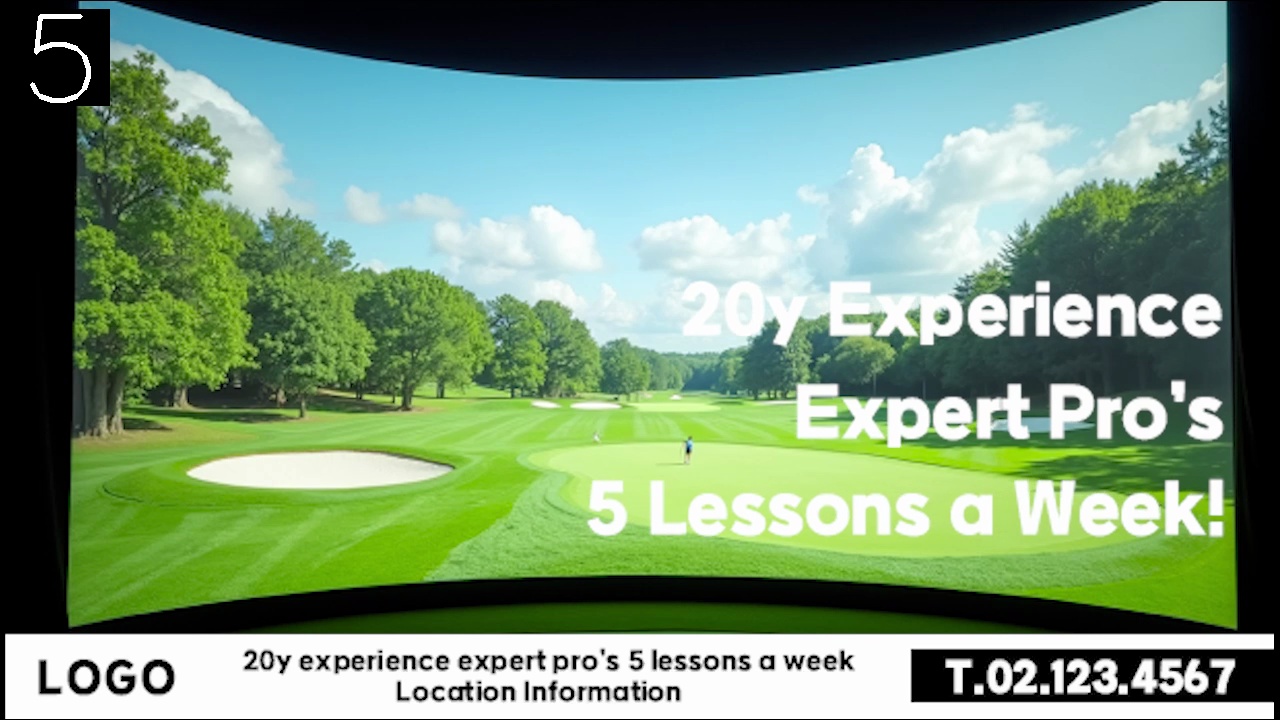} &
            \includegraphics[width=0.2\linewidth]{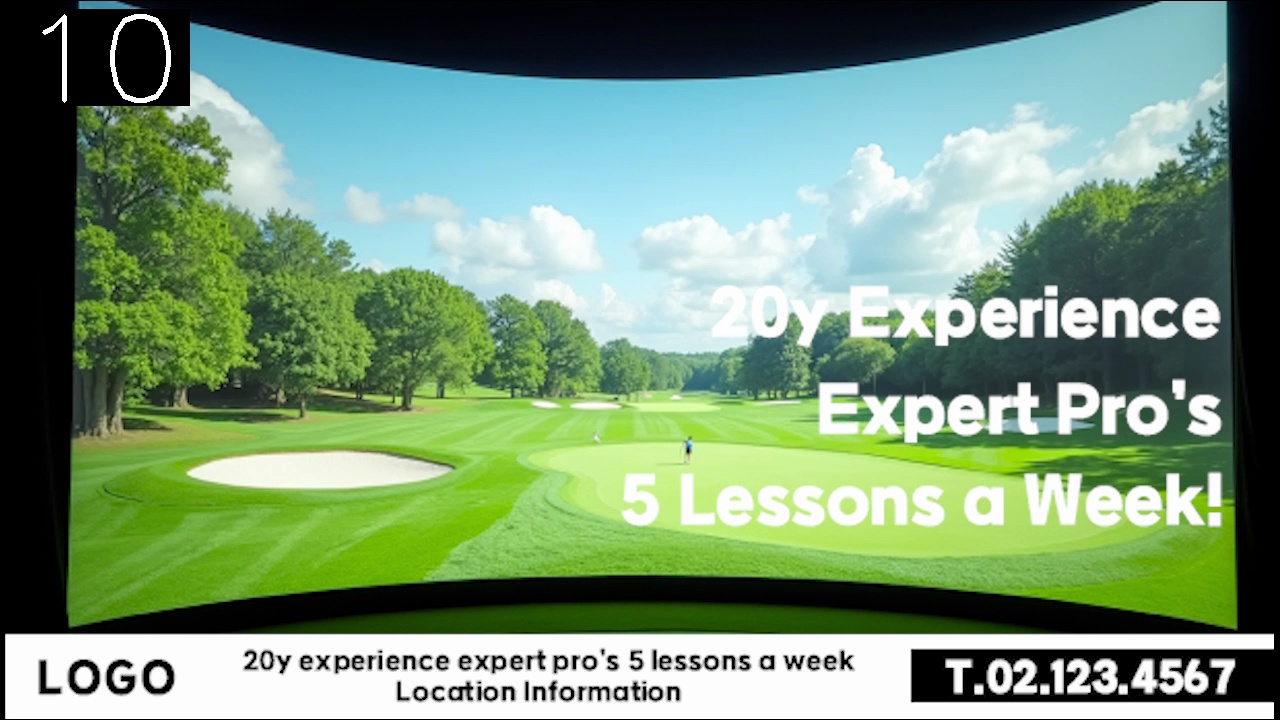} &
            \includegraphics[width=0.2\linewidth]{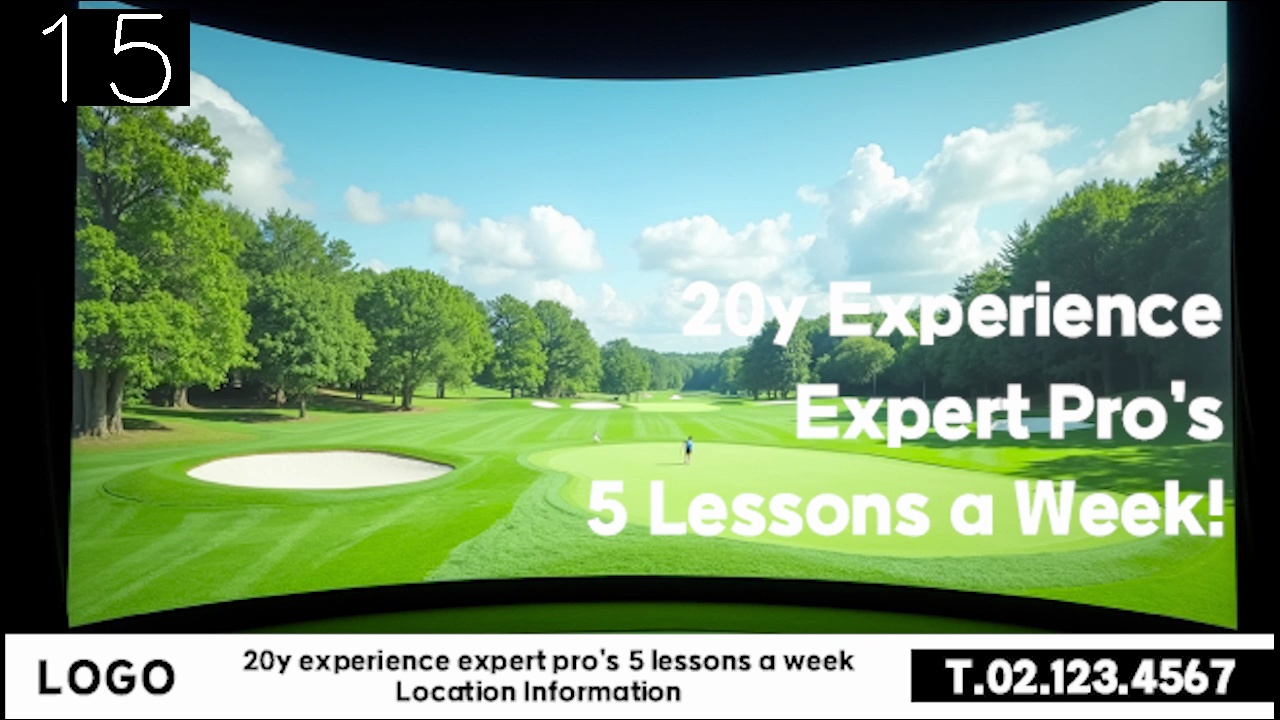} &
            \includegraphics[width=0.2\linewidth]{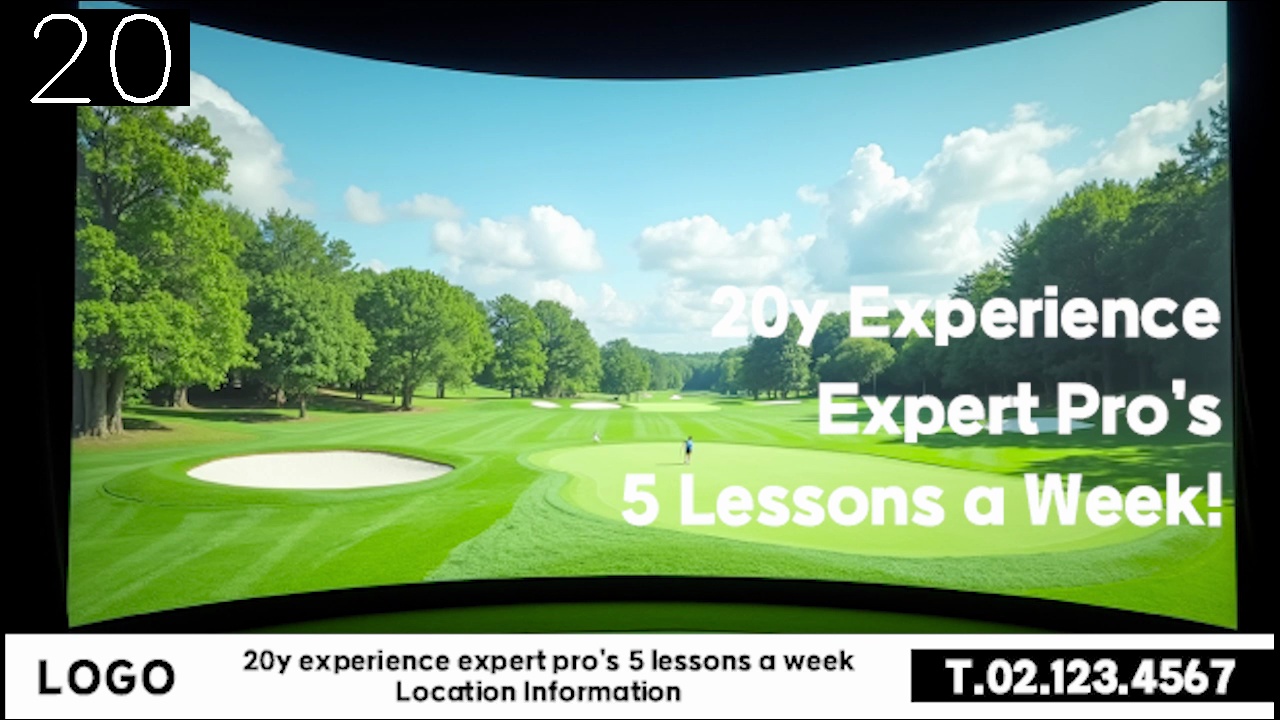} \\
            \multicolumn{5}{c}{\parbox{\linewidth}{\small (b) \textsf{The bottom banner contains a logo and three text elements. The texts include ``20y experience expert pro's 5 lessons a week," ``Location Information," and ``T.02.123.4567." There are no banners on the top-left or top-right of the screen.
The scene features an image of a golf course displayed on a large screen. Superimposed on this background is a text that states ``20y Experience," followed by ``Expert Pro's" signaling training by a certified instructor. Additionally, it highlights ``5 Lessons a Week!".}\vspace{3mm}}}\\
            \includegraphics[width=0.2\linewidth]{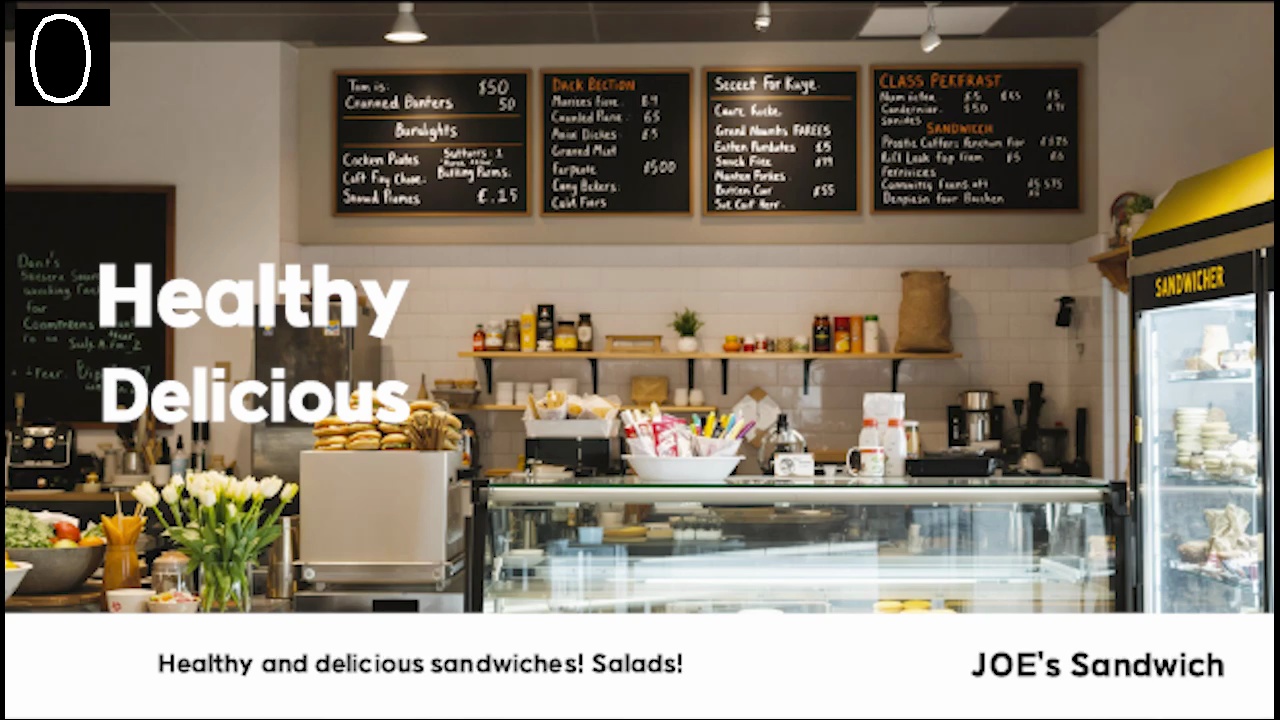} &
            \includegraphics[width=0.2\linewidth]{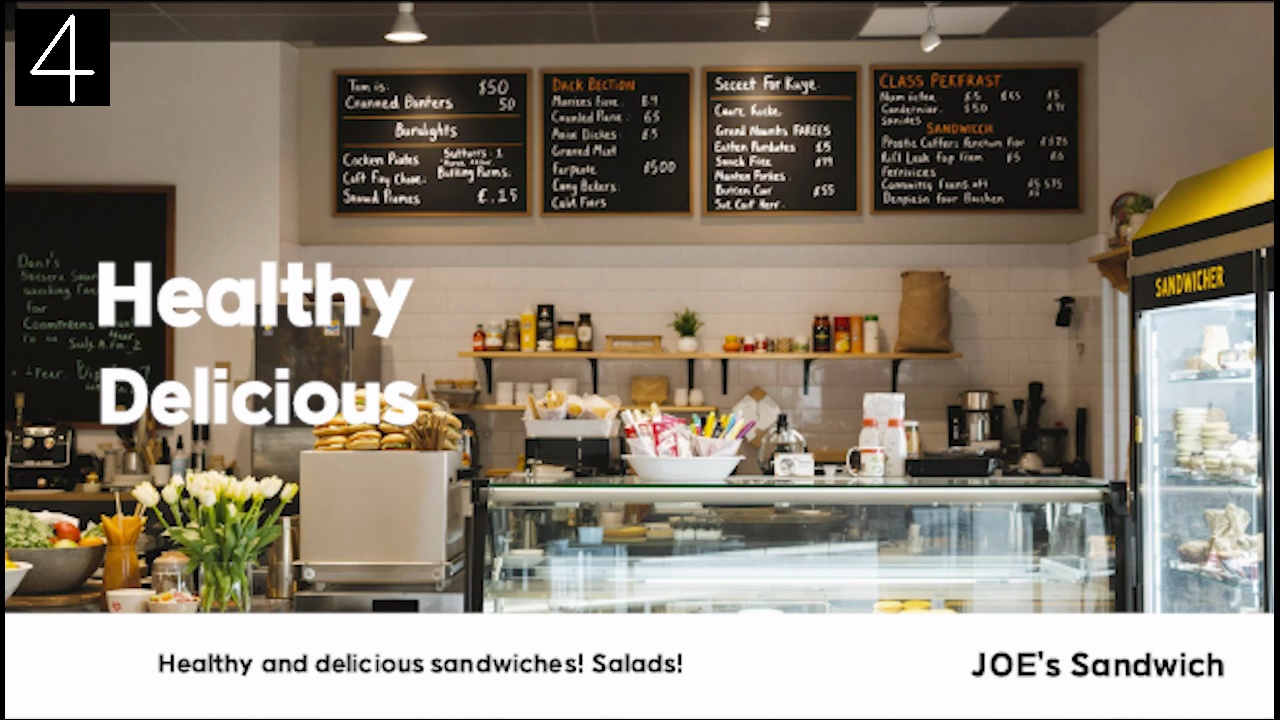} &
            \includegraphics[width=0.2\linewidth]{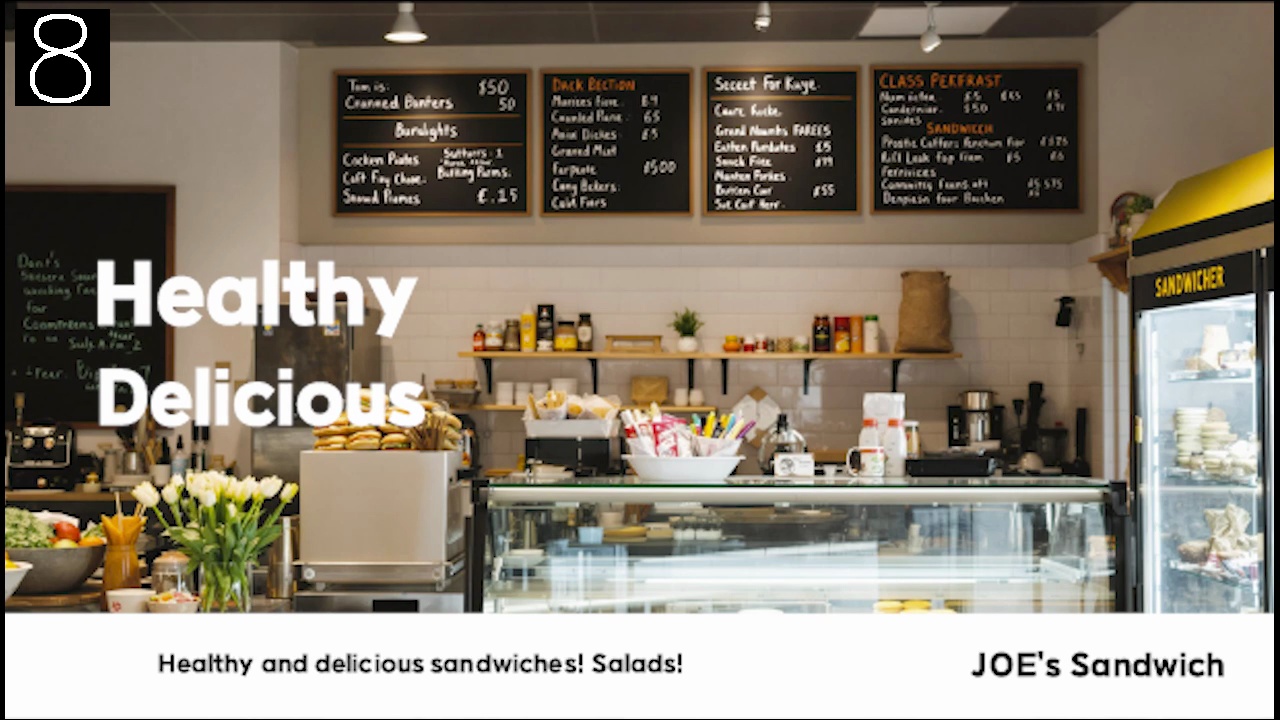} &
            \includegraphics[width=0.2\linewidth]{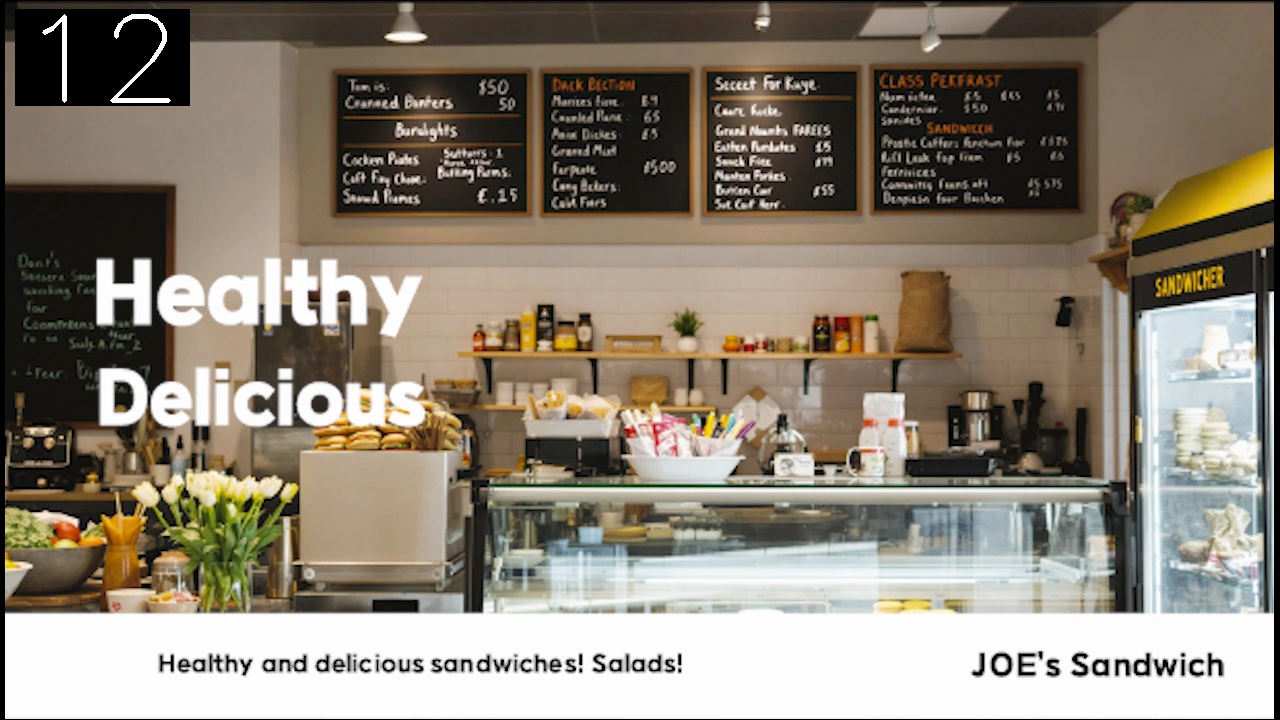} &
            \includegraphics[width=0.2\linewidth]{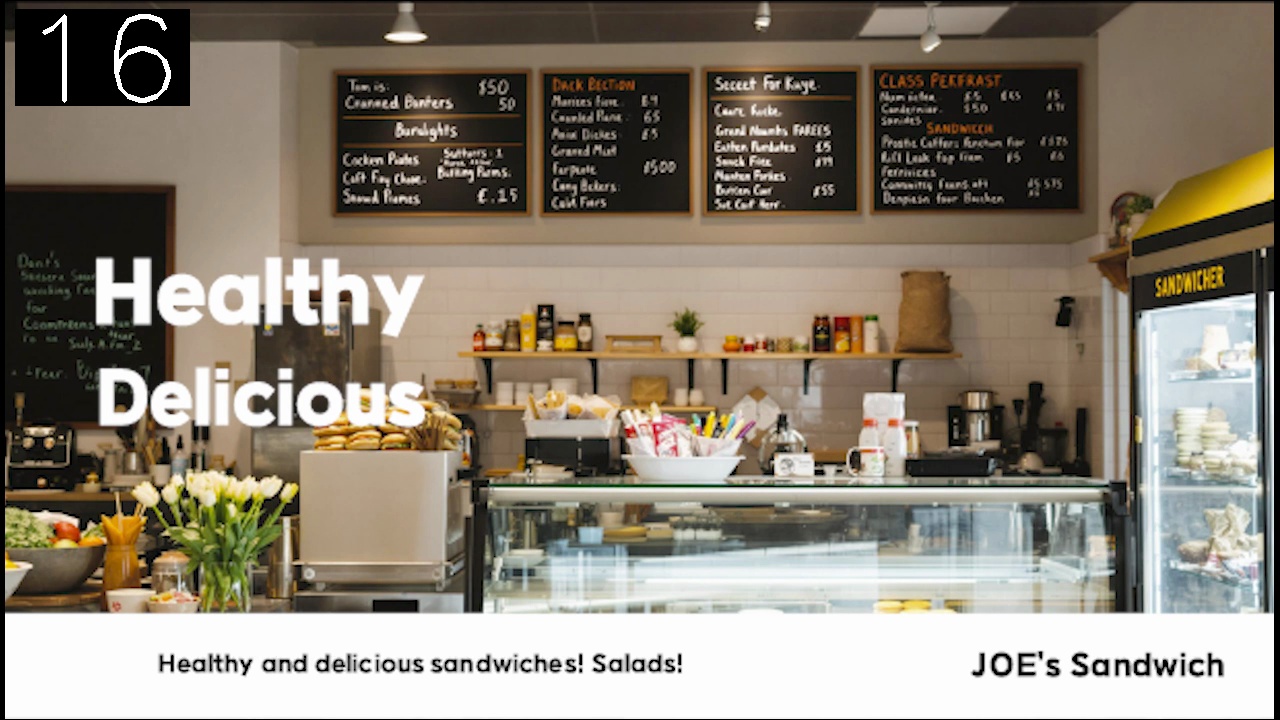} \\
            \multicolumn{5}{c}{\parbox{\linewidth}{\small (c) \textsf{The bottom banner contains two text objects. The first text object reads, ``Healthy and delicious sandwiches! Salads!" The second text object says, ``JOE's Sandwich." There are no banners at the top-left or top-right of the screen.
The video advertisement features an interior image of a sandwich shop as its background. It contains text elements with messages like ``Healthy" and ``Delicious," presented with distinctive typography.}\vspace{3mm}}}\\
            \includegraphics[width=0.2\linewidth]{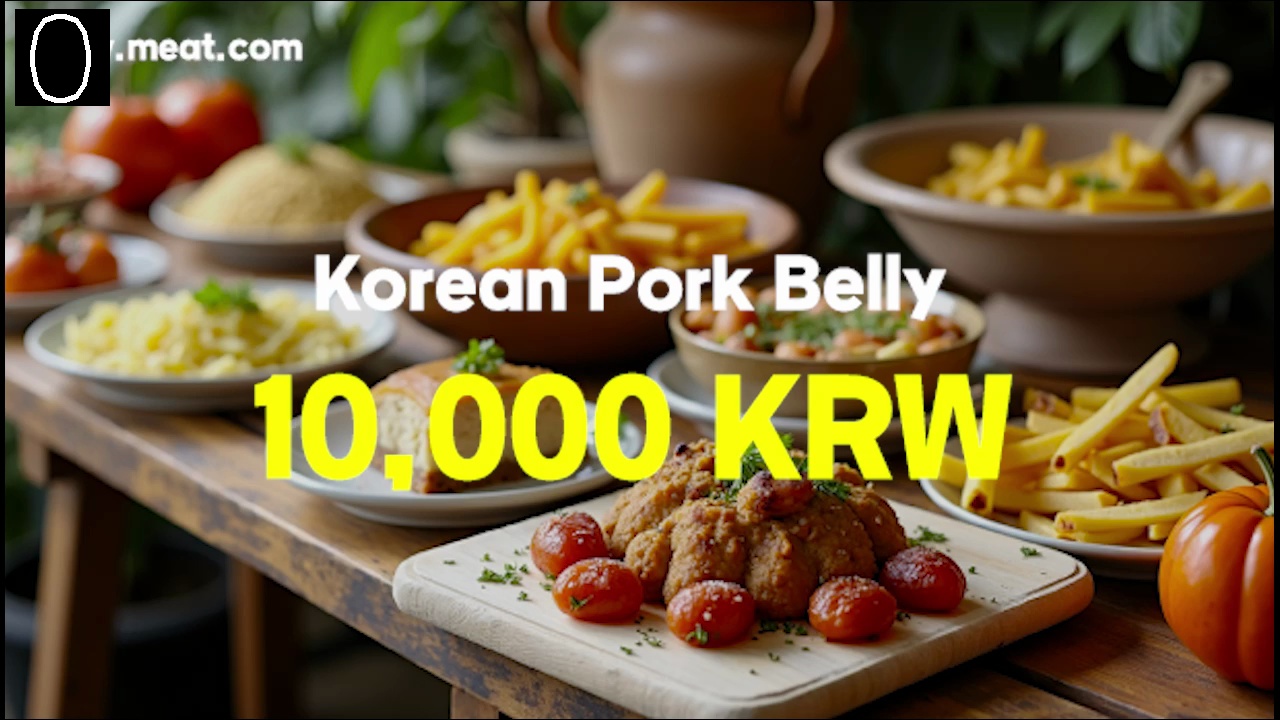} &
            \includegraphics[width=0.2\linewidth]{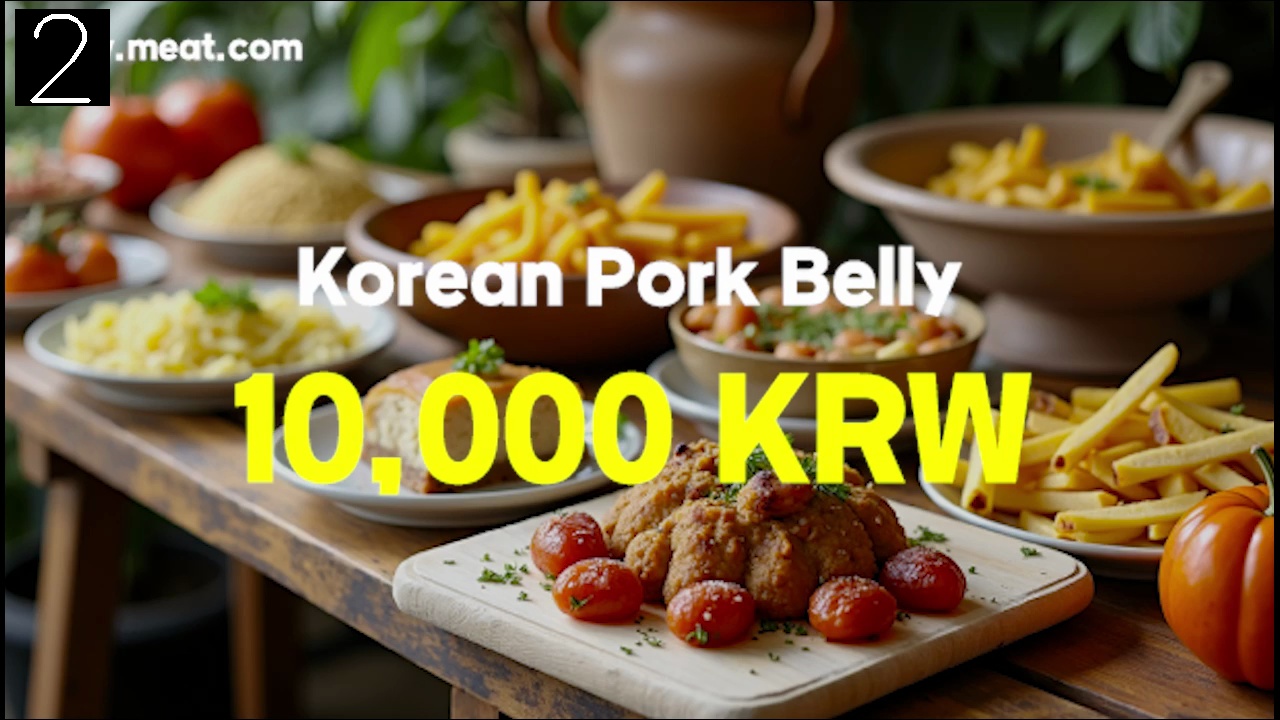} &
            \includegraphics[width=0.2\linewidth]{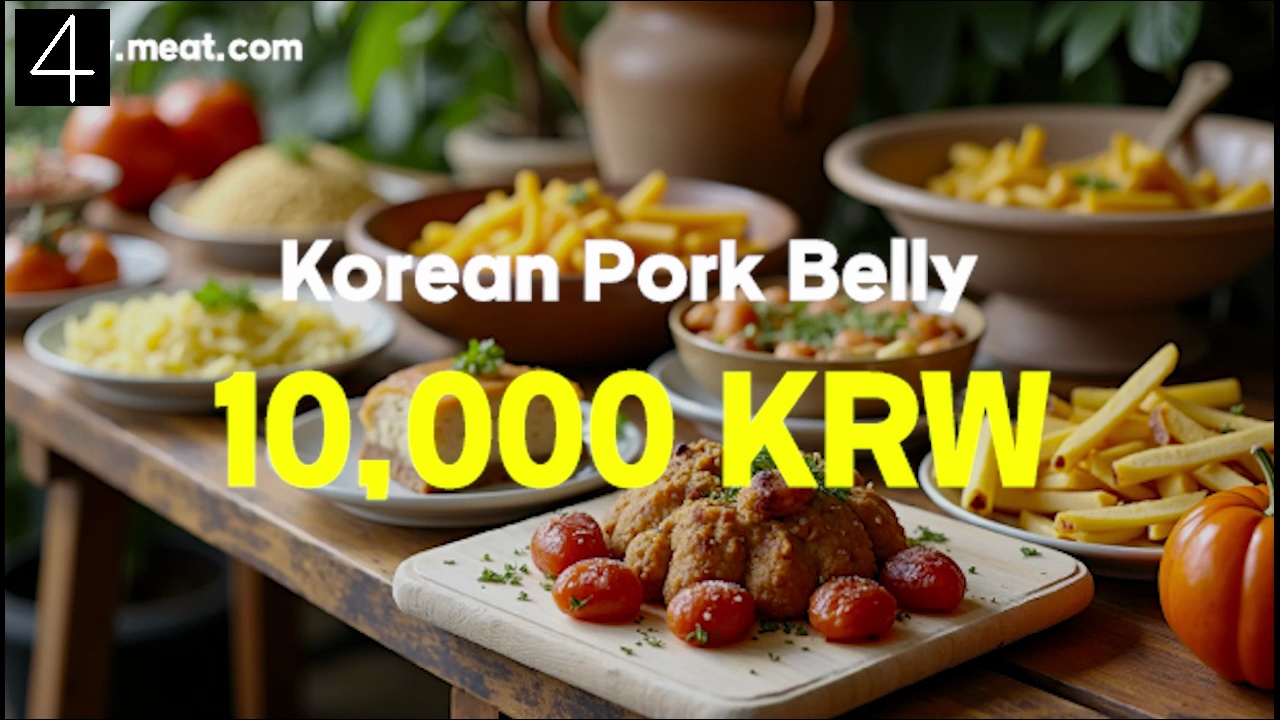} &
            \includegraphics[width=0.2\linewidth]{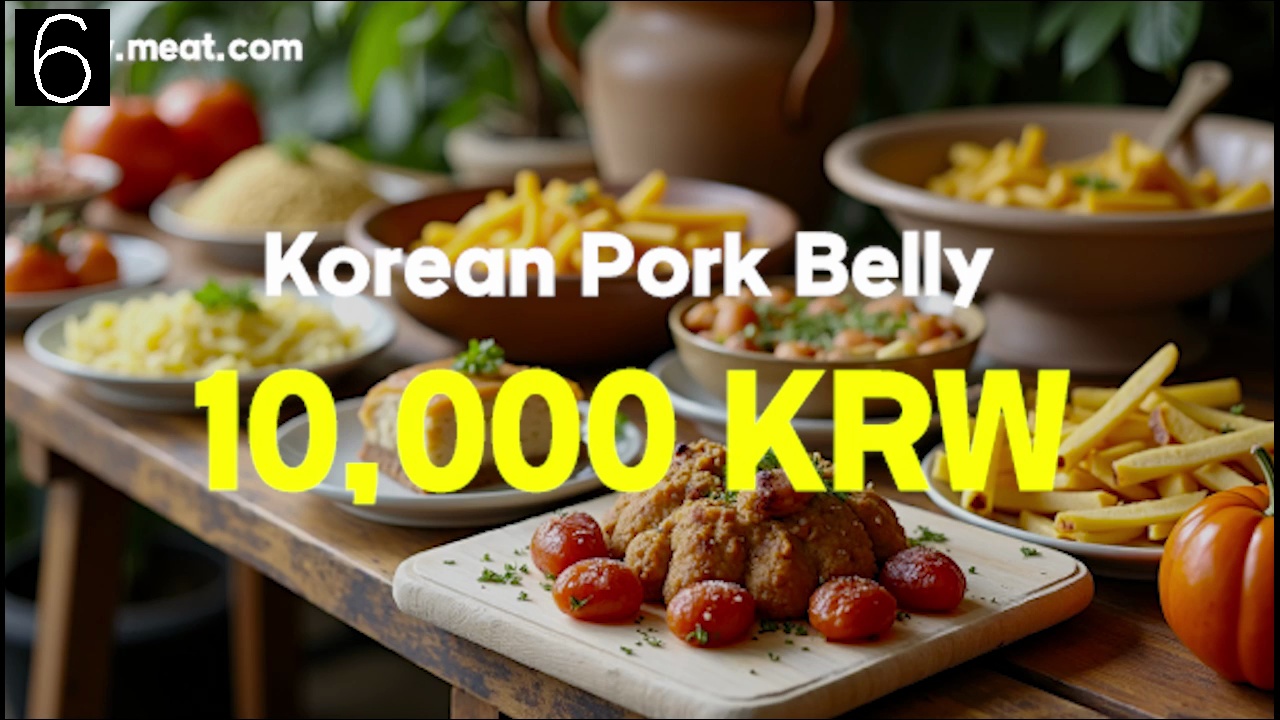} &
            \includegraphics[width=0.2\linewidth]{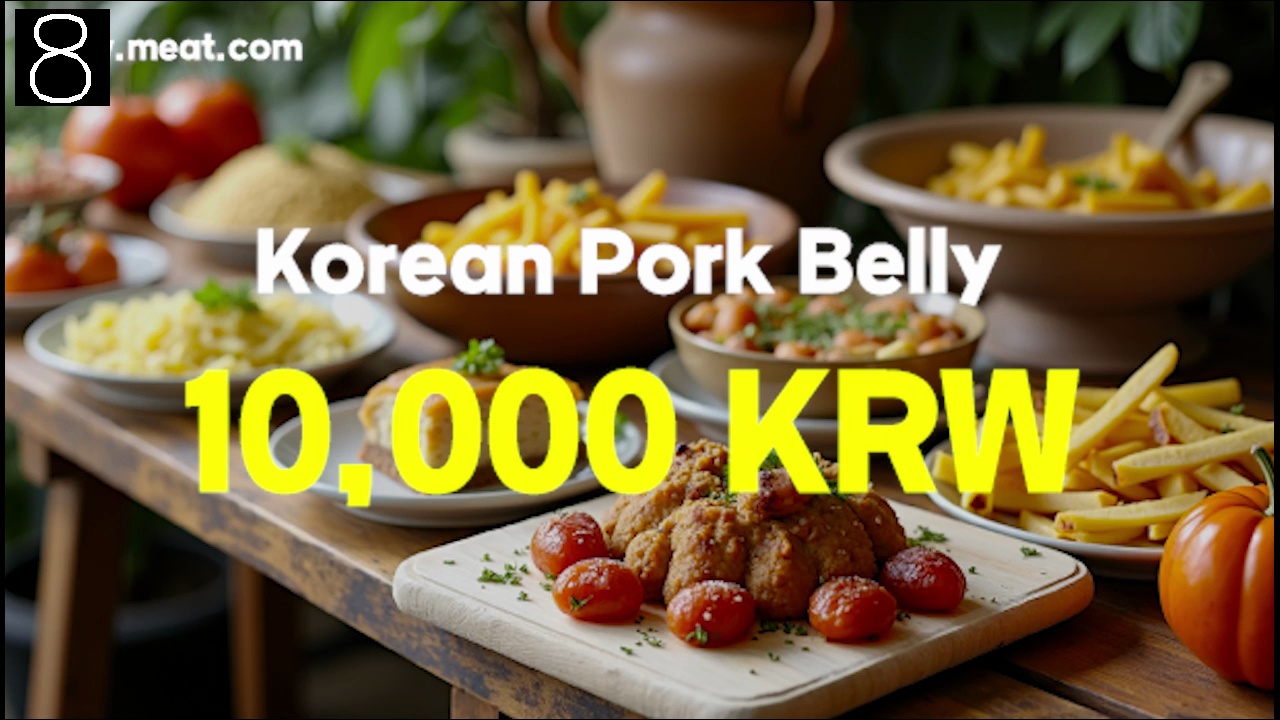} \\
            \multicolumn{5}{c}{\parbox{\linewidth}{\small (d) \textsf{"The top-left banner includes a single text stating, ``http://www.meat.com." There is no top-right and bottom banner.
The advertisement scene consists of two text objects. The first text displays ``Korean Pork Belly." The second text shows the price of the product as ``10,000 KRW."}\vspace{3mm}}}\\
            \includegraphics[width=0.2\linewidth]{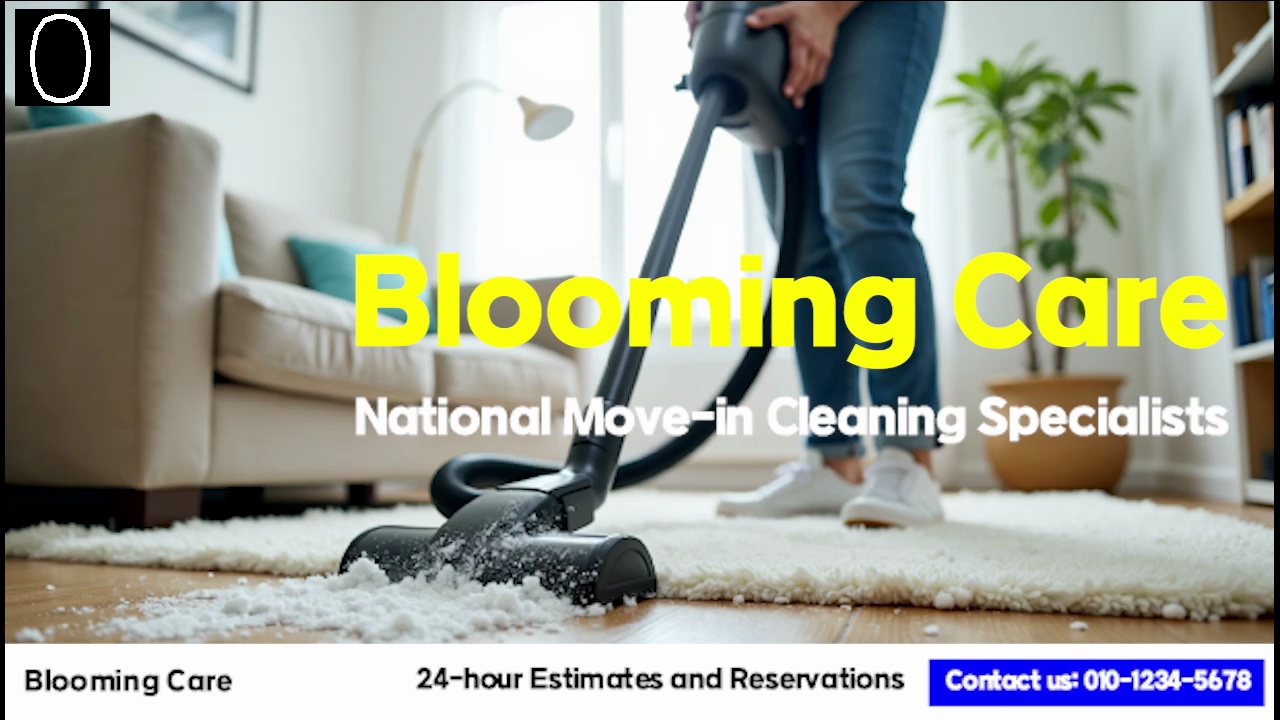} &
            \includegraphics[width=0.2\linewidth]{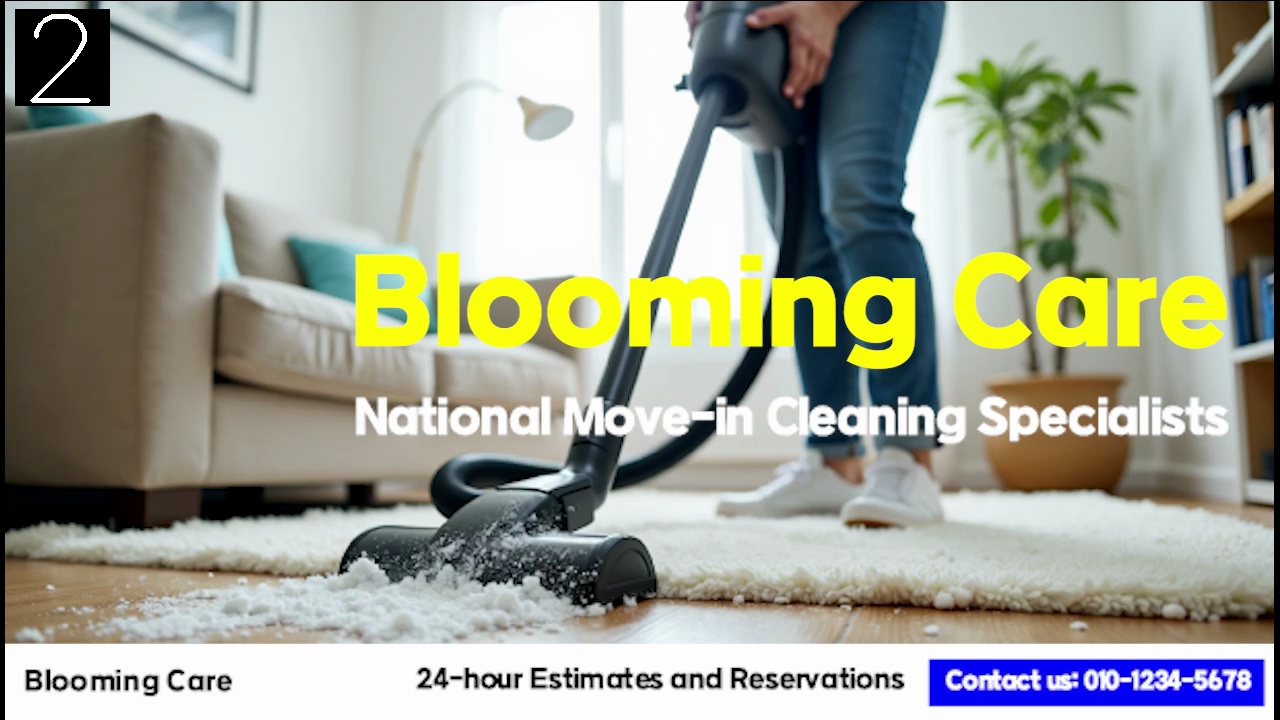} &
            \includegraphics[width=0.2\linewidth]{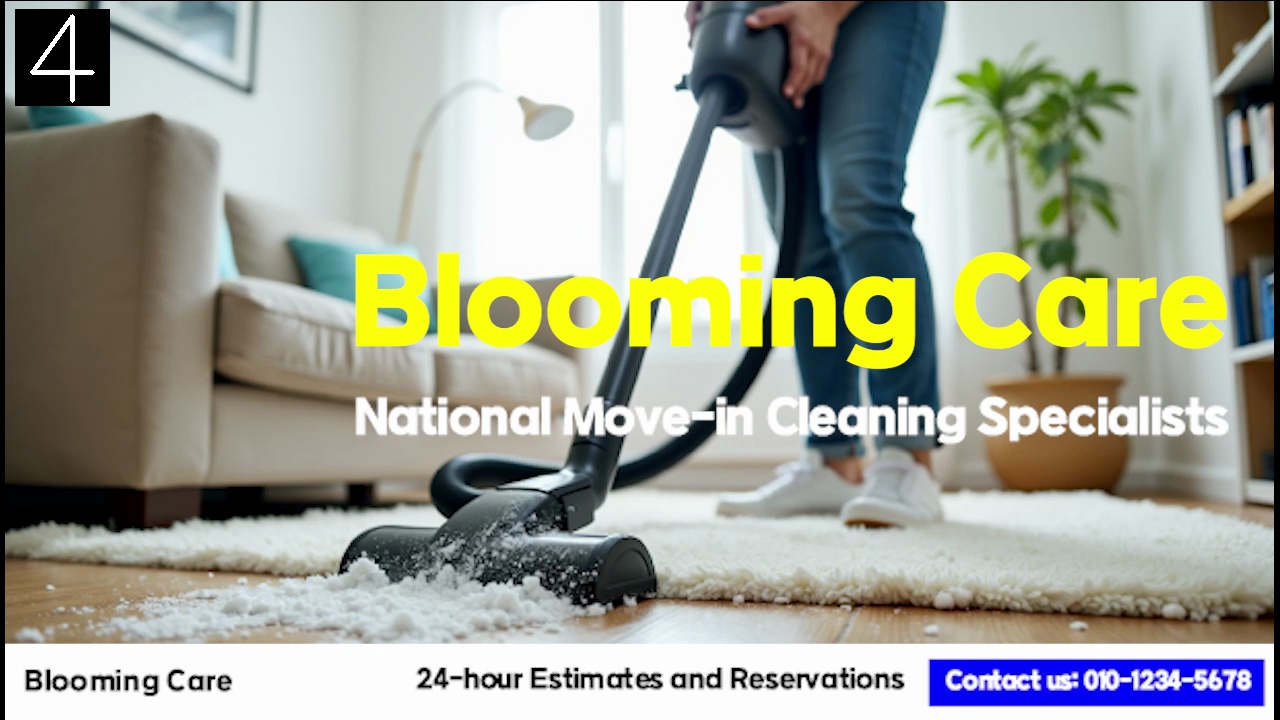} &
            \includegraphics[width=0.2\linewidth]{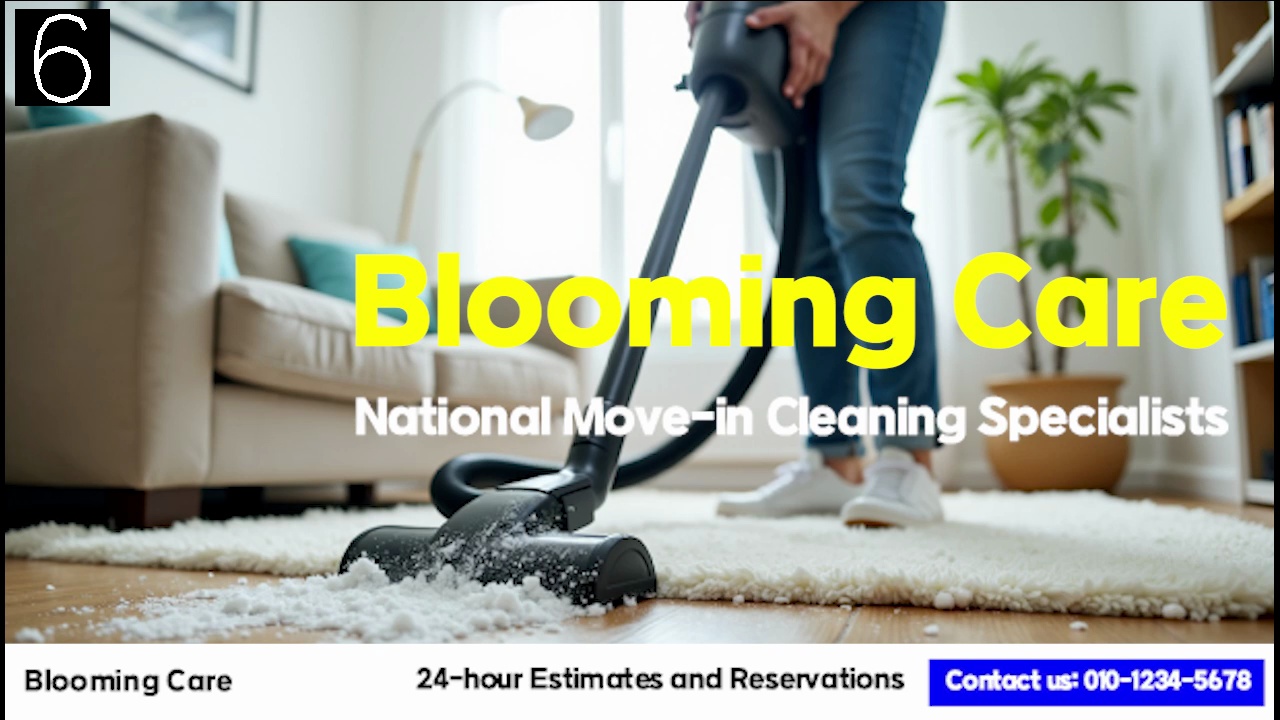} &
            \includegraphics[width=0.2\linewidth]{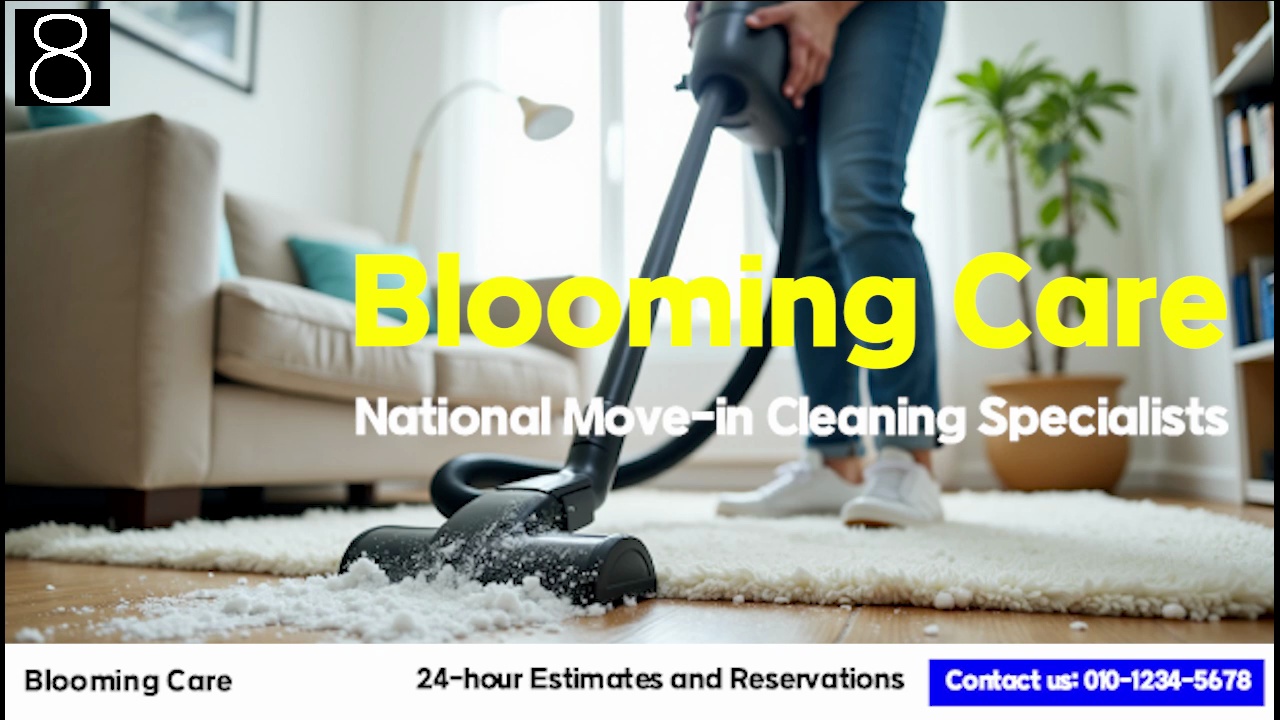} \\
            \multicolumn{5}{c}{\parbox{\linewidth}{\small (e) \textsf{The bottom banner contains three text objects: ``Blooming Care," ``24-hour Estimates and Reservations," and ``Contact us: 010-1234-5678." There are no banners at the top-left or top-right of the screen.
The advertisement features two text objects. The first text object displays the content ``Blooming Care." The second text object reads ``National Move-In Cleaning Specialist."}\vspace{3mm}}}\\
            \includegraphics[width=0.2\linewidth]{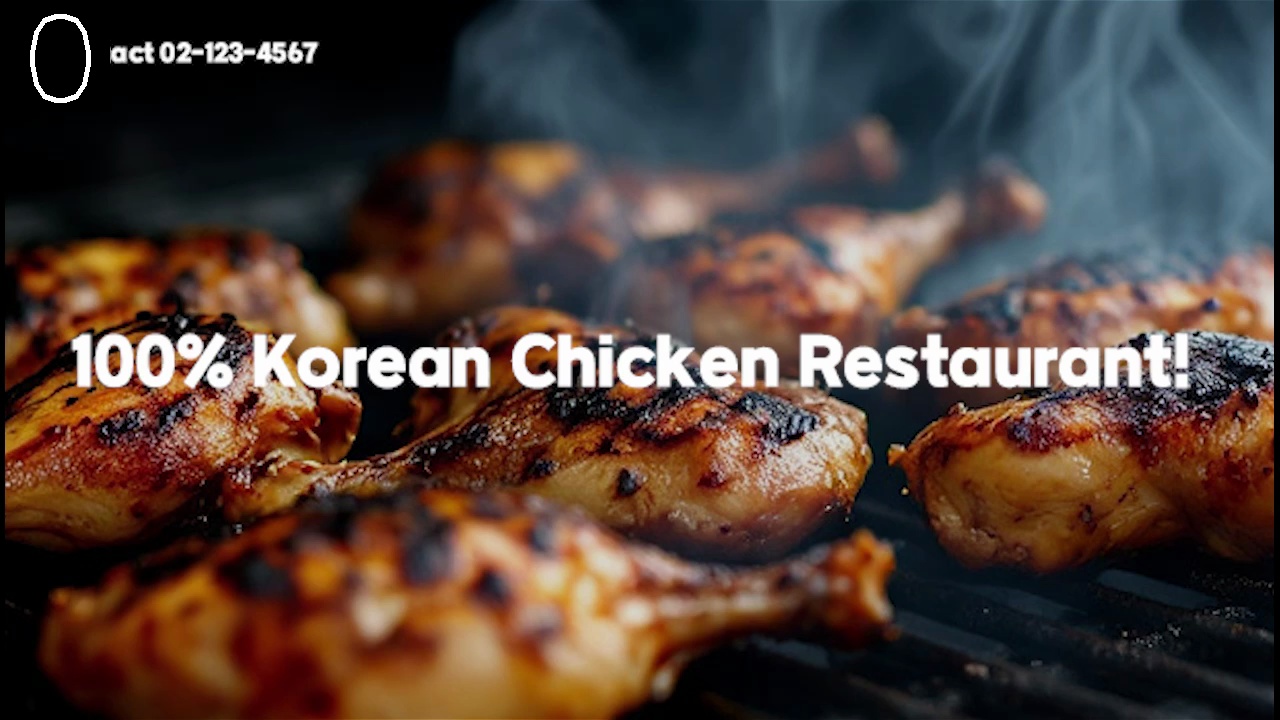} &
            \includegraphics[width=0.2\linewidth]{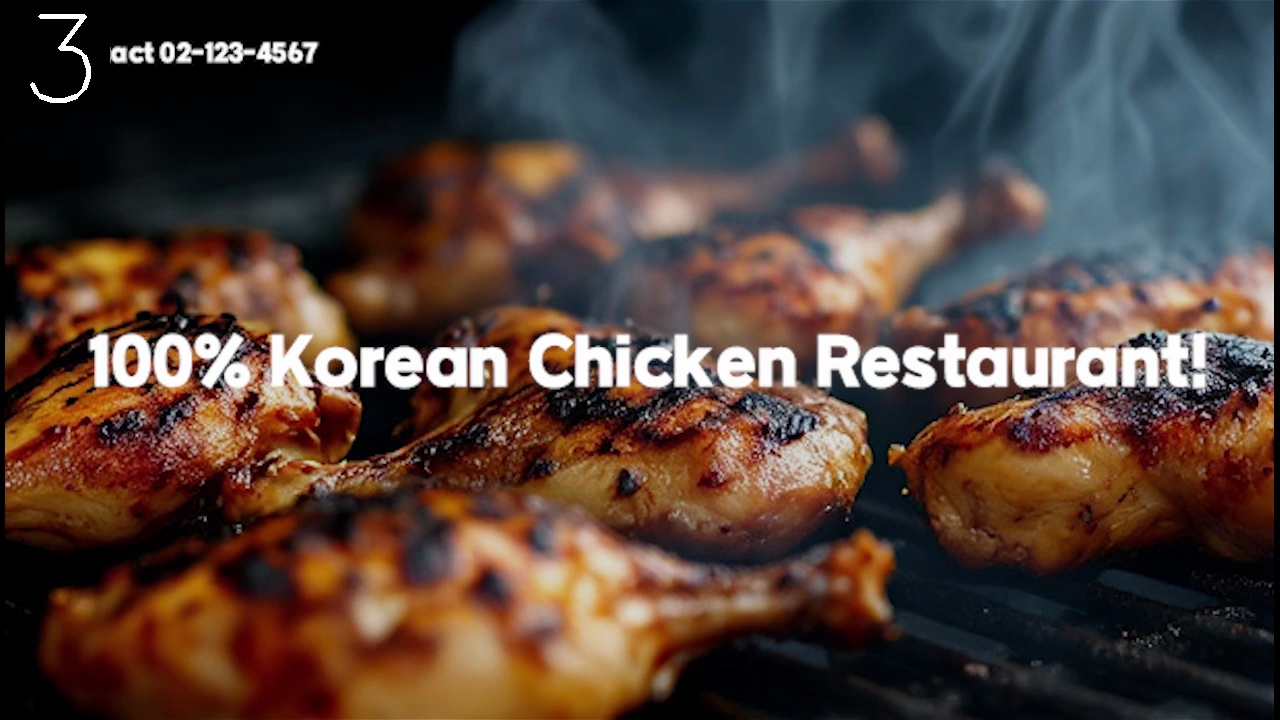} &
            \includegraphics[width=0.2\linewidth]{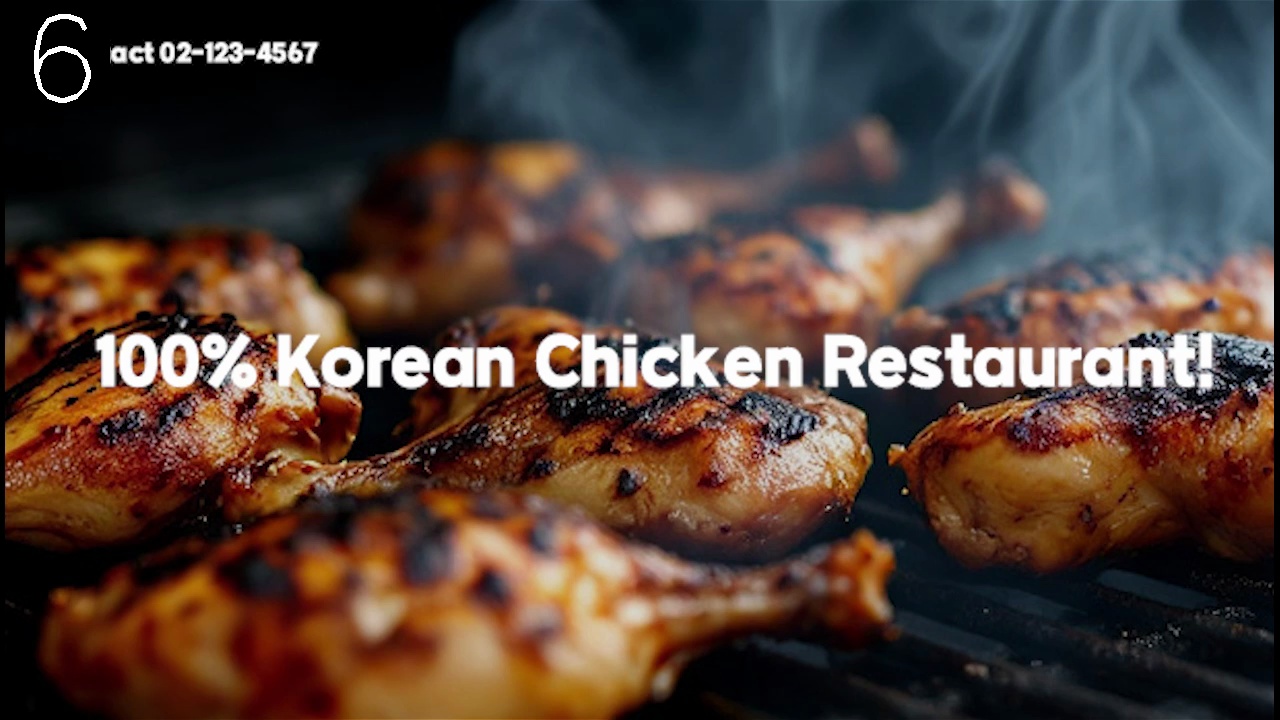} &
            \includegraphics[width=0.2\linewidth]{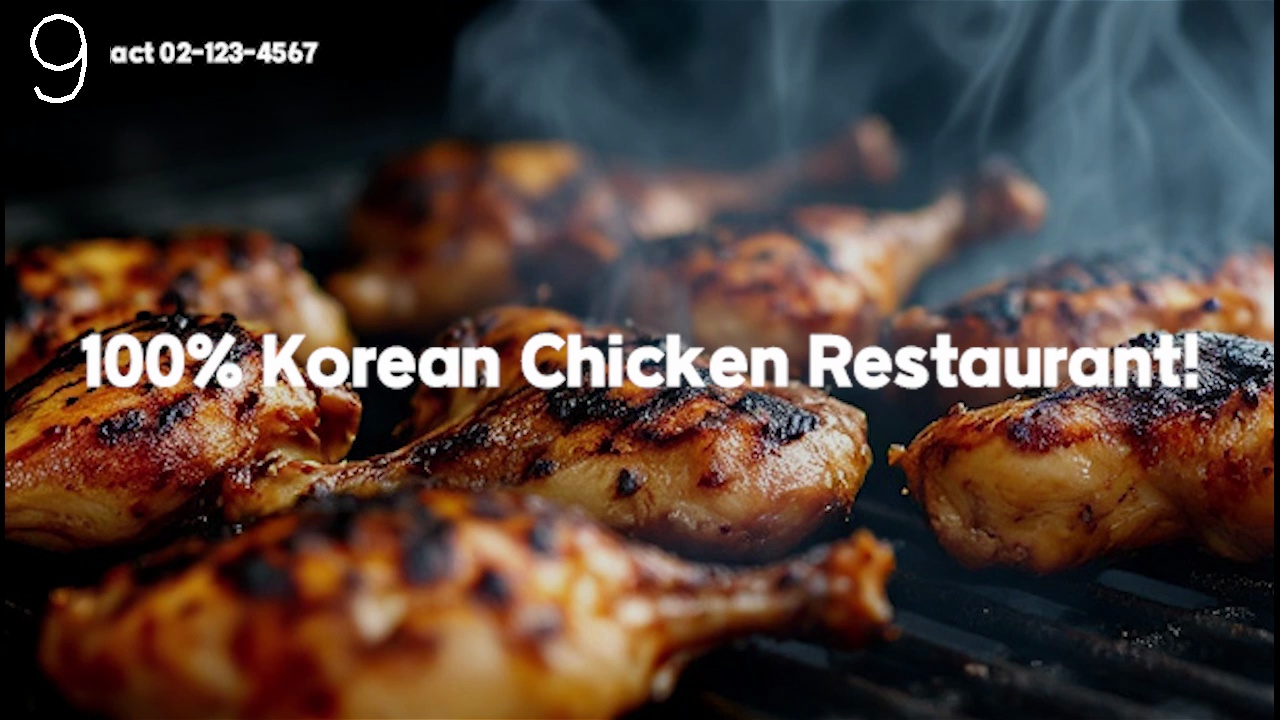} &
            \includegraphics[width=0.2\linewidth]{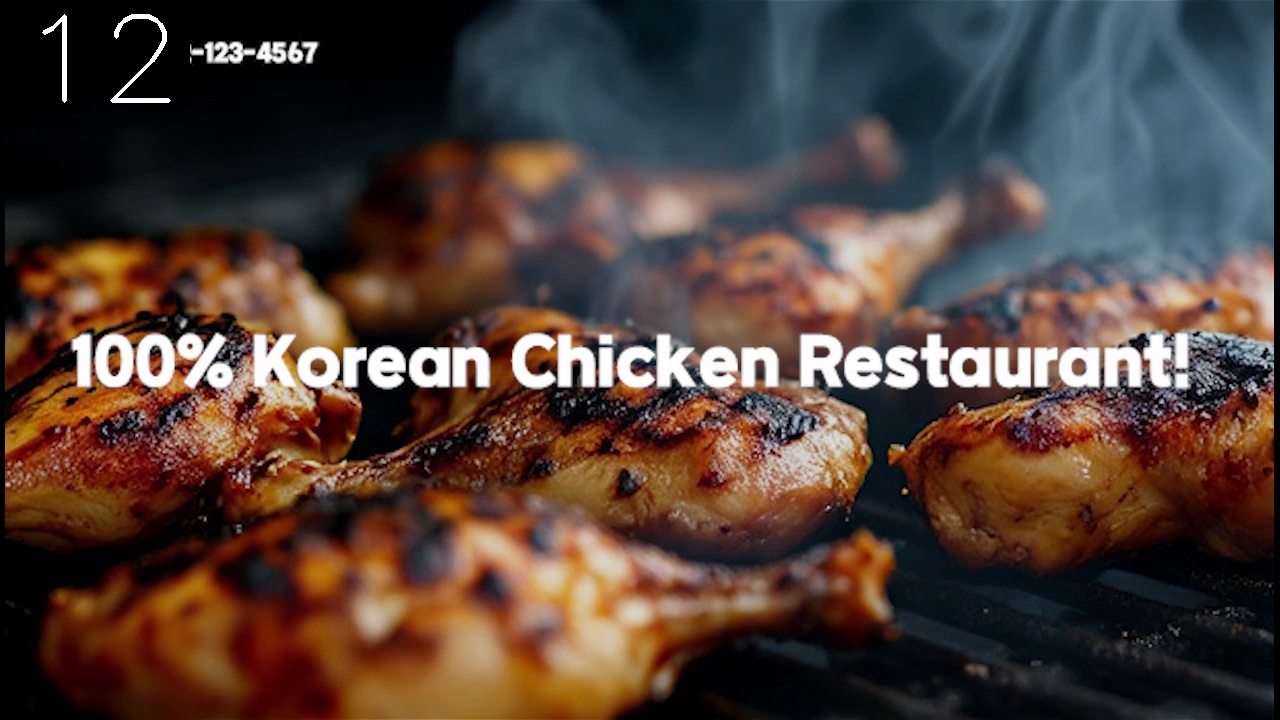} \\
            \multicolumn{5}{c}{\parbox{\linewidth}{\small (f) \textsf{The top-left banner contains a text object displaying the message ``Contact 02-123-4567." There are no banners at the bottom or top-right positions.
The scene features an image of grilled or cooked chicken in the background. Overlaying the image is text that reads ``100\% Korean Chicken Restaurant!". The text presented is clearly visible against the background.}\vspace{3mm}}}\\
        \end{tabular}
    }
        \caption{
            Videos generated by VAKER.
            The number on the top-left corner of each frame indicates the frame index.
        } \label{fig:qual_video_1}
\end{figure*}

In \cref{fig:qual_video,fig:qual_video_1}, we provide video results generated by VAKER.
To effectively visualize the generated animations, we sample frames at intervals of 2 to 5 frames.
\vspace{5mm}

\clearpage

\section{ST-Representation for VAKER}
\label{sec:st_rep_vaker}

We present examples of all ST-Representation components for VAKER: Banner Prompt, Banner UT, Banner ST, Mainground Prompt, Mainground UT, Mainground ST, Animation UT, and Animation ST.
For data privacy, all brand names, locations, and numerical information have been modified --- numbers are standardized to 012.345.6789 and specific locations are set to 0.

\begin{figure}[h]
    \centering
    \includegraphics[width=0.9\linewidth]{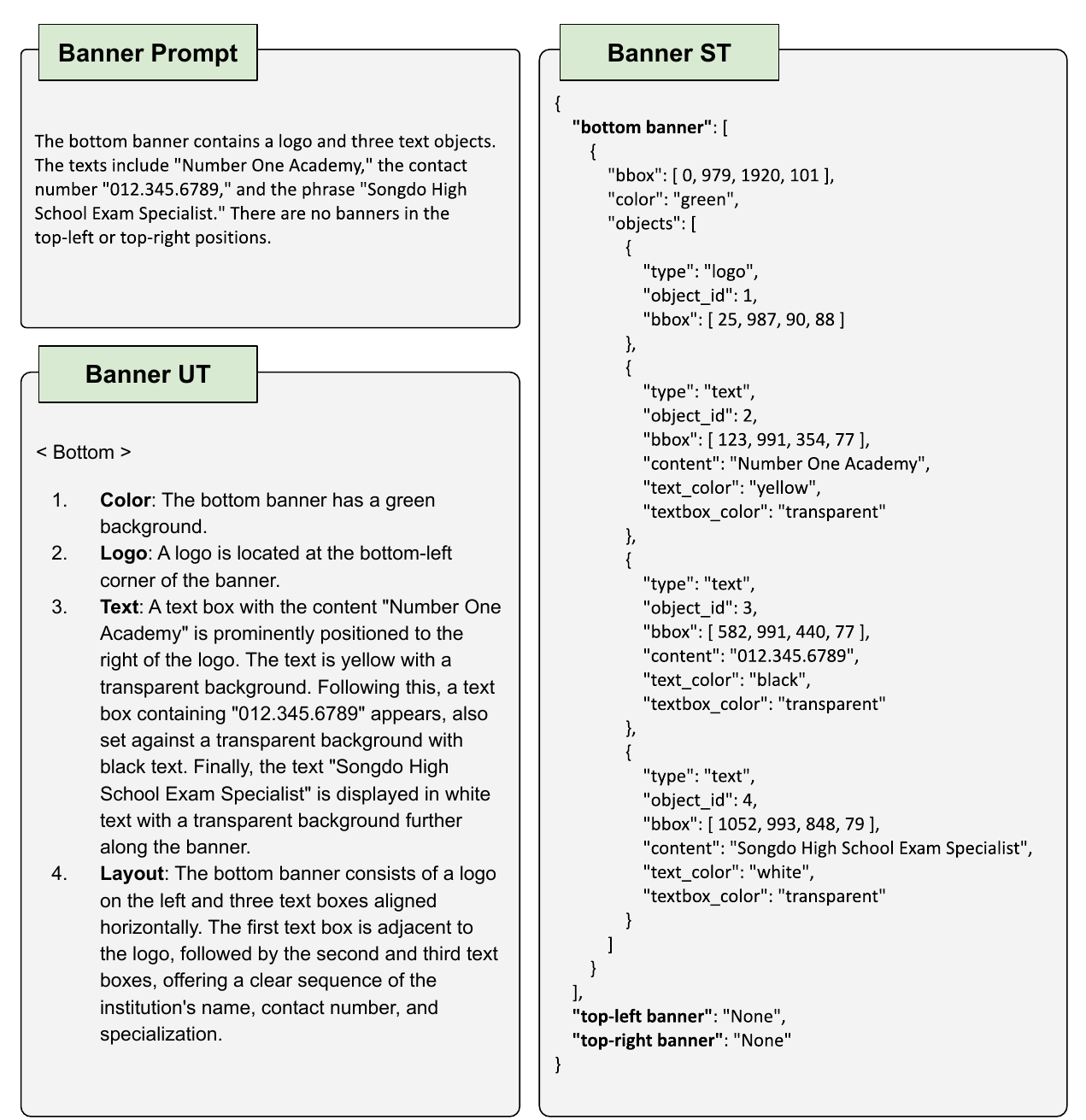}
    \caption{Example 1 for Banner Prompt, Banner UT, and Banner ST}
    \label{fig:strep_banner_1}
\end{figure}

\begin{figure}[h]
    \centering
    \includegraphics[width=0.9\linewidth]{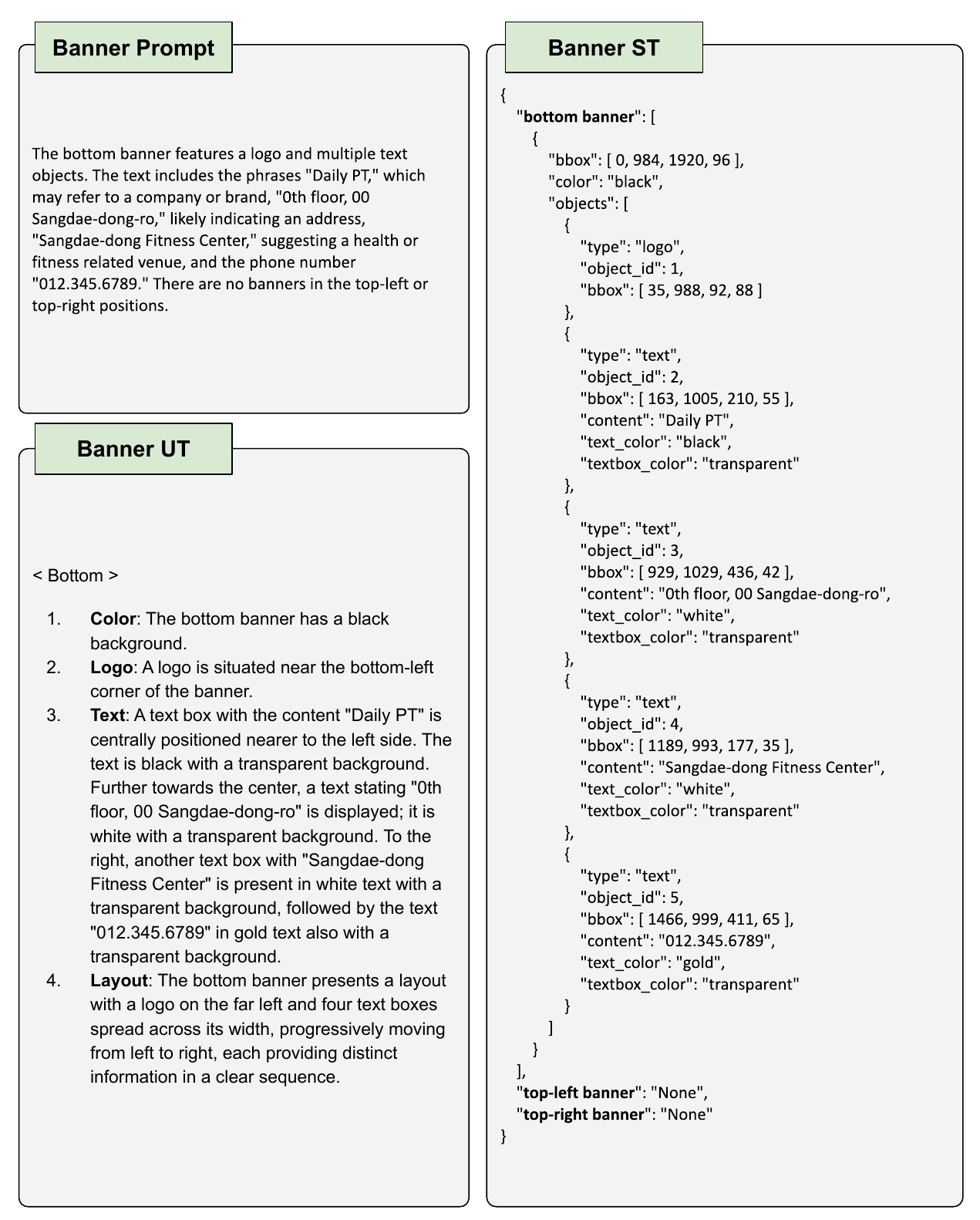}
    \caption{Example 2 for Banner Prompt, Banner UT, and Banner ST}
    \label{fig:strep_banner_2}
\end{figure}

\begin{figure}[h]
    \centering
    \includegraphics[width=0.9\linewidth]{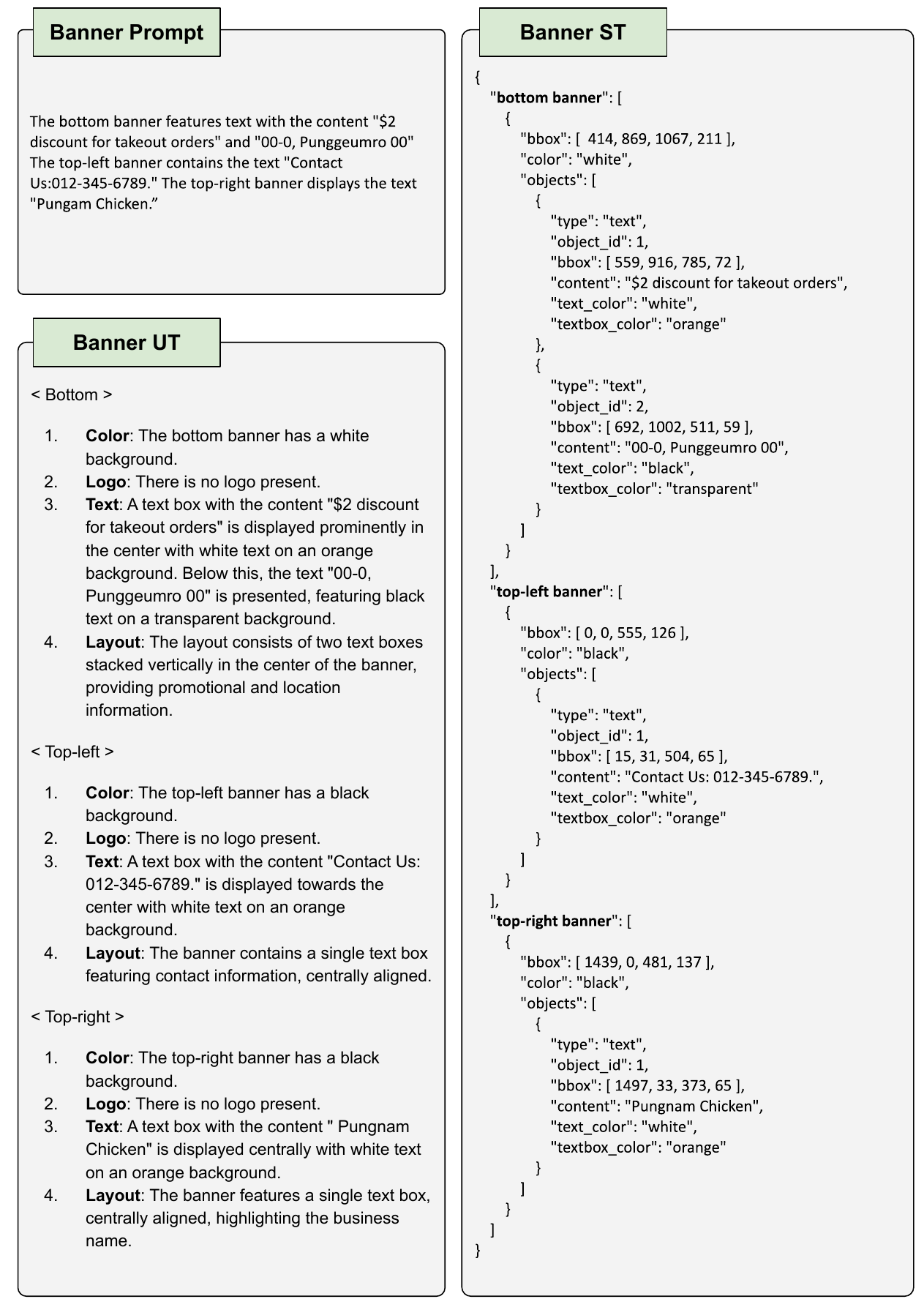}
    \caption{Example 3 for Banner Prompt, Banner UT, and Banner ST}
    \label{fig:strep_banner_3}
\end{figure}

\begin{figure}[h]
    \centering
    \includegraphics[width=0.9\linewidth]{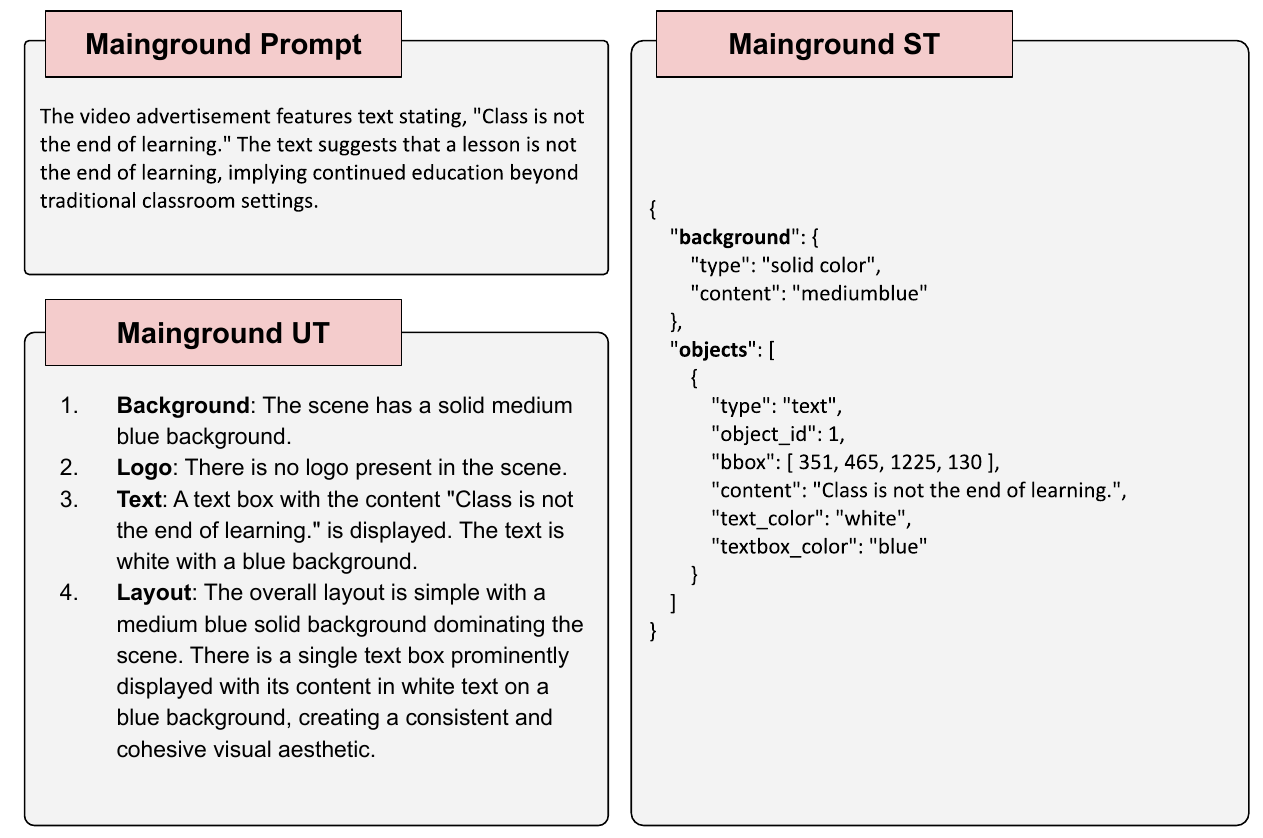}
    \caption{Example 1 for Mainground Prompt, Mainground UT, and Mainground ST}
    \label{fig:strep_mainground_1}
\end{figure}

\begin{figure}[h]
    \centering
    \includegraphics[width=0.9\linewidth]{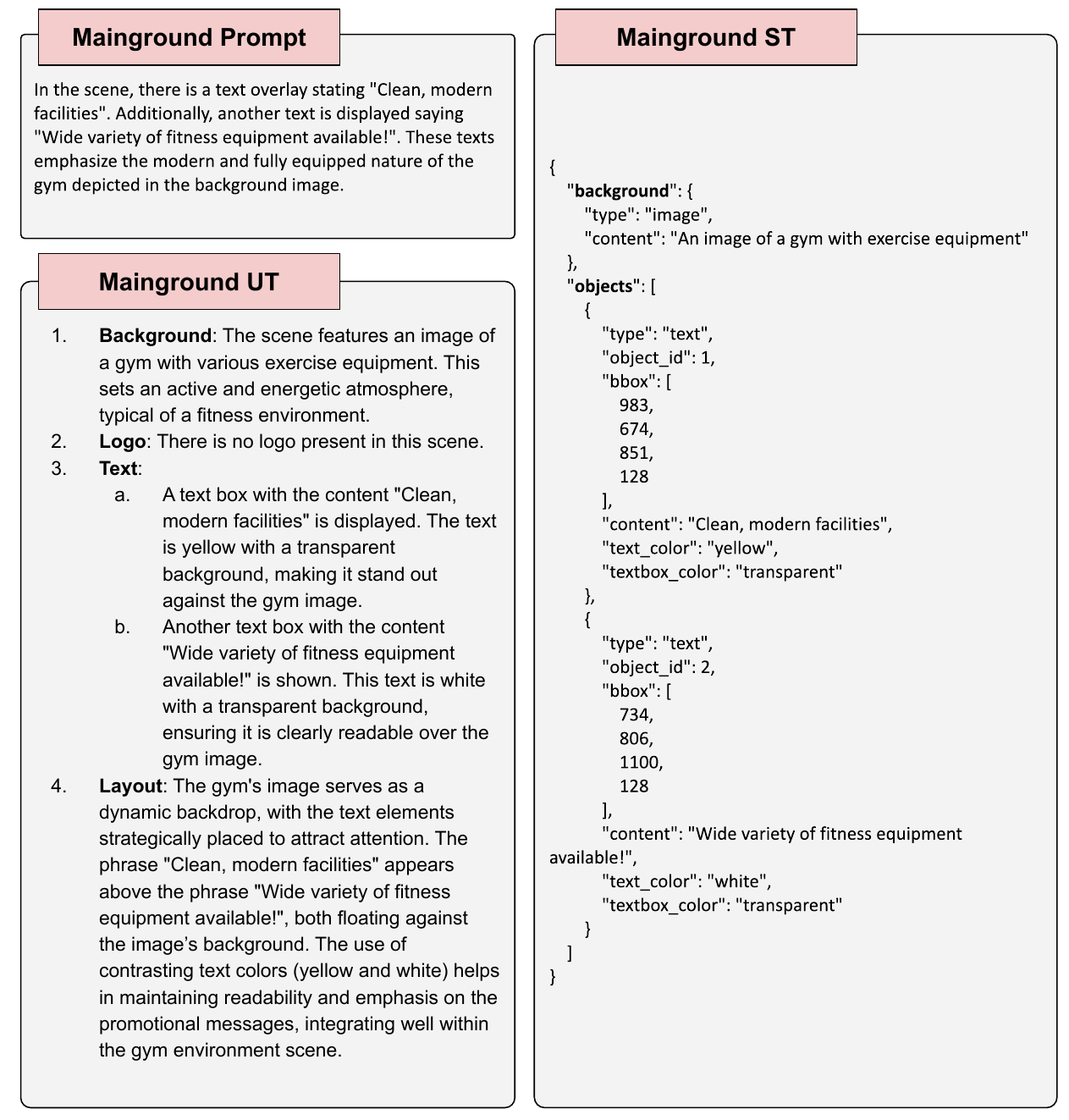}
    \caption{Example 2 for Mainground Prompt, Mainground UT, and Mainground ST}
    \label{fig:strep_mainground_2}
\end{figure}

\begin{figure}[h]
    \centering
    \includegraphics[width=0.9\linewidth]{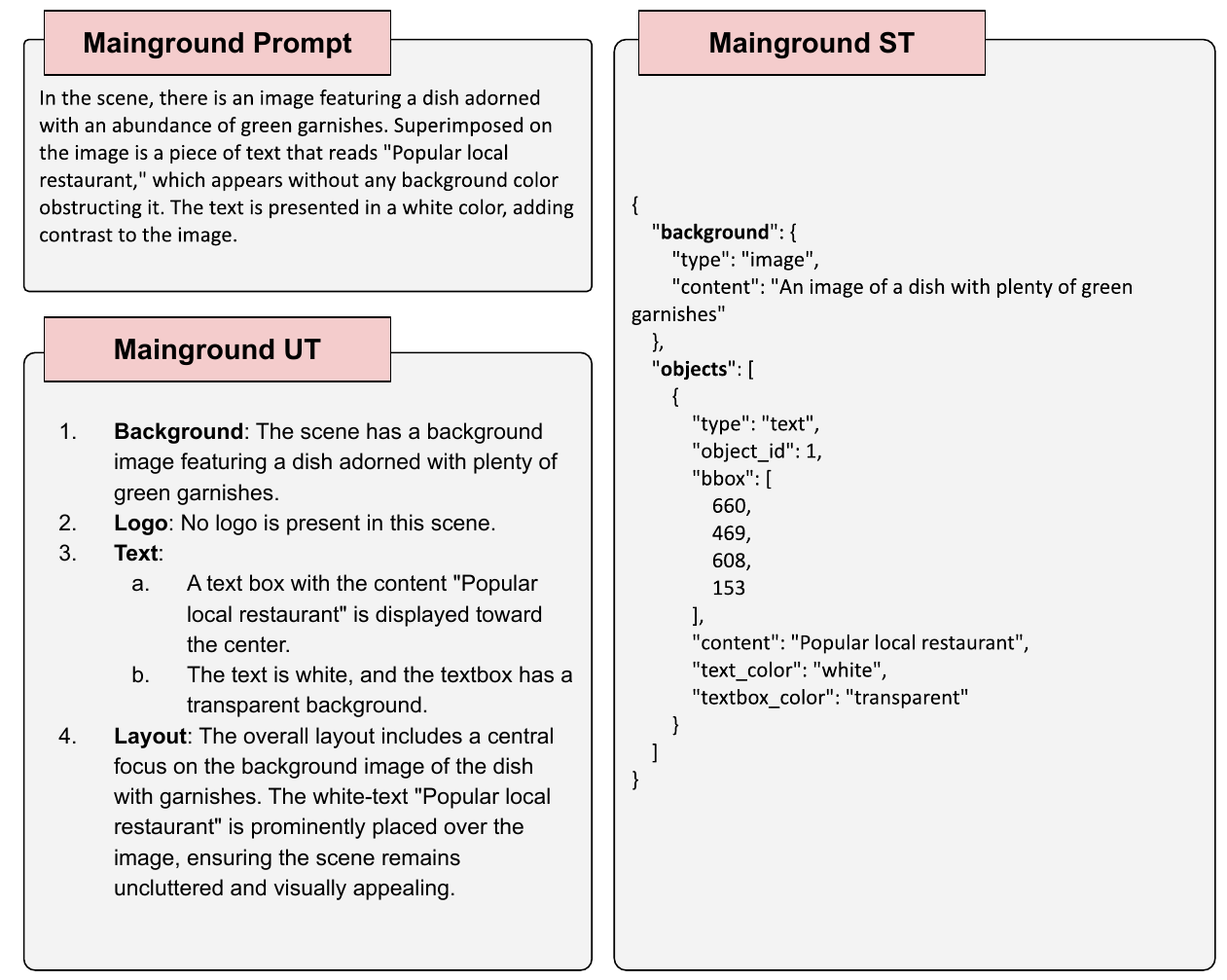}
    \caption{Example 3 for Mainground Prompt, Mainground UT, and Mainground ST}
    \label{fig:strep_mainground_3}
\end{figure}

\begin{figure}[h]
    \centering
    \includegraphics[width=0.9\linewidth]{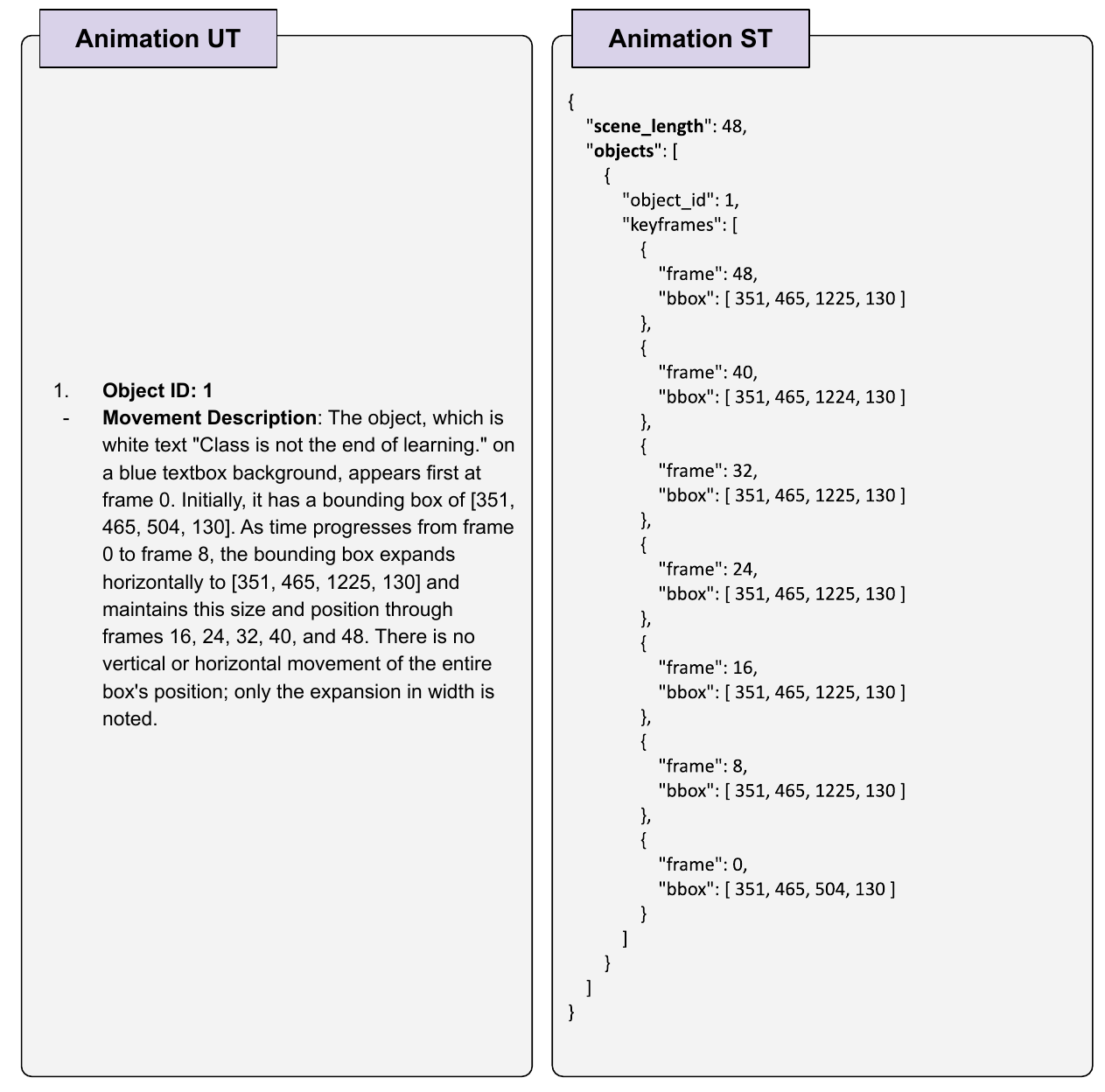}
    \caption{Example 1 for Animation UT and Animation ST}
    \label{fig:strep_animation_1}
\end{figure}

\begin{figure}[h]
    \centering
    \includegraphics[width=0.8\linewidth]{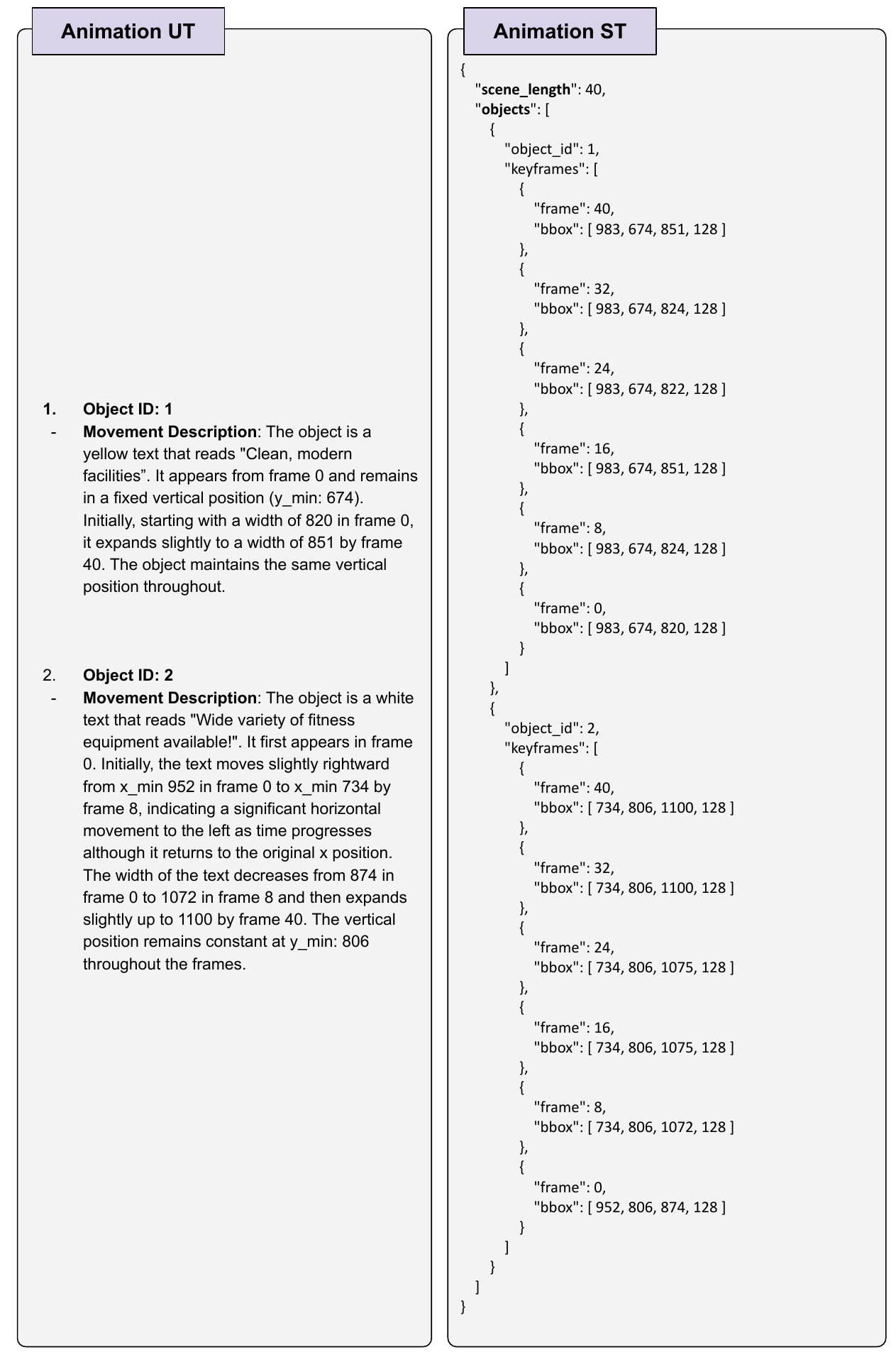}
    \caption{Example 2 for Animation UT and Animation ST}
    \label{fig:strep_animation_2}
\end{figure}

\begin{figure}[h]
    \centering
    \includegraphics[width=0.9\linewidth]{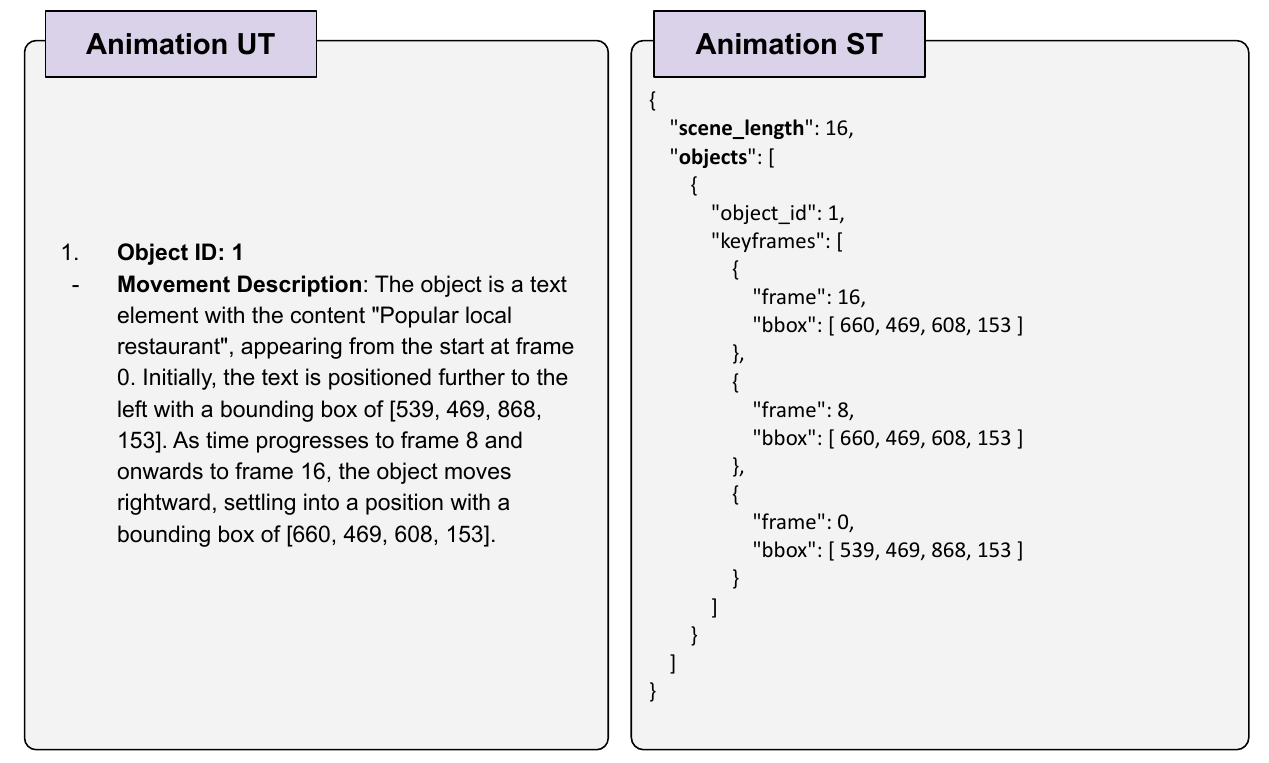}
    \caption{Example 3 for Animation UT and Animation ST}
    \label{fig:strep_animation_3}
\end{figure}

\begin{algorithm}[!t]
    \caption{Bounding Box Post-processing Algorithm}
    \label{alg:bbox}
    \begin{algorithmic}
        \Require~$pred\_mask$, $max\_values$, $is\_firstframe$
        \State \textbf{Output:} $boxes$, $max\_values$
        \vspace{3mm}
        \State $objects \leftarrow \text{unique}(pred\_mask) \setminus \{0\}$ \Comment{Remove background}
        \ForAll{$obj\_id \in objects$}
            \State $max\_value \leftarrow is\_firstframe~?~0 : max\_values[obj\_id - 1]$
            \State $mask \leftarrow pred\_mask = obj\_id$
            \State $y\_indices, x\_indices \leftarrow \text{where}(mask)$
            \vspace{3mm}
            
            \State /* Get vertical bounds via bidirectional scanning */
            \State $y\_max, y\_min \leftarrow \max(y\_indices), \min(y\_indices)$
            \State $y\_mid \leftarrow (y\_max + y\_min) / 2$
            \State $(y1\_1, y2\_1) \leftarrow \text{ScanTopDown}(mask, y\_min, y\_max)$ \Comment{Check early-stop at 70\%}
            \State $(y1\_2, y2\_2) \leftarrow \text{ScanBottomUp}(mask, y\_min, y\_max)$ \Comment{Check early-stop at 30\%}
            \If{$\text{BothEarlyStop}()$ \textbf{and} $\text{HasContent}(mask[y\_mid, :])$}
                \State $y1, y2 \leftarrow \text{ScanFromMiddle}(mask, y\_mid)$
            \Else
                \State $y1, y2 \leftarrow \text{SelectBestBounds}(y1\_1, y2\_1, y1\_2, y2\_2)$
            \EndIf
            \vspace{3mm}
            
            \State /* Get horizontal bounds */
            \State $x1, x2 \leftarrow \text{GetHorizontalBounds}(mask[y1:y2+1, :])$
            \vspace{3mm}
            
            \State /* Refine boundaries */
            \State $box\_xy \leftarrow \text{ScanXthenY}(mask, x1, y1, x2, y2)$ \Comment{50\% ratio}
            \State $box\_yx \leftarrow \text{ScanYthenX}(mask, x1, y1, x2, y2)$ \Comment{50\% ratio}
            \State $final\_box \leftarrow \text{SelectLargerBox}(box\_xy, box\_yx)$
            \vspace{3mm}
            
            \State /* Size filtering */
            \State $size \leftarrow (final\_box.x2 - final\_box.x1) \times (final\_box.y2 - final\_box.y1)$
            \If{$is\_firstframe$ \textbf{or} $size \geq 0.2 \times max\_value$}
                \State $boxes[obj\_id] \leftarrow final\_box$
                \If{$is\_firstframe$}
                    \State $max\_values.\text{append}(size)$
                \EndIf
            \EndIf
        \EndFor
        \State \textbf{return} $boxes, max\_values$
    \end{algorithmic}
\end{algorithm}

\clearpage
\section{Prompt template}
\begin{figure}[h]
    \centering
    \includegraphics[width=0.9\linewidth]{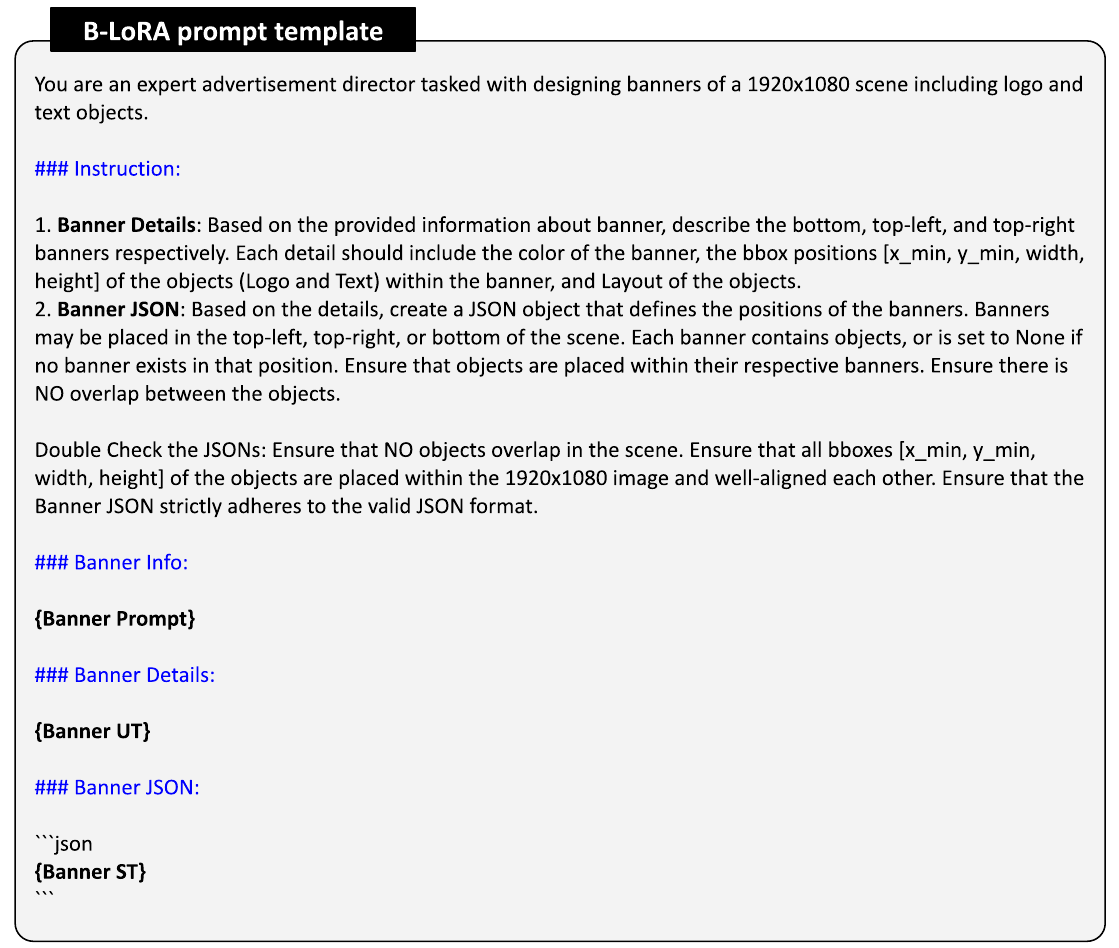}
    \caption{B-LoRA prompt template.}
    \label{fig:blora_prompt}
\end{figure}

\begin{figure}
    \centering
    \includegraphics[width=0.9\linewidth]{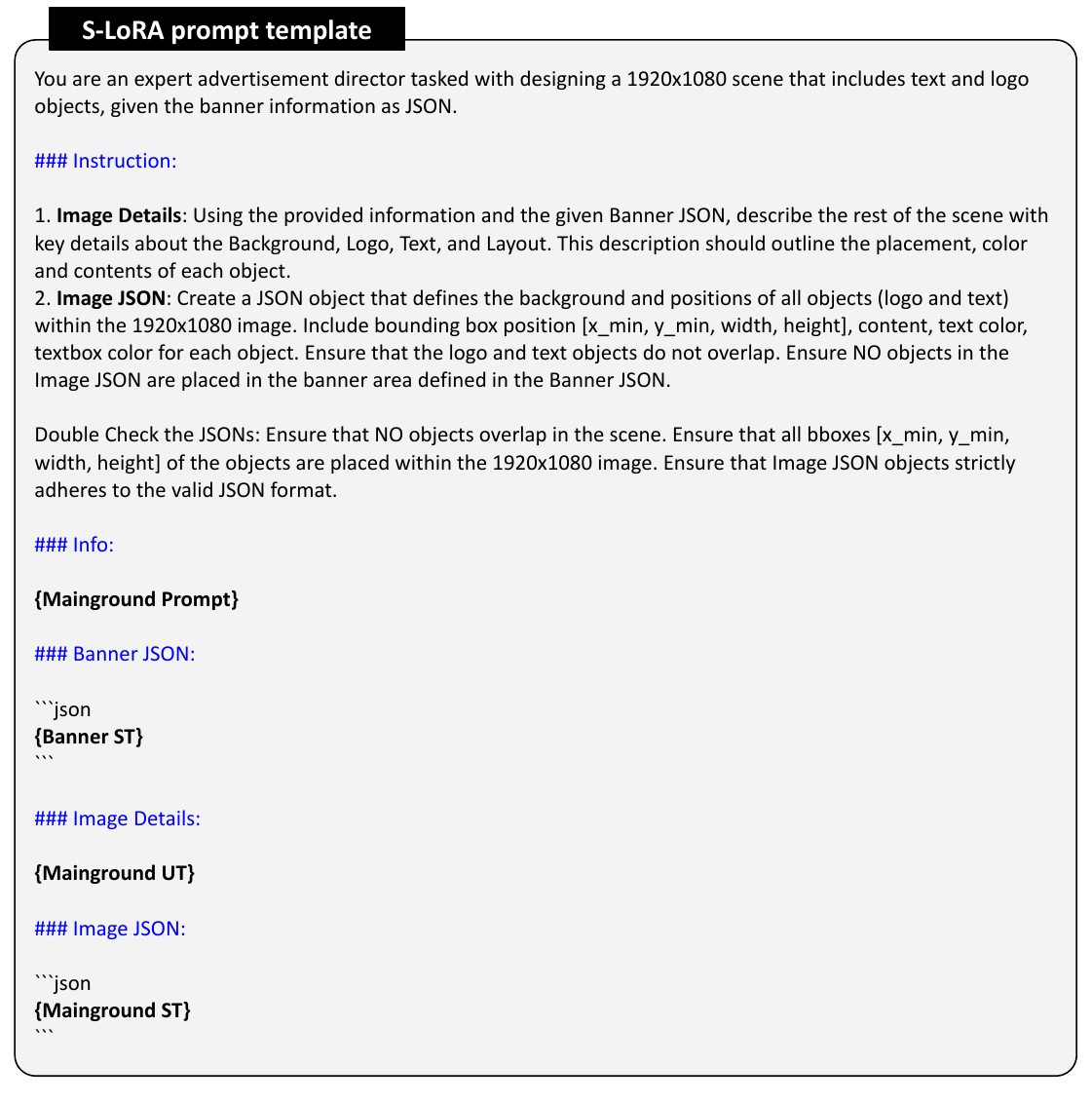}
    \caption{S-LoRA prompt template.}
    \label{fig:slora_prompt}
\end{figure}

\begin{figure}
    \centering
    \includegraphics[width=0.9\linewidth]{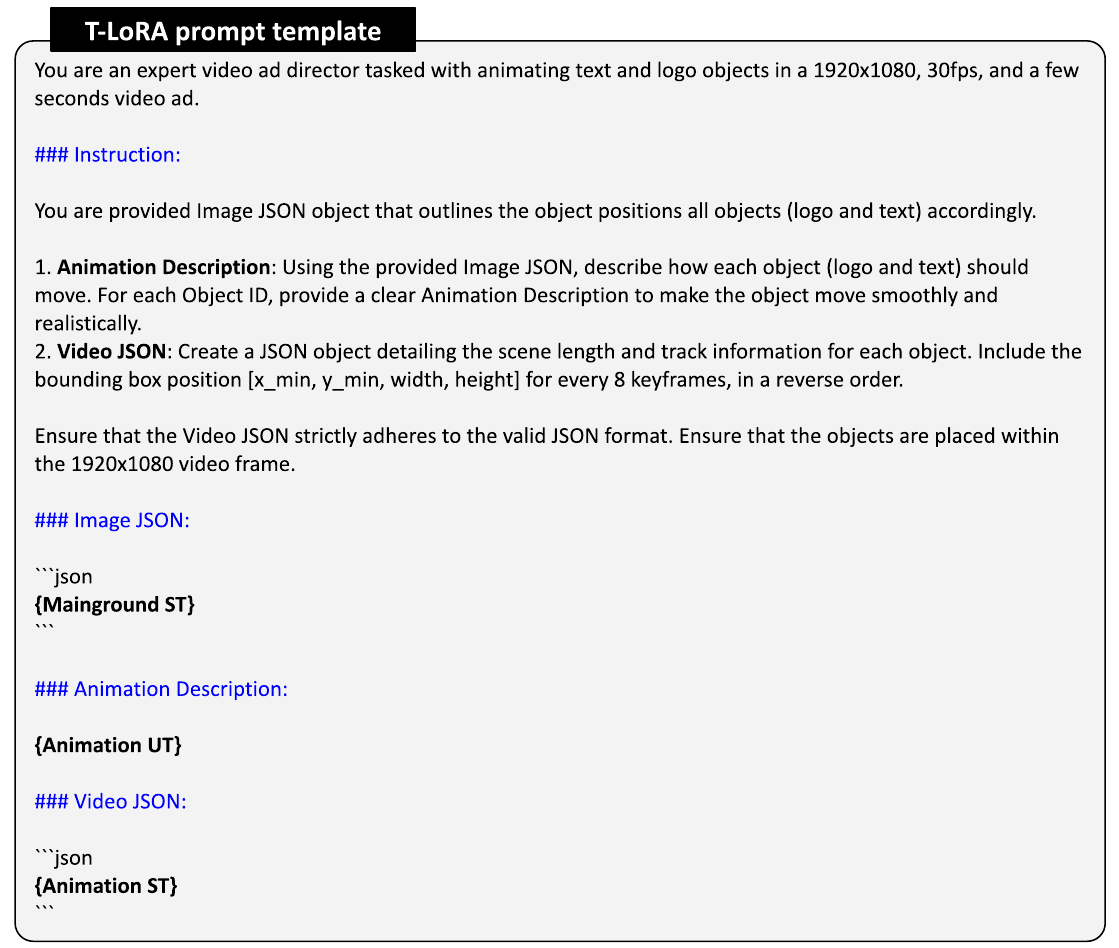}
    \caption{T-LoRA prompt template.}
    \label{fig:tlora_prompt}
\end{figure}

We provide the prompt templates for training B-LoRA~(\cref{fig:blora_prompt}), S-LoRA~(\cref{fig:slora_prompt}), and T-LoRA~(\cref{fig:tlora_prompt}).
\vspace{10mm}

\end{document}